\documentclass[lettersize,journal]{IEEEtran}

\usepackage{tikz}
\usetikzlibrary{arrows.meta, positioning}

\usepackage{amsmath,amsfonts}
\usepackage{algorithmic}
\usepackage{algorithm}
\usepackage{array}
\usepackage{textcomp}
\usepackage{url}
\usepackage{verbatim}
\usepackage{graphicx}
\usepackage{booktabs}
\usepackage{subcaption}
\usepackage{cite}
\usepackage{dsfont}
\usepackage{bm}
\usepackage{multirow}

\usepackage{siunitx}
\usepackage{hyperref}
\usepackage{listings}
\usepackage{xspace}
\usepackage{rotating}
\usepackage{breqn}
\usepackage{titlesec}
\usepackage{placeins}
\usepackage{dblfloatfix}
\usepackage{xcolor}

\definecolor{mygreen}{rgb}{0,0.6,0}
\definecolor{mygray}{rgb}{0.5,0.5,0.5}
\definecolor{mymauve}{rgb}{0.58,0,0.82}
\definecolor{myblue}{rgb}{0,0,0}  

\lstdefinestyle{mystyle}{
    backgroundcolor=\color{white},      
    commentstyle=\color{mygreen},       
    keywordstyle=\color{myblue},        
    identifierstyle=\color{black},      
    numberstyle=\tiny\color{mygray},    
    stringstyle=\color{mymauve},        
    basicstyle=\ttfamily\footnotesize,  
    breakatwhitespace=false,            
    breaklines=true,                    
    captionpos=b,                       
    keepspaces=true,                    
    numbers=left,                       
    numbersep=2pt,                    
    showspaces=false,                   
    showstringspaces=false,             
    showtabs=false,                     
    tabsize=2,                          
    language=Python,
    literate={all}{{\textcolor{black}{all}}}3
             {any}{{\textcolor{black}{any}}}3
             {sum}{{\textcolor{black}{sum}}}3,
    xleftmargin=0pt,                    
}

\usepackage[capitalize,noabbrev]{cleveref}
\usepackage[textsize=tiny]{todonotes}

\DeclareMathOperator{\argmin}{argmin}

\DeclareMathOperator{\cov}{cov}

\crefname{figure}{Fig.}{Figs.}
\crefname{table}{Table}{Tables}
\crefname{equation}{Eq.}{Eqs.}

\titleformat{\subsubsection}{\normalfont\normalsize\itshape}{\arabic{subsubsection})}{1em}{}

\author{Tao Jiang, 
        Kebin Sun,
        Zhenyu Liang,
        Ran Cheng, 
        Yaochu Jin,
        and Kay Chen Tan
        
\thanks{The first two authors contributed equally to this work.} 

}

\begin{document}

\title{Evolutionary Generative Optimization: Towards Fully Data-Driven Evolutionary Optimization via Generative Learning}

\maketitle

\newcommand{\change}[1]{#1}
\newcommand{\typo}[1]{#1}
\newcommand{\SmallTitle}[1]{\emph{#1}~~}
\newcommand{\MethodName}{EvoGO\xspace}
\makeatletter
\newcommand{\FullMethodName}{\@ifstar{\@FullMethodNamestar}{\@FullMethodNamenostar}} 
\newcommand\@FullMethodNamestar{evolutionary generative optimization}
\newcommand\@FullMethodNamenostar{\textbf{Evo}lutionary \textbf{G}enerative \textbf{O}ptimization (\MethodName)\xspace}
\makeatother


\newcommand{\subfigPerf}[1]{ %
\begin{subfigure}[c]{0.325\linewidth} %
    \raggedleft %
    \includegraphics[scale=0.525]{figs/ablation/#1_1.pdf} %
    \vskip -0.25em %
    \caption{Ackley (200D)} %
\end{subfigure} %
\begin{subfigure}[c]{0.325\linewidth} %
    \raggedleft %
    \includegraphics[scale=0.525]{figs/ablation/#1_2.pdf} %
    \vskip -0.25em %
    \caption{Rosenbrock (200D)} %
\end{subfigure} %
%
\begin{subfigure}[c]{0.325\linewidth} %
    \raggedleft %
    \includegraphics[scale=0.525]{figs/ablation/#1_3.pdf} %
    \vskip -0.25em %
    \caption{Rastrigin (200D)} %
\end{subfigure} %
\hfill \vspace{0.5em} %
\begin{subfigure}[c]{0.325\linewidth} %
    \raggedleft %
    \includegraphics[scale=0.525]{figs/ablation/#1_4.pdf} %
    \vskip -0.25em %
    \caption{Levy (200D)} %
\end{subfigure} %
%
\begin{subfigure}[c]{0.325\linewidth} %
    \raggedleft %
    \includegraphics[scale=0.525]{figs/ablation/#1_5.pdf} %
    \vskip -0.25em %
    \caption{Landing (12D)} %
\end{subfigure} %
\begin{subfigure}[c]{0.325\linewidth} %
    \raggedleft %
    \includegraphics[scale=0.525]{figs/ablation/#1_6.pdf} %
    \vskip -0.25em %
    \caption{Rover (60D)} %
\end{subfigure} %
%
\hfill \vspace{0.5em} %
\begin{subfigure}[c]{0.325\linewidth} %
    \raggedleft %
    \includegraphics[scale=0.525]{figs/ablation/#1_7.pdf} %
    \vskip -0.25em %
    \caption{Walker (198D)} %
\end{subfigure} %
\begin{subfigure}[c]{0.325\linewidth} %
    \raggedleft %
    \includegraphics[scale=0.525]{figs/ablation/#1_8.pdf} %
    \vskip -0.25em %
    \caption{Ant (404D)} %
\end{subfigure} %
\begin{subfigure}[c]{0.325\linewidth} %
    \centering %
    \includegraphics[scale=1]{figs/ablation/#1_legend.pdf} %
\end{subfigure} %
}

\newcommand{\figFramework}{%
\begin{figure*}[bt]
\begin{center}
\includegraphics[width=0.9\linewidth]{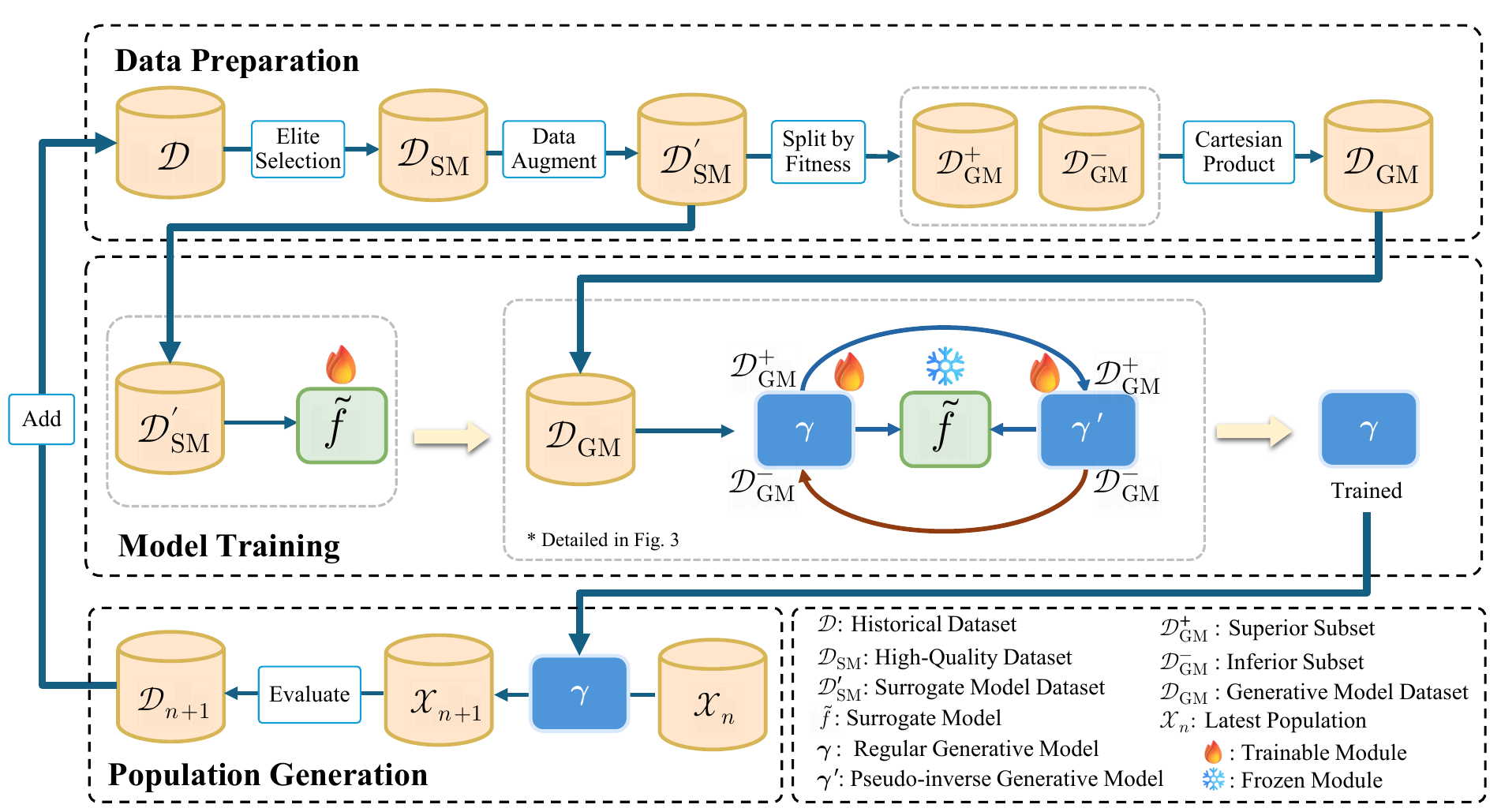}
\caption{ The overall framework of \MethodName. The historical dataset $\mathcal{D}$ is composed of $\{\mathcal{D}_0 \cup \cdots \cup \mathcal{D}_t\}$, where each $\mathcal{D}_t = \left\{ (\bm{x}_{t,i}, y_{t,i}) \mid 1 \le i \le N \right\}$ , with $\bm{x}_{t,i}$ and $y_{t,i}$ denoting the population and its fitness, respectively. In \textbf{data preparation} phase, the historical population is first selected to form the training set of current generation and augmented by an optional learning-based data augmentation strategy. Then, it is separated into the \emph{superior} and \emph{inferior} groups with smaller and larger fitnesses, respectively, before a Cartesian product is applied to generate the paired training dataset. In \textbf{model training} phase, the framework first trains a surrogate model solely for landscape estimation. Then, the surrogate model is frozen and combined with a pair of generative models to form a composite model for effective generative model training. Later, the composite model is trained via the proposed training loss tailored for comoplex optimization. In \textbf{population generation} phase, the inferior-to-superior generative model is used to generate the new solutions in parallel. }
\label{fig:framework}
\end{center}
\end{figure*}
}

\newcommand{\figModel}{%
\begin{figure*}[htb]
    \centering
    \includegraphics[width=0.9\linewidth]{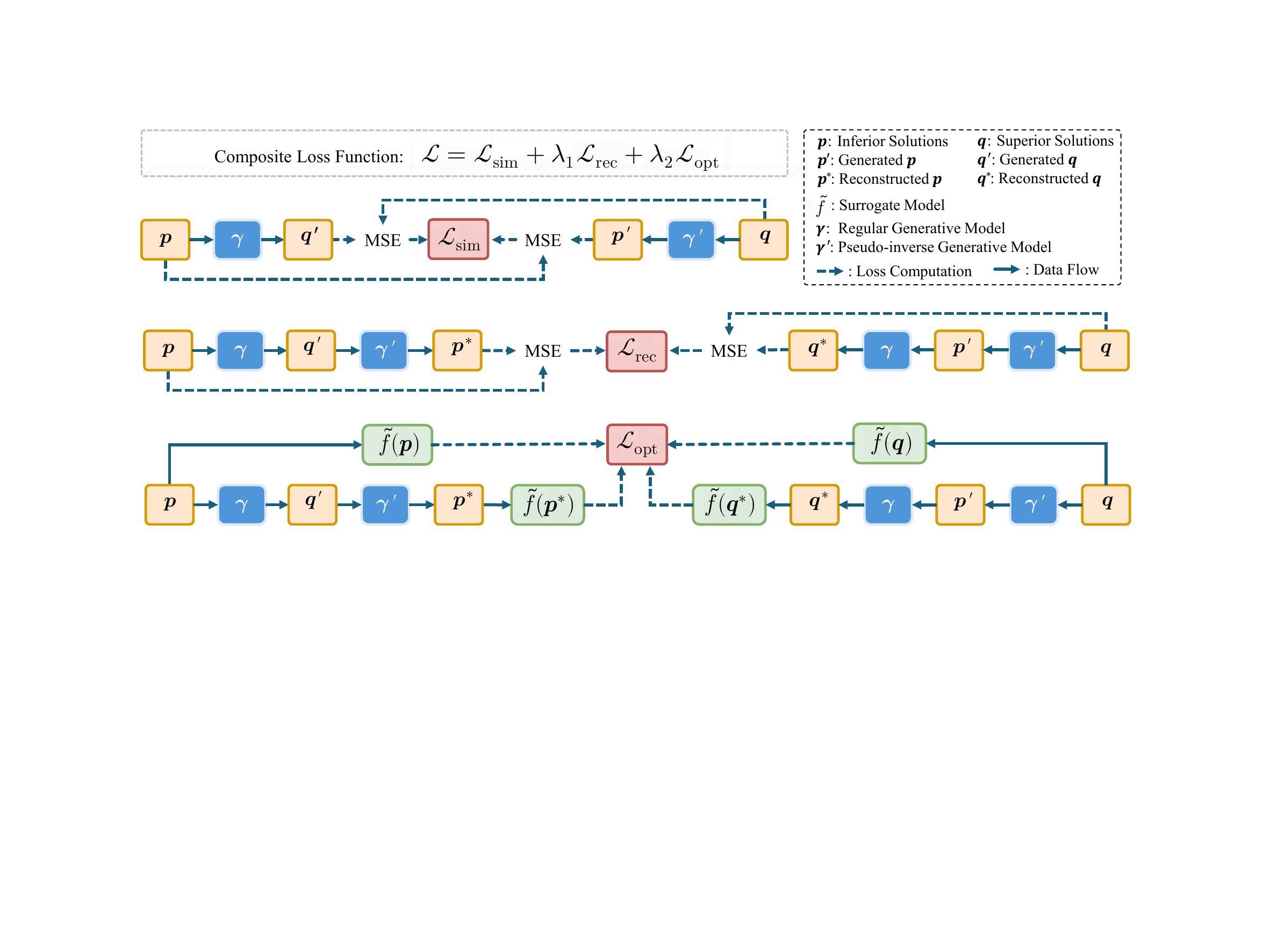}
    \caption{Overview of the model architecture and composite loss functions $\mathcal{L}$ in \MethodName. The top, middle, and bottom rows illustrate the computational workflows for the similarity loss $\mathcal{L}_{\text{sim}}$, the reconstruction loss $\mathcal{L}_{\text{rec}}$, and the optimization loss $\mathcal{L}_{\text{opt}}$, respectively. The input pair $(\bm{p}, \bm{q})$ corresponds to solutions with \emph{inferior} and \emph{superior} fitness, respectively.}
    \label{fig:model}
\end{figure*}
}

\newcommand{\figResultAll}{%
\begin{figure*}[htpb]
    \centering
    \resizebox{0.9\textwidth}{!}{%
    \begin{minipage}{\textwidth}
        \centering

        \begin{subfigure}[c]{0.325\linewidth}
            \centering
            \includegraphics[scale=0.510]{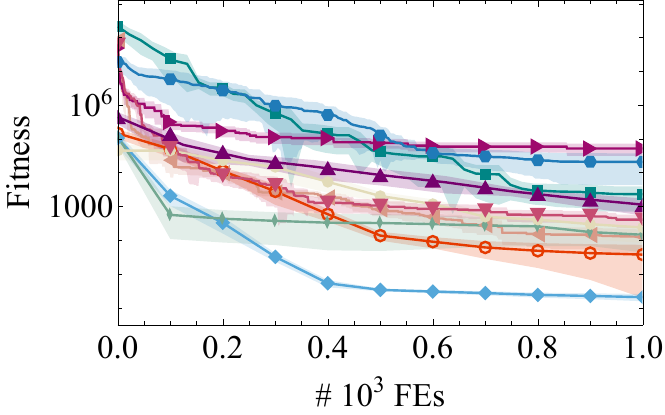}
            \vskip -0.5em
            \caption{Rosenbrock (5)}\label{fig:res-Rosenbrock-5}
        \end{subfigure}
        \begin{subfigure}[c]{0.325\textwidth}
            \centering
            \includegraphics[scale=0.525]{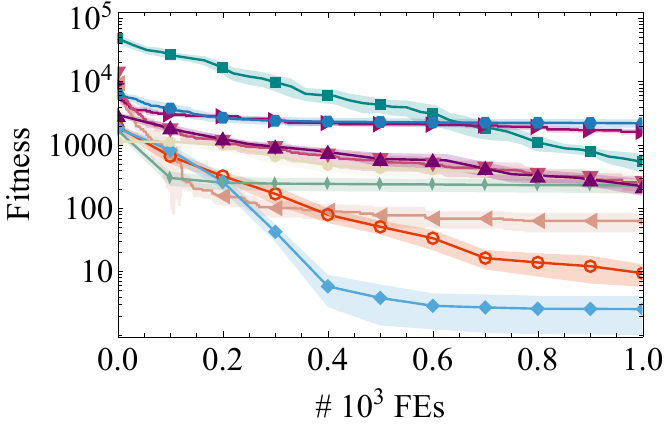}
            \vskip -0.5em
            \caption{Levy (10D)}\label{fig:res-Levy-10}
        \end{subfigure}
        \begin{subfigure}[c]{0.325\linewidth}
            \centering
            \includegraphics[scale=0.525]{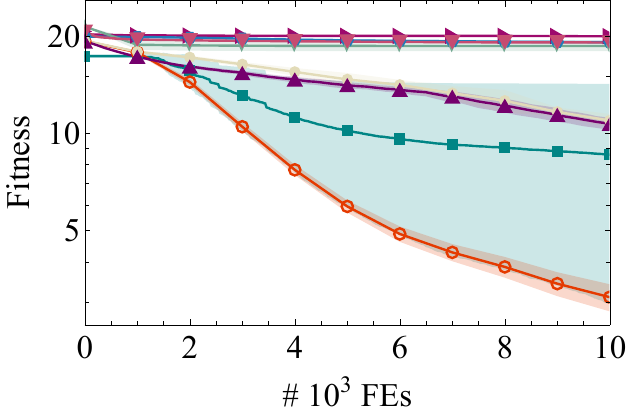}
            \vskip -0.5em
            \caption{Ackley (200D)}\label{fig:res-Ackley-200}
        \end{subfigure}

        \vskip 0.5em

        \begin{subfigure}[c]{0.325\linewidth}
            \centering
            \includegraphics[scale=0.525]{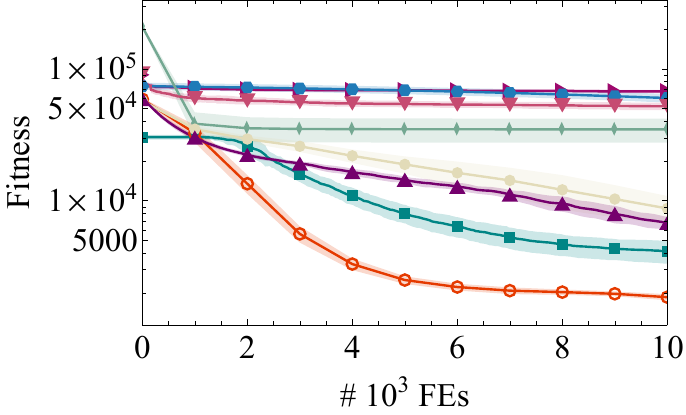}
            \vskip -0.5em
            \caption{Rastrigin (200D)}\label{fig:res-Rastrigin-200}
        \end{subfigure}
        \begin{subfigure}[c]{0.325\linewidth}
            \centering
            \!\includegraphics[scale=0.525]{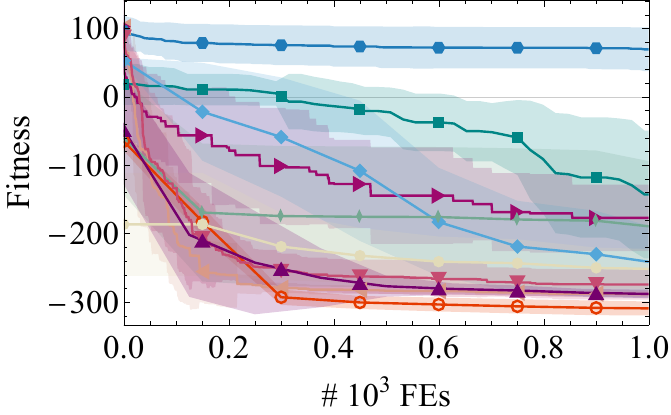}
            \vskip -0.5em
            \caption{Landing (12D)}\label{fig:res-Landing}
        \end{subfigure}
        \begin{subfigure}[c]{0.325\linewidth}
            \centering
            \includegraphics[scale=0.525]{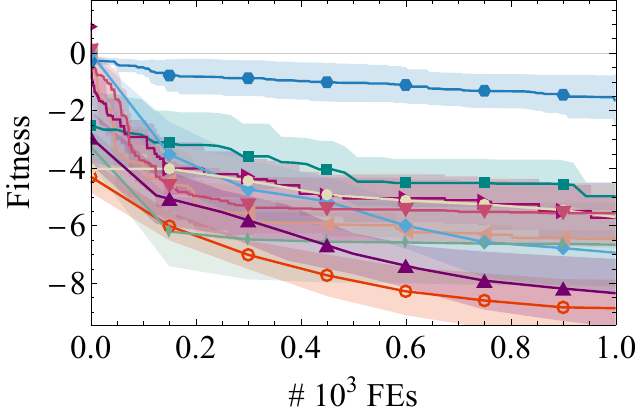}
            \vskip -0.5em
            \caption{Pushing (14D)}\label{fig:res-Pushing}
        \end{subfigure}

        \vskip 0.5em

        \begin{subfigure}[c]{0.325\linewidth}
            \centering
            \hskip -1em
            \includegraphics[scale=0.525]{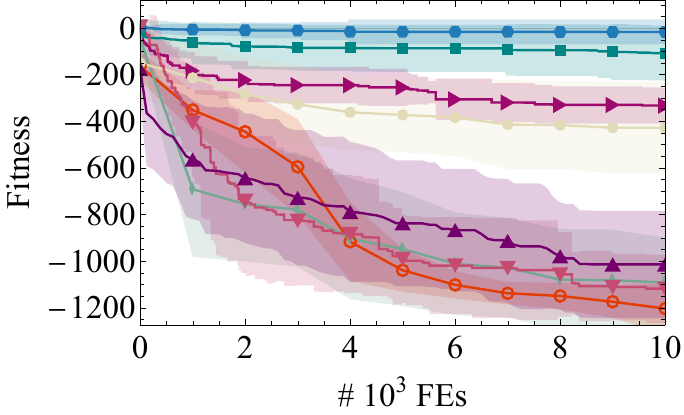}
            \vskip -0.5em
            \caption{Walker (198D)}\label{fig:res-Walker}
        \end{subfigure}
        \begin{subfigure}[c]{0.325\linewidth}
            \centering
            \includegraphics[scale=0.525]{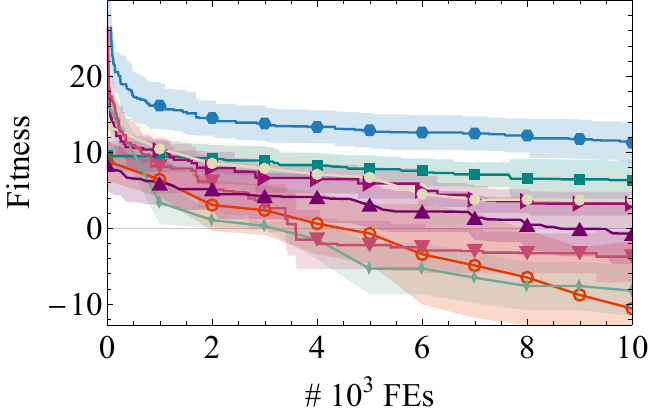}
            \vskip -0.5em
            \caption{Ant (404D)}\label{fig:res-Ant}
        \end{subfigure}
        \begin{subfigure}[c]{0.325\linewidth}
            \centering
            \vskip 1.6em
            \includegraphics[scale=0.9]{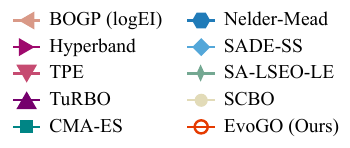}
            \vskip 1.6em
            \caption{Plot legend}\label{fig:res-legend}
        \end{subfigure}

    \end{minipage}
    }

    \caption{The convergence curves of \MethodName and the compared baselines in standard settings. 
    The $y$-axes for numerical functions are in logarithmic scale, while those for other tasks are in linear scale since their true optimal is unknown. 
    One-sigma error bars are adopted since two-sigma ones result in excessively large error bars for some algorithms. 
    Please refer to Supplementary Document E for the complete figure.}
    \label{fig:result-all}
\end{figure*}
}

\newcommand{\figHigh}{%
\begin{figure*}[htb]
    \centering
    \begin{subfigure}[c]{0.325\linewidth}
        \raggedleft
        \includegraphics[scale=0.52]{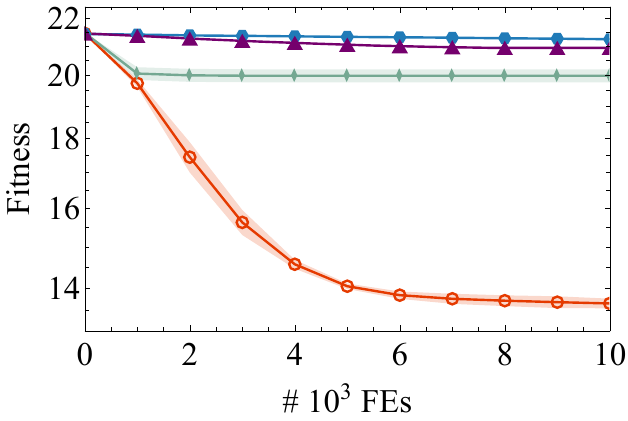}
        \vskip -0.5em
        \caption{Ackley (1000D)}
    \end{subfigure}
    \begin{subfigure}[c]{0.325\linewidth}
        \raggedleft
        \includegraphics[scale=0.52]{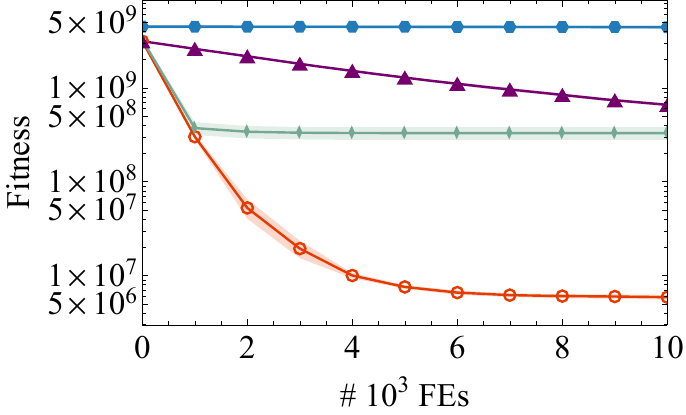}
        \vskip -0.5em
        \caption{Rosenbrock (1000D)}
    \end{subfigure}
    \begin{subfigure}[c]{0.325\linewidth}
        \raggedleft
        \includegraphics[scale=0.52]{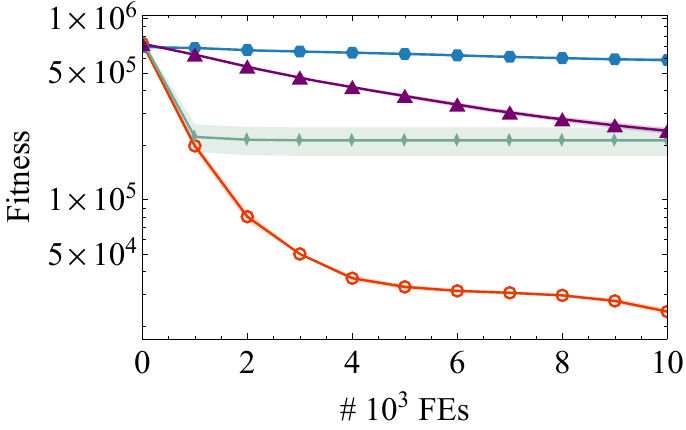}
        \vskip -0.5em
        \caption{Levy (1000D)}
    \end{subfigure}
    \hfill
    \vskip 0.5em
    \begin{subfigure}[c]{0.325\linewidth}
        \raggedleft
        \hskip -1em
        \includegraphics[scale=0.52]{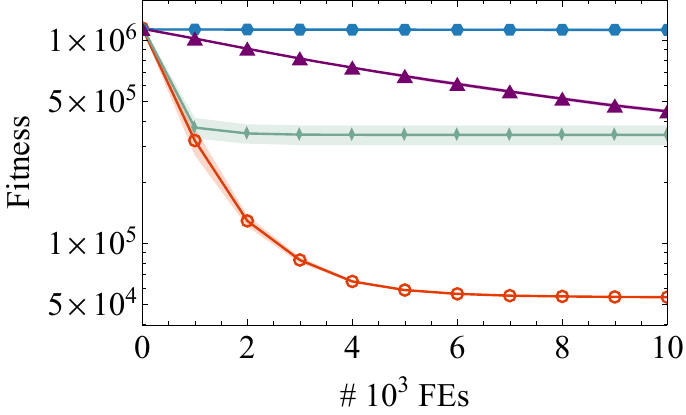}
        \vskip -0.5em
        \caption{Rastrigin (1000D)}
    \end{subfigure}
    \begin{subfigure}[c]{0.325\linewidth}
        \raggedleft
        \vskip 1.6em
        \includegraphics[scale=0.85]{figs/mainexp/res_legend_high.pdf}
        \vskip 1.6em
        \caption{Plot legend}
    \end{subfigure}

    \caption{The convergence curves of \MethodName and representative baseline algorithms on 1000-dimensional numerical problems, under a total evaluation budget of 10,000 FEs. Owing to the substantial computational cost, one representative method is selected from each algorithmic category.}
    \label{fig:high}
\end{figure*}
}

\newcommand{\figSameBatch}{%
\begin{figure*}[ htb]
    \centering
    \begin{subfigure}[c]{0.325\linewidth}
        \centering
        \includegraphics[scale=0.525]{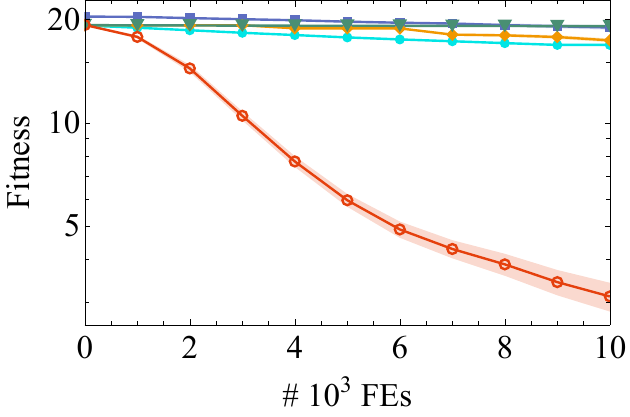}
        \vskip -0.5em
        \caption{Ackley (200D)}
    \end{subfigure}
    \begin{subfigure}[c]{0.325\linewidth}
        \centering
        \includegraphics[scale=0.525]{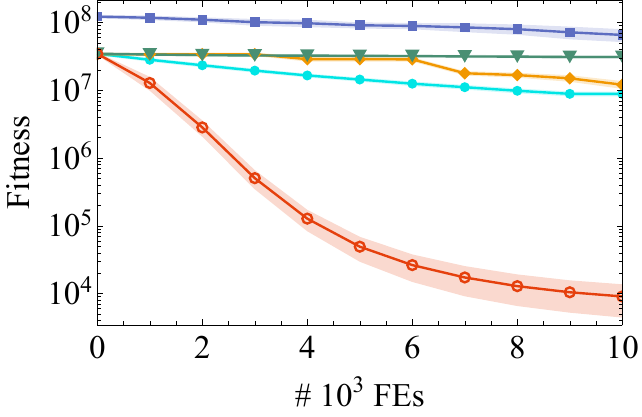}
        \vskip -0.5em
        \caption{Rosenbrock (200D)}
    \end{subfigure}
    \begin{subfigure}[c]{0.325\linewidth}
        \centering
        \!\!\!\includegraphics[scale=0.525]{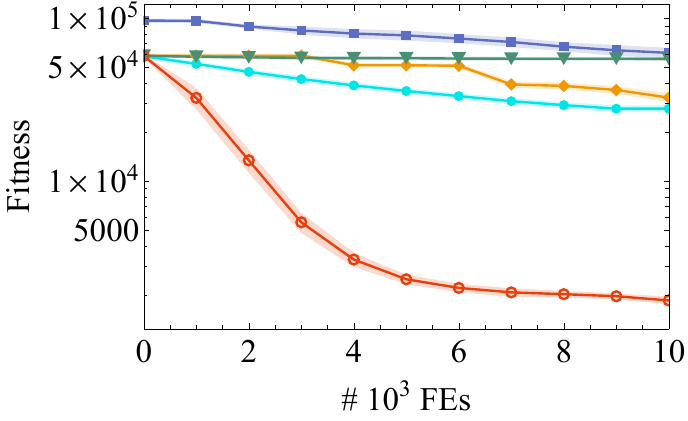}
        \vskip -0.5em
        \caption{Levy (200D)}
    \end{subfigure}

    \vskip 0.25em

    \begin{subfigure}[c]{0.325\linewidth}
        \centering
        \hskip -1em
        \!\!\includegraphics[scale=0.525]{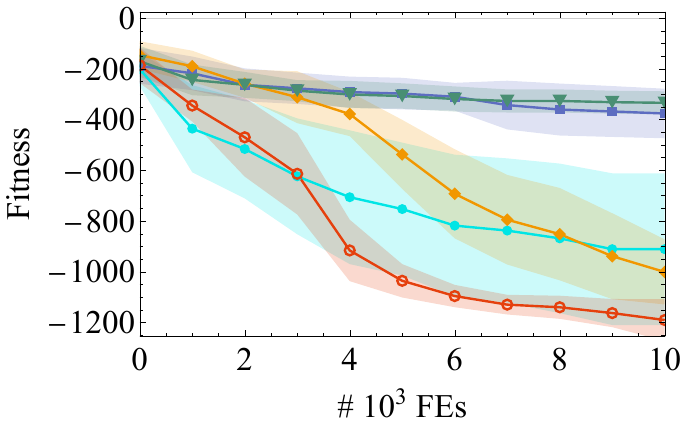}
        \vskip -0.5em
        \caption{Walker (198D)}
    \end{subfigure}
    \begin{subfigure}[c]{0.325\linewidth}
        \centering
        \!\includegraphics[scale=0.525]{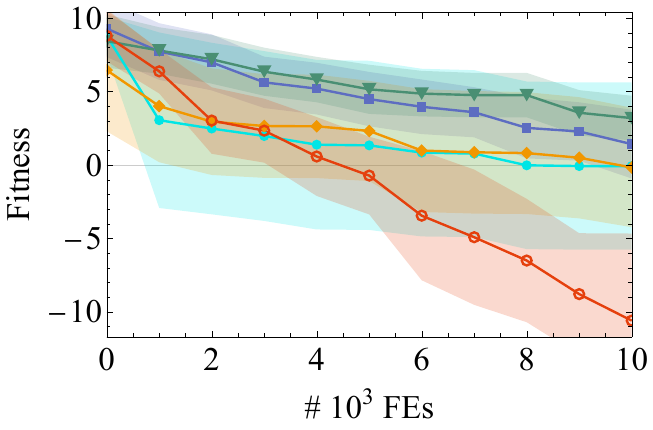}
        \vskip -0.5em
        \caption{Ant (404D)}
    \end{subfigure}
    \begin{subfigure}[c]{0.325\linewidth}
        \centering
        \vskip 1.2em
        \includegraphics[scale=0.9]{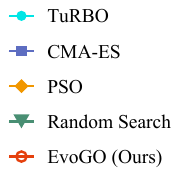}
        \vskip 1.2em
        \caption{Plot Legend}
    \end{subfigure}

    \caption{The convergence curves of \MethodName and the compared baselines (TuRBO, CMA-ES, PSO, and Random Search) in a large-population setting (1000 FEs per generation). All configurations are consistent with those in \cref{fig:result-all}, except that sequential acquisition algorithms are removed since due to unadjustable population sizes. Please refer to Supplementary Document E for complete figure.}
    \label{fig:same-batch}
\end{figure*}
}

\newcommand{\figBrax}{%
\begin{figure}[htb]
    \centering
    \includegraphics[width=0.9\linewidth]{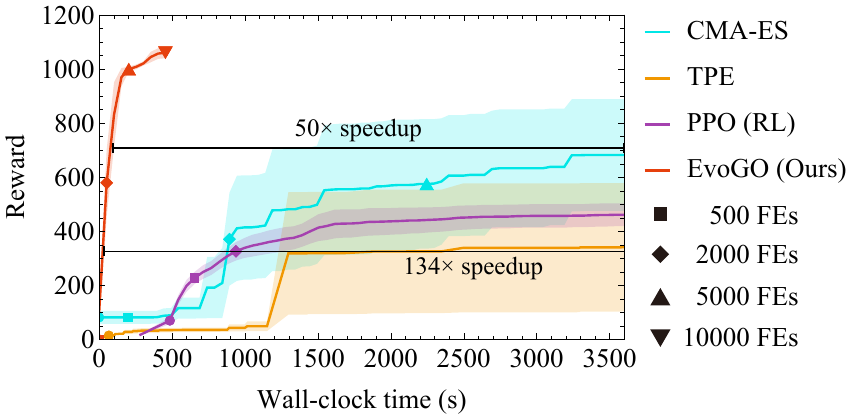}
    \caption{The rewards of TPE, CMA-ES, PPO, and \MethodName on the Hopper environment provided by Brax. To highlight the advantage in runtime of large population sizes, the rewards are plotted versus the wall-clock time, while the FEs are indicated by different markers. We limited the number of FEs and runtime to 10,000 and one hour respectively for all algorithms.}
    \label{fig:brax}
\end{figure}
}

\newcommand{\figDirect}{%
\begin{figure*}[!b]
    \footnotesize
    \centering

    \resizebox{0.9\textwidth}{!}{%
    \begin{minipage}{\textwidth}
        \centering

        \begin{subfigure}[c]{0.325\linewidth}
            \centering
            \!\!\!\includegraphics[scale=0.525]{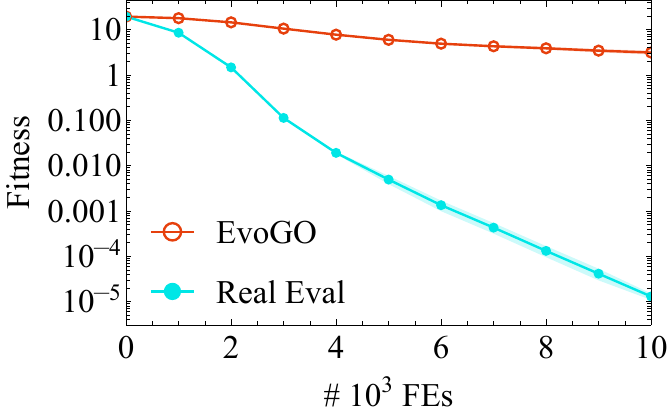}
            \vskip -0.5em
            \caption{Ackley (200D)}
        \end{subfigure}
        \begin{subfigure}[c]{0.325\linewidth}
            \centering
            \includegraphics[scale=0.525]{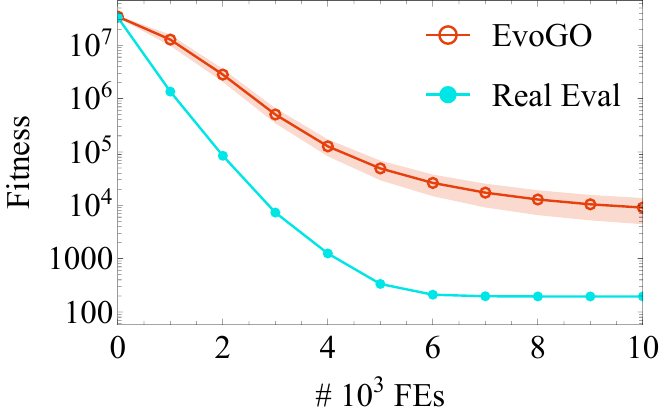}
            \vskip -0.5em
            \caption{Rosenbrock (200D)}
        \end{subfigure}
        \begin{subfigure}[c]{0.325\linewidth}
            \centering
            \includegraphics[scale=0.525]{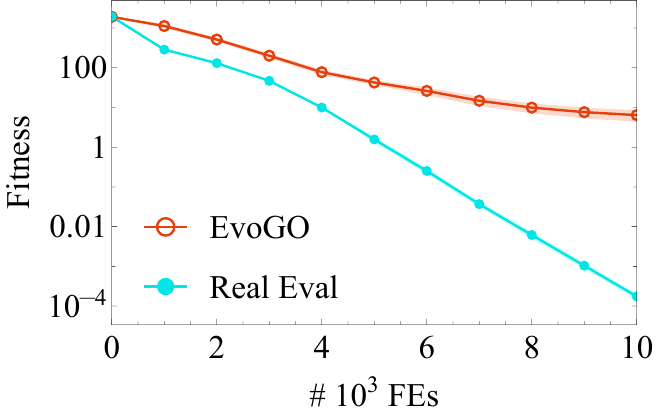}
            \vskip -0.5em
            \caption{Levy (200D)}
        \end{subfigure}

    \end{minipage}
    }

    \caption{The convergence curves of \MethodName and its upper-bound variant ``Real Eval'', 
    which directly utilizes real numerical functions instead of surrogate models.}
    \label{fig:direct}
\end{figure*}
}

\newcommand{\figAblation}{%
\begin{figure*}[!htb]
    \footnotesize
    \centering

    \resizebox{0.9\textwidth}{!}{%
    \begin{minipage}{\textwidth}
        \centering

        \begin{subfigure}[c]{0.325\linewidth}
            \centering
            \includegraphics[scale=0.525]{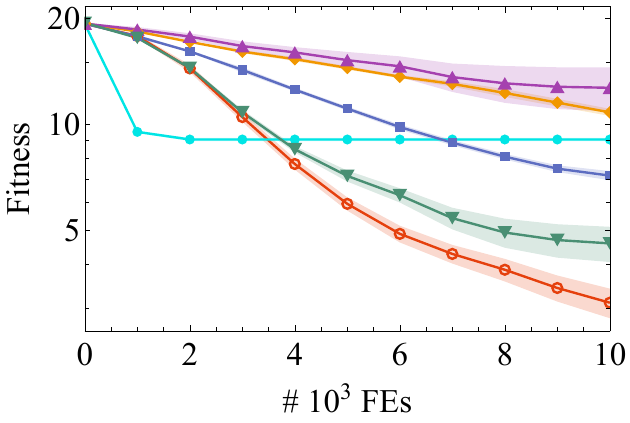}
            \vskip -0.5em
            \caption{Ackley (200D)}
        \end{subfigure}
        \begin{subfigure}[c]{0.325\linewidth}
            \centering
            \includegraphics[scale=0.525]{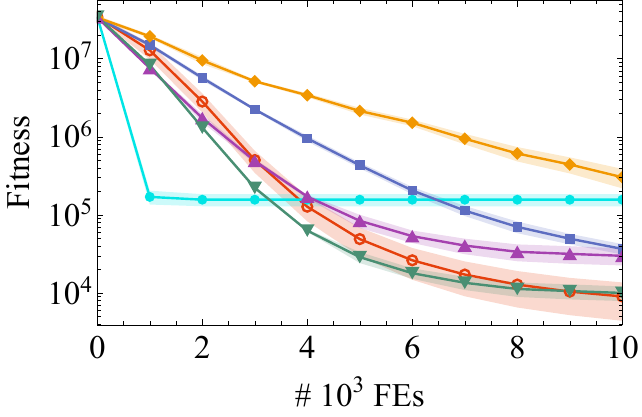}
            \vskip -0.5em
            \caption{Rosenbrock (200D)}
        \end{subfigure}
        \begin{subfigure}[c]{0.325\linewidth}
            \centering
            \includegraphics[scale=0.525]{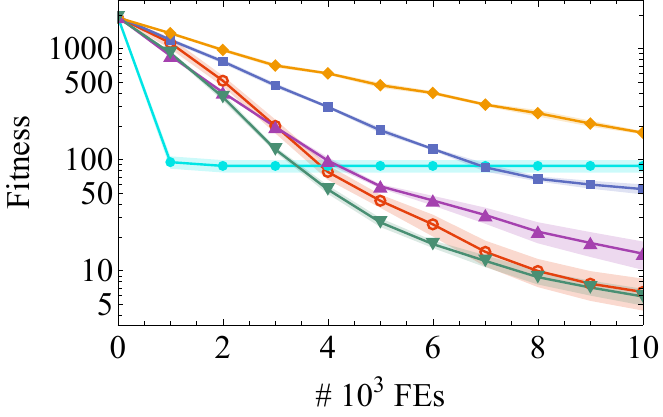}
            \vskip -0.5em
            \caption{Levy (200D)}
        \end{subfigure}

        \vskip 0.25em

        \begin{subfigure}[c]{0.325\linewidth}
            \centering
            \!\!\!\!\!\!\includegraphics[scale=0.525]{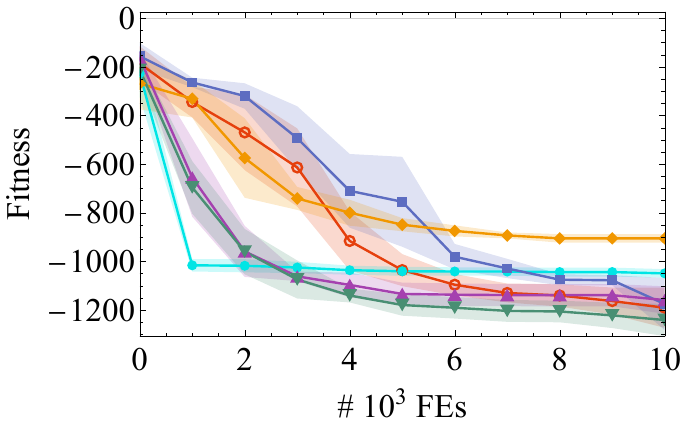}
            \vskip -0.5em
            \caption{Walker (198D)}
        \end{subfigure}
        \begin{subfigure}[c]{0.325\linewidth}
            \centering
            \includegraphics[scale=0.525]{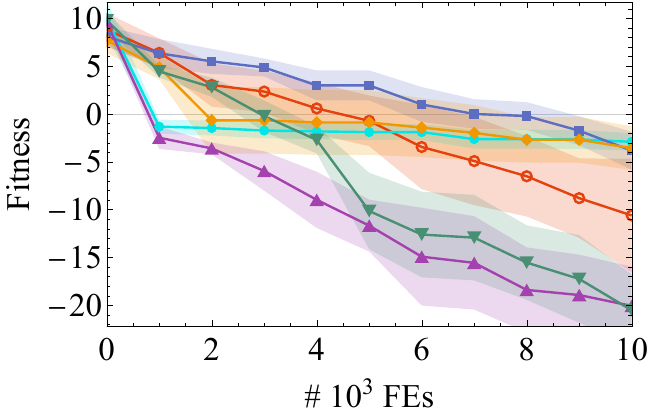}
            \vskip -0.5em
            \caption{Ant (404D)}
        \end{subfigure}
        \begin{subfigure}[c]{0.325\linewidth}
            \centering
            \vskip 1.25em
            \includegraphics[scale=0.9]{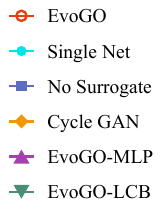}
            \vskip 1.25em
            \caption{Plot Legend}
        \end{subfigure}

    \end{minipage}
    }

    \caption{The convergence curves of \MethodName and its ablations. 
    ``\MethodName'' represents the original method; ``Single Net'' excludes the generative model $\bm{\gamma}$ and related loss terms; 
    ``Cycle GAN'' adapts the Cycle GAN; ``\MethodName-MLP'' utilizes MLP for surrogate modeling; 
    ``\MethodName-LCB'' employs the LCB infill criteria. 
    Please refer to Supplementary Document F for the complete figure.}
    \label{fig:ablation}
\end{figure*}
}

\newcommand{\figVisToy}{%
\begin{figure*}[htpb]
    \small
    \centering

    \resizebox{0.9\textwidth}{!}{%
    \begin{minipage}{\textwidth}
        \centering

        \begin{subfigure}[c]{0.275\linewidth}
            \centering
            \hskip -2em
            Generation 1
        \end{subfigure}
        \begin{subfigure}[c]{0.275\linewidth}
            \centering
            \hskip -2em
            Generation 3
        \end{subfigure}
        \begin{subfigure}[c]{0.275\linewidth}
            \centering
            \hskip -2em
            Generation 5
        \end{subfigure}
        \begin{subfigure}[c]{0.1\linewidth}
            \textcolor{white}{PH}
        \end{subfigure}
        \hfill

        \begin{subfigure}[c]{0.275\linewidth}
            \centering
            \includegraphics[scale=0.63]{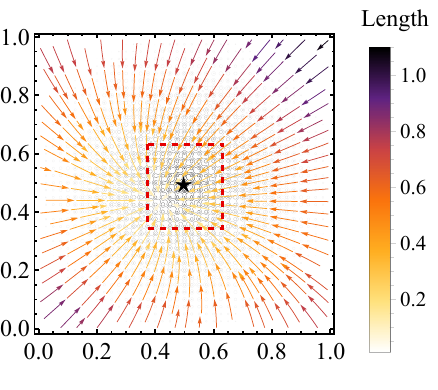}
        \end{subfigure}
        \begin{subfigure}[c]{0.275\linewidth}
            \centering
            \includegraphics[scale=0.63]{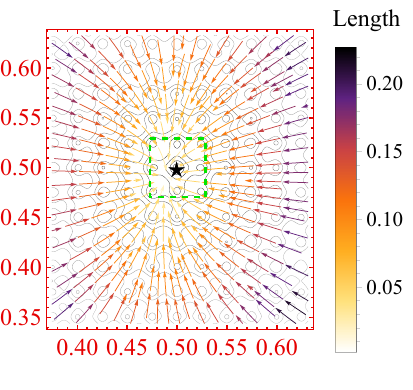}
        \end{subfigure}
        \begin{subfigure}[c]{0.275\linewidth}
            \centering
            \includegraphics[scale=0.63]{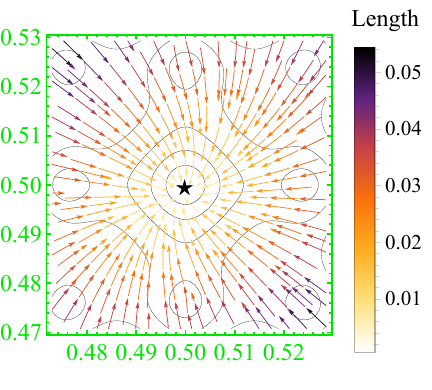}
        \end{subfigure}
        \begin{subfigure}[c]{0.03\linewidth}
            \textcolor{white}{PH}
        \end{subfigure}
        \begin{subfigure}[c]{0.05\linewidth}
            \centering
            \includegraphics[width=\textwidth]{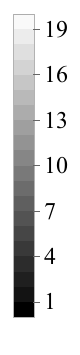}
        \end{subfigure}
        \begin{subfigure}[c]{0.02\linewidth}
            \centering
            \begin{turn}{-90}Fitness\end{turn}
        \end{subfigure}
        \hfill

    \end{minipage}
    }

    \caption{Visual representation of vectors from $\bm{x}_{\text{in}}$ to $\bm{x}_{\text{out}}$ on the Ackley function landscape. 
    The vectors were generated by the trained generative model across different generations. 
    They illustrate the transformation from input solutions $\bm{x}_{\text{in}}$ to the corresponding output solutions $\bm{x}_{\text{out}}$. 
    The central star indicates the global optimum, while the dashed rectangles highlight regions covered by the solutions generated at different generations. 
    For clarity, any landscape shifting and rotation have been omitted from this visualization.}
    \label{fig:vis-toy}
\end{figure*}
}

\newcommand{\algData}{%
\begin{algorithm}[bt]
    \caption{Data Preparation}
    \label{alg:data}
    \begin{algorithmic}[1]
        \raggedright
        \STATE \textbf{Input:} historical data $\mathcal{D}$, dataset split factor $\eta \in (0,1)$, sliding window factor $\epsilon >= 0$, data augmentation threshold $N$, and factor $\rho$
        \STATE \textbf{Output:} $\mathcal{D}_{\text{SM}}$ for surrogate model training, $\mathcal{D}_{\text{GM}}$ for generative model training
        \STATE /************** Elite Selection ***************/
        \STATE $\mathcal{D}' \leftarrow$ select $|\mathcal{D}_0|$ elements from $\mathcal{D} \equiv \mathcal{D}_0 \cup \cdots \cup \mathcal{D}_t$, in descending order of fitnesses
        \STATE $\mathcal{D}_{\text{SM}} \leftarrow \mathcal{D}' \cup \left\{ (\bm{x}, y) \in \mathcal{D}_{t-1} \backslash \mathcal{D}_{\text{all}} \middle| y \leq \max\{y_j^t\} + \epsilon\cdot\mathrm{std}\{y_j^t\} \right\}$
        \STATE /******** Data Augmentation (Optional) ********/
        \IF{$|\mathcal{D}_0| < N$}
        \STATE $\tilde{f} \leftarrow $ trained surrogate model on $\mathcal{D}_{\text{SM}}$
        \STATE $\bm{\mathcal{X}}' \leftarrow$ randomly generate $\rho \cdot |\mathcal{D}'|$ solutions according to the distribution of solutions in $\mathcal{D}'$
        \STATE $\mathcal{D}' \leftarrow \{(\bm{x}, \mu_t(\bm{x}))|\bm{x}\in \bm{\mathcal{X}}\}$ where $\mu_t(\bm{x}) = \mathbb{E}[\tilde{f}(\bm{x})]$
        \ENDIF
        \STATE /*********** Paired Set Construction ***********/
        \STATE $\mathcal{D}_{\text{GM}^+} \leftarrow$ the $\eta$-quantile of $\mathcal{D}'$ based on fitnesses
        \STATE $\mathcal{D}_{\text{GM}^-} \leftarrow $ remove $\mathcal{D}_{\text{GM}^+}$ from $\mathcal{D}_{\text{SM}}$
        \STATE $\mathcal{D}_{\text{GM}} \leftarrow $ Cartesian product of $ \mathcal{D}_{\text{GM}^-} $ and $ \mathcal{D}_{\text{GM}^+}$
    \end{algorithmic}
\end{algorithm}
}

\newcommand{\algTrain}{%
\begin{algorithm}[htpb]
    \caption{Model Training}
    \label{alg:train}
    \begin{algorithmic}[1]
        \STATE \textbf{Input:} maximum training epochs $E$, optimizer \texttt{opt}, training datasets $\mathcal{D}_{\text{SM}}$ and $\mathcal{D}_{\text{GM}}$
        \STATE \textbf{Output:} trained generative model $\bm{\gamma}$
        \STATE $\tilde{f} \leftarrow$ Train the surrogate model using $\mathcal{D}_{\text{SM}}$
        \FOR{$e=1$ \textbf{to} $E$}
        \FOR{\textbf{each pair} $((\bm{x}_-, y_-), (\bm{x}_+, y_+))$ \textbf{in} $\mathcal{D}_{\text{GM}}$}
        \STATE $(\bm{p}, y_{\bm{p}}), (\bm{q}, y_{\bm{q}}) \leftarrow (\bm{x}_-, y_-), (\bm{x}_+, y_+)$
        \STATE $\mathcal{L}_b \leftarrow \mathcal{L}(\bm{\phi}, \bm{\phi}')$ where $\mathcal{L}$ is \cref{eq:loss}
        \STATE $\bm{\phi}, \bm{\phi}' \leftarrow \texttt{opt}(\bm{\phi}, \nabla_{\bm{\phi}} \mathcal{L}_b), \texttt{opt}(\bm{\phi}', \nabla_{\bm{\phi}'} \mathcal{L}_b)$
        \ENDFOR
        \ENDFOR
        \STATE \textbf{Return:} $\bm{\gamma}(\cdot;\bm{\phi}^*)$ where $\bm{\phi}^* \equiv \argmin_{\bm{\phi}}{\mathcal{L}}(\bm{\phi}, \bm{\phi}')$
    \end{algorithmic}
\end{algorithm}
}

\newcommand{\algGenerate}{%
\begin{algorithm}[htpb]
    \caption{Population Generation}
    \label{alg:generate}
    \begin{algorithmic}[1]
        \STATE \textbf{Input:} generative model $\bm{\gamma}$ with optimal parameter $\bm{\phi}^*$, latest population $\mathcal{D}_t$  
        \STATE \textbf{Output:} new population $\mathcal{D}_{t+1}$
        \STATE $\mathcal{D}_{t+1} \leftarrow \emptyset$
        \FOR{\textbf{each} $\left(\bm{x}_{i}^t, y_{i}^t\right) \in \mathcal{D}_t$}
        \STATE $\bm{x}_{i}^{t+1} \leftarrow \bm{\gamma}(\bm{x}_{i}^t;\bm{\phi}^*)$
        \hfill // ``$\bm{\gamma}$'' for trained generative model
        \STATE $y_i^{t+1} \leftarrow f(\bm{x}_{i}^{t+1})$
        \hfill // Function evaluation
        \STATE $\mathcal{D}_{t+1} \leftarrow \mathcal{D}_{t+1} \cup \left\{\left(\bm{x}_{i}^{t+1}, y_i^{t+1}\right) \right\}$
        \ENDFOR
    \end{algorithmic}
\end{algorithm}
}

\newcommand{\tabIters}{%
\begin{table}[hb]
    \caption{The number of generations to complete each benchmark by each method. \textbf{Note:} sequential methods are not listed here since they use exactly the same number of generations as FEs.}
    \label{tab:iters}
    \centering
    \setlength{\tabcolsep}{2.25pt}
    \begin{tabular}{l | c c c c c}
    \hline
    Benchmark        & TuRBO & CMA-ES & PSO & Nelder-Mead & \MethodName \\
    \hline
    Numerical        & 10$\sim$100 & 83$\sim$526  & 25$\sim$50   & 956$\sim$9395 & \textbf{10} \\
    Landing          & \textbf{10} & 91   & 25   & 823$\sim$901 & \textbf{10} \\
    Pushing          & \textbf{10} & 91   & 25   & 862$\sim$933 & \textbf{10} \\
    Rover            & 100 & 625  & 25   & 3163$\sim$3216 & \textbf{10} \\
    Walker           & 100 & 526  & 50$\sim$51 & 1833$\sim$1839 & \textbf{10} \\
    Ant              & 100 & 455  & 24$\sim$25 & 1500$\sim$1503 & \textbf{10} \\
    \hline
    \end{tabular}
\end{table}
}

\newcommand{\tabNumerical}{%
\begin{table}[htb]
    \renewcommand{\arraystretch}{1.5}
    \caption{Formulations of four numerical functions. Here, $d$ denotes the dimensionality and $\bm{s}$ a random shift in the search space.}
    \label{tab:numerical}
    \vskip 0.5em
    \centering
    \begin{tabular}{l p{10.5cm}}
    \toprule
    Function Name & Formulation \\
    \midrule
    Ackley & $f(\bm{x}) = -a \exp\left(-b \sqrt{\sum_{i=1}^{d} \bm{w}_i^2 / d}\right) - \exp\left(\sum_{i=1}^{d} \cos(2\pi \bm{w}_i) / d\right) + a + \exp(1)$, where $a = 20$, $b = 0.2$, and $\bm{w} = 40\bm{x} - 20 - \bm{s}$ \\
    Rosenbrock & $f(\bm{x}) = \sum_{i=1}^{d-1} [100(\bm{w}_{i+1} - \bm{w}_i^2)^2 + (\bm{w}_i - 1)^2]$, where $\bm{w} = 20\bm{x} - 10 - \bm{s}$ \\
    Rastrigin & $f(\bm{x}) = 10d + \sum_{i=1}^{d} [\bm{w}_i^2 - 10 \cos(2\pi \bm{w}_i)]$, where $\bm{w} = 64\bm{x} - 32 - \bm{s}$ \\
    Levy & $f(\bm{x}) = \sin^2(\pi \bm{w}_1) + \sum_{i=1}^{d-1} [(\bm{w}_i - 1)^2 (1 + 10 \sin^2(\pi \bm{w}_i + 1))] + (\bm{w}_d - 1)^2 (1 + \sin^2(2\pi \bm{w}_d))$, where $\bm{w} = 1 + (20\bm{x} - 10 - \bm{s})/{4}$ \\
    \bottomrule
    \end{tabular}
    \renewcommand{\arraystretch}{1}
\end{table}
}

\newcommand{\tabRuntimes}{%
\begin{table}[!htb]
    \caption{The averaged wall-clock runtime of \MethodName and all compared baselines for each problem, exclusive of function evaluation costs.}
    \footnotesize
    \label{tab:runtimes}
    \centering
    \setlength{\tabcolsep}{2.25pt}
    \begin{tabular}{l | c c c c c c c c c}
    \toprule
    Benchmark       & BOGP (logEI) & Hyperband & TPE & TuRBO & GPEME & CMA-ES & PSO & Nelder-Mead & \MethodName \\
    \midrule
    Numerical (5D)  & \SI{8.5}{\minute} & \SI{18}{\second} & \SI{16}{\second} & \SI{17}{\second} & \SI{16}{\second} & \SI{4.3}{\second} & \SI{4.2}{\second} & \SI{10}{\second} & \SI{40}{\second}  \\

    Numerical (10D) & \SI{9.2}{\minute} & \SI{30}{\second} & \SI{30}{\second} & \SI{20}{\second} & \SI{17}{\second} & \SI{4.8}{\second} & \SI{4.5}{\second} & \SI{15}{\second} & \SI{41}{\second} \\

    Numerical (20D) & \SI{11}{\minute} & \SI{35}{\second} & \SI{35}{\second} & \SI{24}{\second} & \SI{19}{\second} & \SI{4.9}{\second} & \SI{4.7}{\second} & \SI{18}{\second} & \SI{40}{\second} \\

    Numerical (200D)& $ > 1\ \mathrm{month}$ & \SI{5.2}{\hour} & \SI{4.0}{\hour} & \SI{35}{\minute} & \SI{44}{\minute} & \SI{5.7}{\minute} & \SI{4.2}{\minute} & \SI{5.5}{\minute} & \SI{6.9}{\minute} \\

    Numerical (1000D)& $ > 1\ \mathrm{month}$ & \SI{5.2}{\hour} & \SI{4.0}{\hour} & \SI{35}{\minute} & \SI{44}{\minute} & \SI{5.7}{\minute} & \SI{4.2}{\minute} & \SI{5.5}{\minute} & \SI{34.4}{\minute} \\

    Landing (12D)   & \SI{10}{\minute} & \SI{42}{\second} & \SI{46}{\second} & \SI{41}{\second} & \SI{33}{\second} & \SI{5.0}{\second} & \SI{4.7}{\second} & \SI{17}{\second} & \SI{41}{\second} \\

    Pushing (14D)   & \SI{10}{\minute} & \SI{47}{\second} & \SI{42}{\second} & \SI{43}{\second} & \SI{33}{\second} & \SI{4.9}{\second} & \SI{4.6}{\second} & \SI{18}{\second} & \SI{42}{\second} \\

    Rover (60D)     & $ > 1\ \mathrm{month}$ & \SI{4.3}{\hour} & \SI{3.1}{\hour} & \SI{19}{\minute} & \SI{38}{\minute} & \SI{8.5}{\minute} & \SI{5.0}{\minute} & \SI{5.6}{\minute} & \SI{7.2}{\minute} \\

    Walker (198D)   & $ > 1\ \mathrm{month}$ & \SI{5.2}{\hour} & \SI{4.0}{\hour} & \SI{33}{\minute} & \SI{42}{\minute} & \SI{5.8}{\minute} & \SI{5.2}{\minute} & \SI{5.5}{\minute} & \SI{7.7}{\minute} \\

    Ant (404D)      & $ > 1\ \mathrm{month}$ & \SI{6.0}{\hour} & \SI{5.1}{\hour} & \SI{42}{\minute} & \SI{44}{\minute} & \SI{4.6}{\minute} & \SI{5.2}{\minute} & \SI{5.6}{\minute} & \SI{23}{\minute} \\
    \bottomrule
    \end{tabular}
\end{table}
}

\newcommand{\tabParams}{%
\begin{table}[!htb]
    \caption{Number of parameters of the generative model (neural network) applied to each problem.}
    \label{tab:params}
    \vskip 0.5em
    \centering
    \renewcommand{\arraystretch}{1.25}
    \begin{tabular}{c | c c c}
    \toprule
    Problem & \MethodName & \MethodName-Wide & \MethodName-Deep \\
    \midrule
    High-dimensional functions (200D) & 1.84M & 2.24M & 2.08M \\
    Low-dimensional functions ($\le$20D) & 169K & 204K & 201K \\
    Rover (60D) & 188K & 228K & 213K \\
    Walker (198D) & 1.81M & 2.20M & 2.04M \\
    Ant (404D) & 7.51M & 9.15M & 8.50M \\
    \bottomrule
    \end{tabular}
    \renewcommand{\arraystretch}{1}
\end{table}
}

\newcommand{\tabEvoGOResults}{%
\renewcommand{\thetable}{S.\Roman{table}}
\begin{table*}[t]
    \centering
    \caption{Statistical Results (Mean and Standard Deviation) of the Tested Algorithms on 1000-Dimensional Benchmark Problems. 
    Each Algorithm Was Executed Ten Times, and the Best Results Are Highlighted in Bold.}
    \label{tab:evogo_1000d}
    \vskip 0.5em
    \renewcommand{\arraystretch}{1.2}

    \resizebox{0.9\textwidth}{!}{%
    \begin{tabular}{lccccc}
        \toprule
        Problem & CMA-ES & Nelder-Mead & TuRBO & SA\_LSEO\_LE & \textbf{EvoGO} \\
        \midrule
        Ackley     
        & 2.1192e+01 (2.66e-02) 
        & 2.1228e+01 (8.64e-03) 
        & 2.1236e+01 (2.80e-02) 
        & 2.0169e+01 (1.37e-01) 
        & \textbf{1.3652e+01 (1.10e-01)} \\
        Rosenbrock 
        & 1.4956e+09 (1.55e+08) 
        & 4.4097e+09 (1.73e+08) 
        & 5.9143e+08 (1.93e+07) 
        & 3.6834e+07 (1.32e+07) 
        & \textbf{5.8708e+06 (3.50e+05)} \\
        Rastrigin  
        & 9.2663e+05 (2.17e+04) 
        & 1.1251e+06 (1.73e+04) 
        & 4.2006e+05 (8.88e+03) 
        & 7.6479e+04 (7.78e+03) 
        & \textbf{5.4188e+04 (5.74e+02)} \\
        Levy       
        & 5.4255e+05 (3.66e+04) 
        & 5.8105e+05 (1.34e+04) 
        & 2.2472e+05 (7.57e+03) 
        & 1.6155e+05 (1.35e+04) 
        & \textbf{2.4058e+04 (8.77e+02)} \\
        \bottomrule
    \end{tabular}
    }

    \renewcommand{\arraystretch}{1}
\end{table*}
}


\newcommand{\figResultAllFullA}{%
\begin{figure}[!th]
    \centering
    \begin{subfigure}[c]{0.325\linewidth}
        \raggedleft
        \includegraphics[scale=0.52]{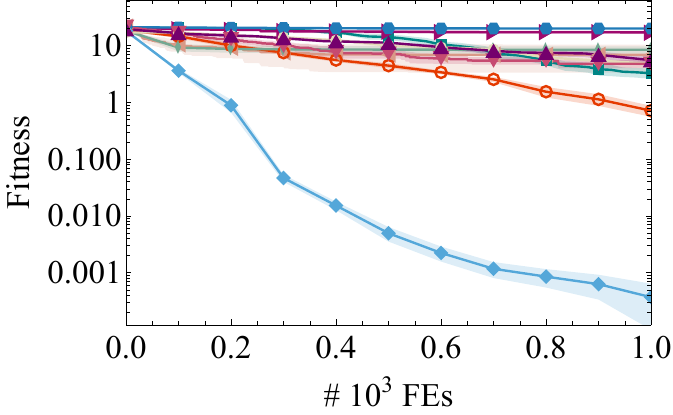}
        \vskip -0.5em
        \caption{Ackley (5D)}
    \end{subfigure}
    \begin{subfigure}[c]{0.325\textwidth}
        \raggedleft
        \includegraphics[scale=0.52]{figs/mainexp/res_13.pdf}
        \vskip -0.5em
        \caption{Rosenbrock (5D)}
    \end{subfigure}
    \begin{subfigure}[c]{0.325\linewidth}
        \raggedleft
        \includegraphics[scale=0.52]{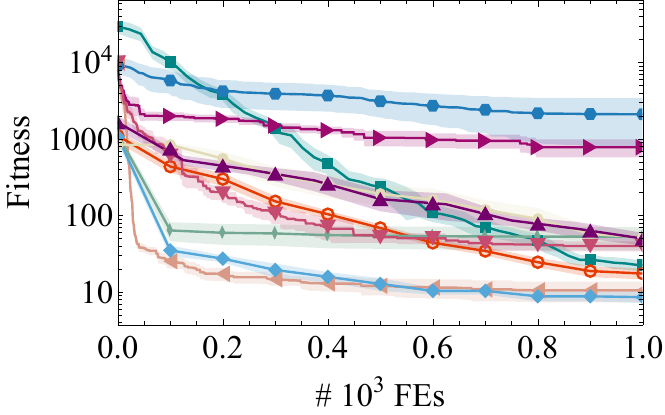}
        \vskip -0.5em
        \caption{Rastrigin (5D)}
    \end{subfigure}
    \hfill
    \vskip 0.5em
    \begin{subfigure}[c]{0.325\linewidth}
        \raggedleft
        \includegraphics[scale=0.52]{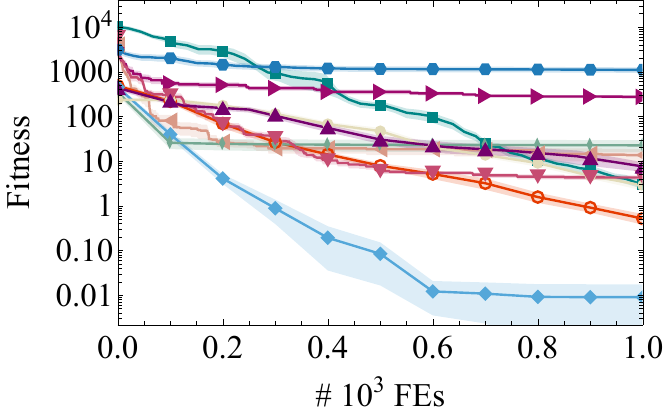}
        \vskip -0.5em
        \caption{Levy (5D)}
    \end{subfigure}
    \begin{subfigure}[c]{0.325\linewidth}
        \raggedleft
        \includegraphics[scale=0.52]{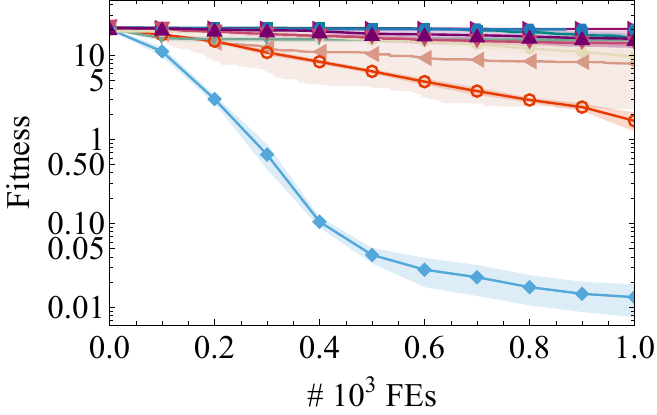}
        \vskip -0.5em
        \caption{Ackley (10D)}
    \end{subfigure}
    \begin{subfigure}[c]{0.325\textwidth}
        \raggedleft
        \includegraphics[scale=0.52]{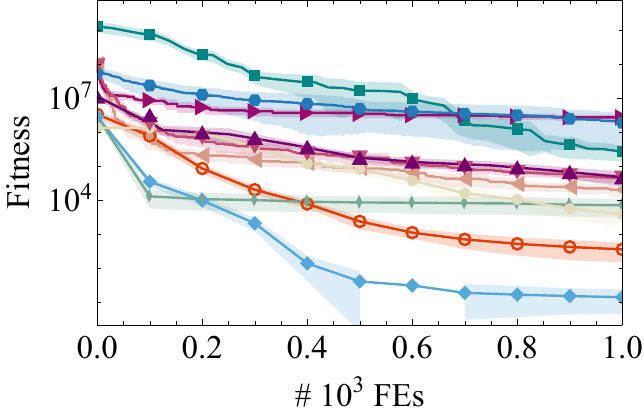}
        \vskip -0.5em
        \caption{Rosenbrock (10D)}
    \end{subfigure}
    \hfill
    \vskip 0.5em
    \begin{subfigure}[c]{0.325\linewidth}
        \raggedleft
        \includegraphics[scale=0.52]{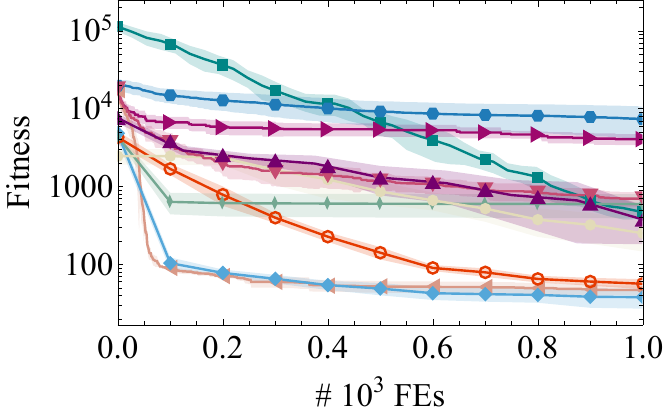}
        \vskip -0.5em
        \caption{Rastrigin (10D)}
    \end{subfigure}
    \begin{subfigure}[c]{0.325\linewidth}
        \raggedleft
        \includegraphics[scale=0.52]{figs/mainexp/res_19.pdf}
        \vskip -0.5em
        \caption{Levy (10D)}
    \end{subfigure}
    \begin{subfigure}[c]{0.325\linewidth}
        \raggedleft
        \includegraphics[scale=0.52]{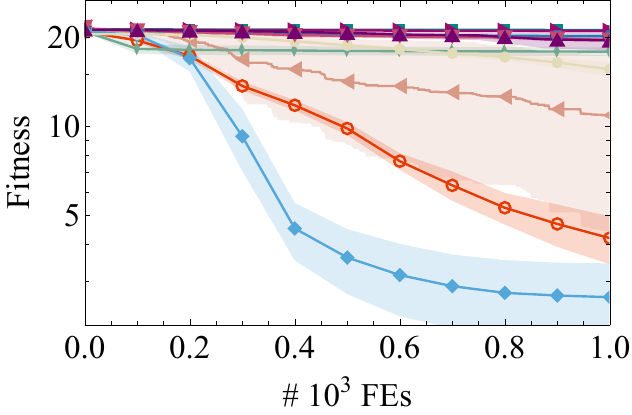}
        \vskip -0.5em
        \caption{Ackley (20D)}
    \end{subfigure}
    \hfill
    \vskip 0.5em
    \begin{subfigure}[c]{0.325\textwidth}
        \raggedleft
        \includegraphics[scale=0.52]{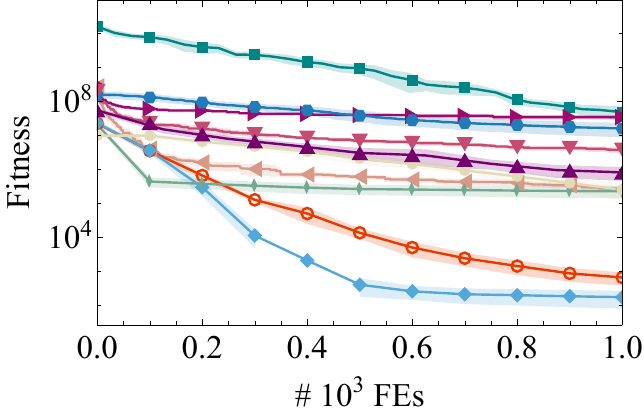}
        \vskip -0.5em
        \caption{Rosenbrock (20D)}
    \end{subfigure}
    \begin{subfigure}[c]{0.325\linewidth}
        \raggedleft
        \includegraphics[scale=0.52]{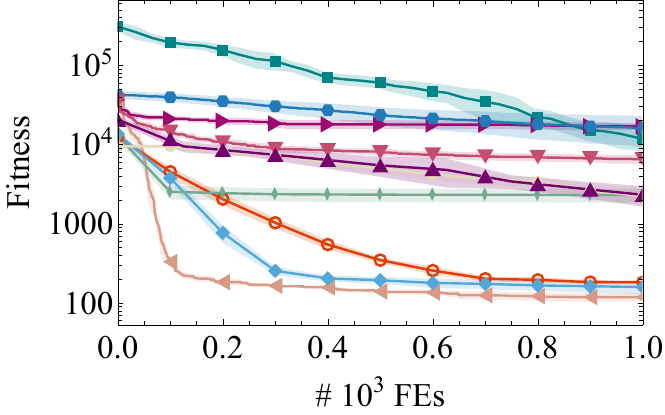}
        \vskip -0.5em
        \caption{Rastrigin (20D)}
    \end{subfigure}
    \begin{subfigure}[c]{0.325\linewidth}
        \raggedleft
        \vskip 1.6em
        \includegraphics[scale=0.85]{figs/mainexp/res_legend.pdf}
        \vskip 1.6em
        \caption{Plot legend}
    \end{subfigure}
    \caption{The convergence curves of \MethodName and the compared baselines in standard settings. The $y$-axes for numerical functions are in logarithmic scale, while those for other tasks are in linear scale since their true optimal is unknown. One-sigma error bars are adopted since two-sigma ones result in excessively large error bars for some algorithms.}
    \label{fig:result-all-full}
\end{figure}
}

\newcommand{\figResultAllFullB}{%
\begin{figure}[!th]
    \ContinuedFloat
    \begin{subfigure}[c]{0.325\linewidth}
        \raggedleft
        \hskip -0.85em \includegraphics[scale=0.52]{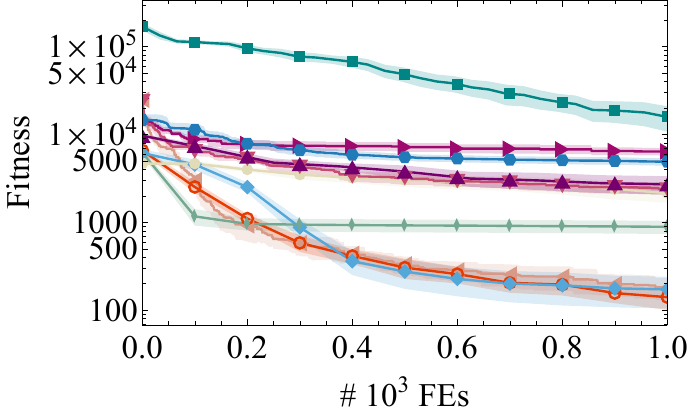}
        \vskip -0.5em
        \caption{Levy (20D)}
    \end{subfigure}
    \begin{subfigure}[c]{0.325\linewidth}
        \raggedleft
        \includegraphics[scale=0.52]{figs/mainexp/res_25.pdf}
        \vskip -0.5em
        \caption{Landing (12D)}
    \end{subfigure}
    \begin{subfigure}[c]{0.325\linewidth}
        \raggedleft
        \includegraphics[scale=0.52]{figs/mainexp/res_24.pdf}
        \vskip -0.5em
        \caption{Pushing (14D)}
    \end{subfigure}
    \hfill
    \vskip 0.5em
    \begin{subfigure}[c]{0.325\linewidth}
        \raggedleft
        \includegraphics[scale=0.52]{figs/mainexp/res_26.pdf}
        \vskip -0.5em
        \caption{Ackley (200D)}
    \end{subfigure}
    \begin{subfigure}[c]{0.325\linewidth}
        \raggedleft
        \includegraphics[scale=0.52]{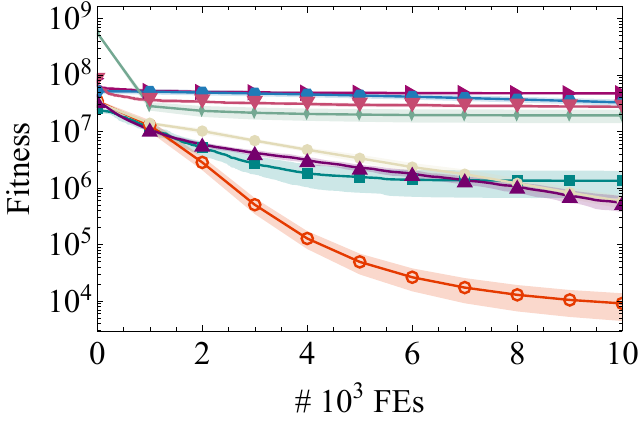}
        \vskip -0.5em
        \caption{Rosenbrock (200D)}
    \end{subfigure}
    \begin{subfigure}[c]{0.325\linewidth}
        \raggedleft
        \includegraphics[scale=0.52]{figs/mainexp/res_28.pdf}
        \vskip -0.5em
        \caption{Rastrigin (200D)}
    \end{subfigure}
    \hfill
    \vskip 0.5em
    \begin{subfigure}[c]{0.325\linewidth}
        \raggedleft
        \includegraphics[scale=0.52]{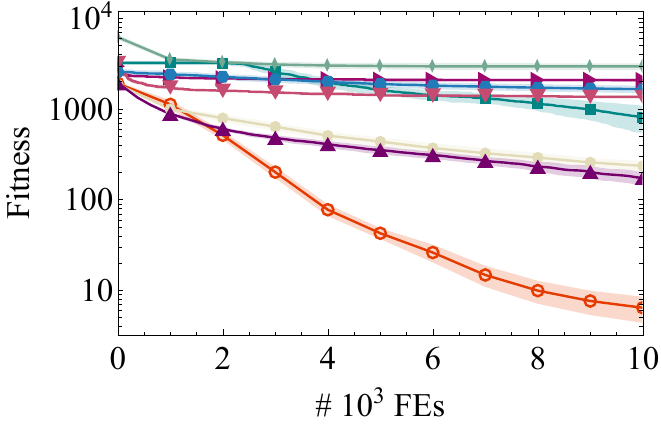}
        \vskip -0.5em
        \caption{Levy (200D)}
    \end{subfigure}
    \begin{subfigure}[c]{0.325\linewidth}
        \raggedleft
        \includegraphics[scale=0.52]{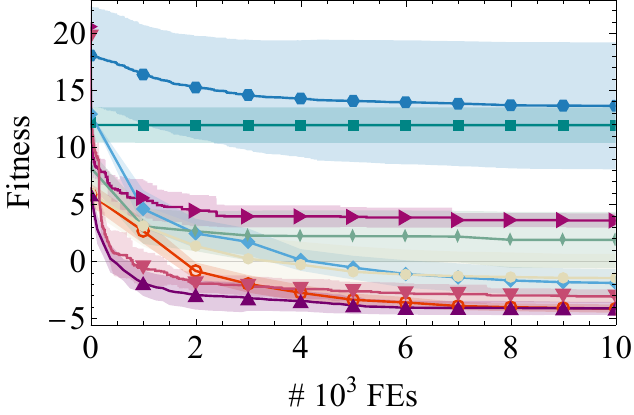}
        \vskip -0.5em
        \caption{Rover (60D)}
    \end{subfigure}
    \begin{subfigure}[c]{0.325\linewidth}
        \raggedleft
        \hskip -1em
        \includegraphics[scale=0.52]{figs/mainexp/res_31.pdf}
        \vskip -0.5em
        \caption{Walker (198D)}
    \end{subfigure}
    \hfill
    \vskip 0.5em
    \begin{subfigure}[c]{0.325\linewidth}
        \raggedleft
        \includegraphics[scale=0.52]{figs/mainexp/res_32.pdf}
        \vskip -0.5em
        \caption{Ant (404D)}
    \end{subfigure}
    \begin{subfigure}[c]{0.325\linewidth}
        \raggedleft
        \vskip 1.6em
        \includegraphics[scale=0.85]{figs/mainexp/res_legend.pdf}
        \vskip 1.6em
        \caption{Plot legend}
    \end{subfigure}
    \caption{(Continued) The convergence curves of \MethodName and the compared baselines in standard settings. The $y$-axes for numerical functions are in logarithmic scale, while those for other tasks are in linear scale since their true optimal is unknown. One-sigma error bars are adopted since two-sigma ones result in excessively large error bars for some algorithms.}
\end{figure}
}

\newcommand{\figSameBatchFull}{%
\begin{figure}[!thb]
    \centering
    \begin{subfigure}[c]{0.325\linewidth}
        \raggedleft
        \includegraphics[scale=0.525]{figs/additional/res_same_batch_1.pdf}
        \vskip -0.5em
        \caption{Ackley (200D)}
    \end{subfigure}
    \begin{subfigure}[c]{0.325\textwidth}
        \raggedleft
        \includegraphics[scale=0.525]{figs/additional/res_same_batch_2.pdf}
        \vskip -0.5em
        \caption{Rosenbrock (200D)}
    \end{subfigure}
    \begin{subfigure}[c]{0.325\linewidth}
        \raggedleft
        \includegraphics[scale=0.525]{figs/additional/res_same_batch_3.pdf}
        \vskip -0.5em
        \caption{Rastrigin (200D)}
    \end{subfigure}
    \hfill
    \vskip 0.5em
    \begin{subfigure}[c]{0.325\linewidth}
        \raggedleft
        \includegraphics[scale=0.525]{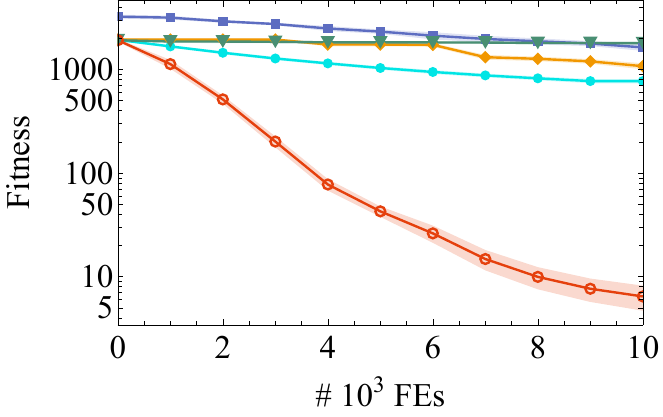}
        \vskip -0.5em
        \caption{Levy (200D)}
    \end{subfigure}
    \begin{subfigure}[c]{0.325\linewidth}
        \raggedleft
        \includegraphics[scale=0.525]{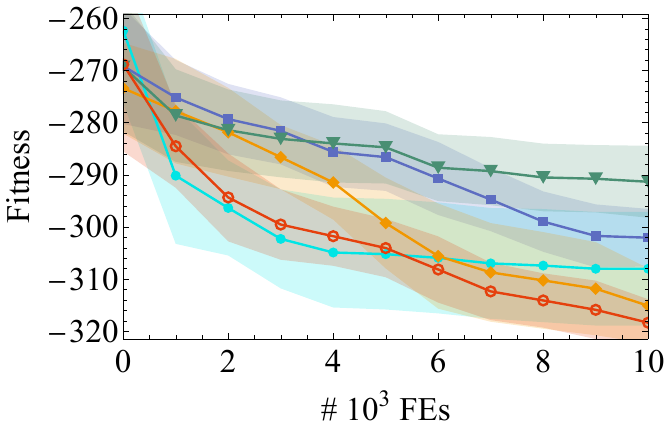}
        \vskip -0.5em
        \caption{Landing (12D)}
    \end{subfigure}
    \begin{subfigure}[c]{0.325\linewidth}
        \raggedleft
        \includegraphics[scale=0.525]{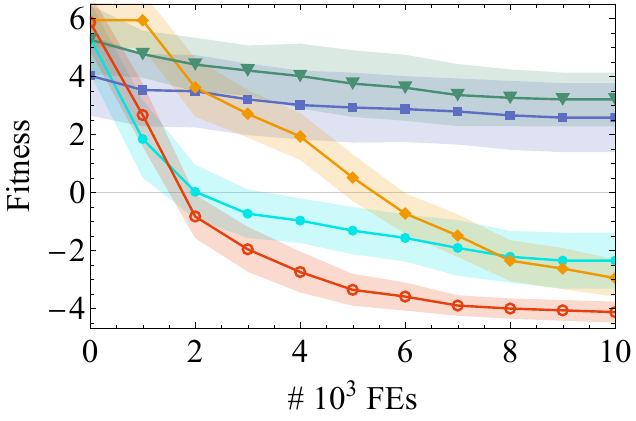}
        \vskip -0.5em
        \caption{Rover (60D)}
    \end{subfigure}
    \hfill
    \vskip 0.5em
    \begin{subfigure}[c]{0.325\textwidth}
        \raggedleft
        \includegraphics[scale=0.525]{figs/additional/res_same_batch_7.pdf}
        \vskip -0.5em
        \caption{Walker (198D)}
    \end{subfigure}
    \begin{subfigure}[c]{0.325\textwidth}
        \raggedleft
        \includegraphics[scale=0.525]{figs/additional/res_same_batch_8.pdf}
        \vskip -0.5em
        \caption{Ant (404D)}
    \end{subfigure}
    \begin{subfigure}[c]{0.325\linewidth}
        \centering
        \vskip 1.6em
        \includegraphics[scale=0.85]{figs/additional/res_same_batch_legend.pdf}
        \vskip 1.6em
        \caption{Plot legend}
    \end{subfigure}

    \caption{The convergence curves of \MethodName and the compared baselines (TuRBO, CMA-ES, PSO, and Random Search) in a large-batch setting (1000 FEs per generation). All configurations are consistent with those in Fig. 4, except that sequential acquisition algorithms are removed since their batch sizes / population sizes cannot be adjusted.}
    \label{fig:same-batch-full}
\end{figure}
}

\newcommand{\figAblationFull}{%
\begin{figure}[!thb]
    \centering
    \begin{subfigure}[c]{0.325\linewidth}
        \raggedleft
        \includegraphics[scale=0.525]{figs/ablation/ablation_1.pdf}
        \vskip -0.5em
        \caption{Ackley (200D)}
    \end{subfigure}
    \begin{subfigure}[c]{0.325\textwidth}
        \raggedleft
        \includegraphics[scale=0.525]{figs/ablation/ablation_2.pdf}
        \vskip -0.5em
        \caption{Rosenbrock (200D)}
    \end{subfigure}
    \begin{subfigure}[c]{0.325\linewidth}
        \raggedleft
        \includegraphics[scale=0.525]{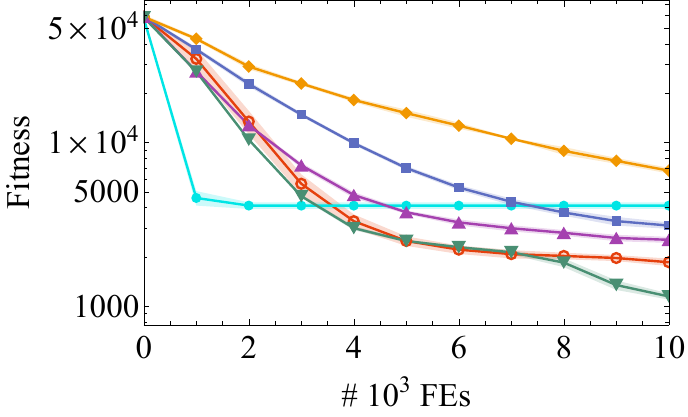}
        \vskip -0.5em
        \caption{Rastrigin (200D)}
    \end{subfigure}
    \hfill
    \vskip 0.5em
    \begin{subfigure}[c]{0.325\linewidth}
        \raggedleft
        \includegraphics[scale=0.525]{figs/ablation/ablation_4.pdf}
        \vskip -0.5em
        \caption{Levy (200D)}
    \end{subfigure}
    \begin{subfigure}[c]{0.325\linewidth}
        \raggedleft
        \includegraphics[scale=0.525]{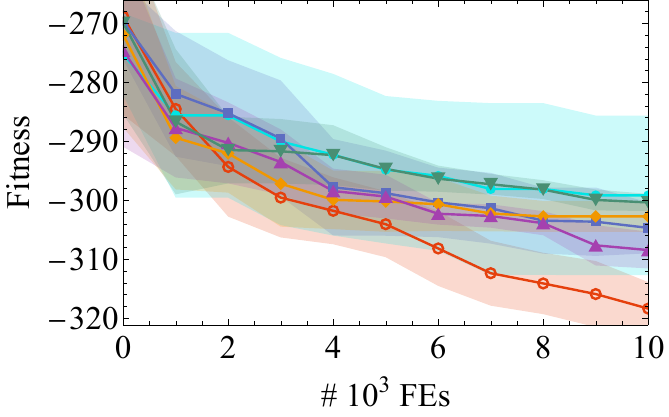}
        \vskip -0.5em
        \caption{Landing (12D)}
    \end{subfigure}
    \begin{subfigure}[c]{0.325\linewidth}
        \raggedleft
        \includegraphics[scale=0.525]{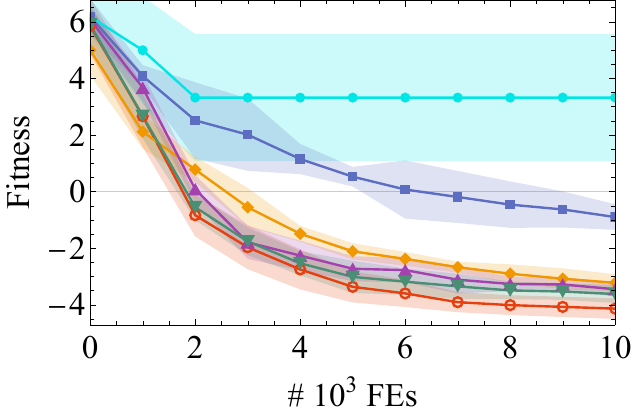}
        \vskip -0.5em
        \caption{Rover (60D)}
    \end{subfigure}
    \hfill
    \vskip 0.5em
    \begin{subfigure}[c]{0.325\textwidth}
        \raggedleft
        \includegraphics[scale=0.525]{figs/ablation/ablation_7.pdf}
        \vskip -0.5em
        \caption{Walker (198D)}
    \end{subfigure}
    \begin{subfigure}[c]{0.325\textwidth}
        \raggedleft
        \includegraphics[scale=0.525]{figs/ablation/ablation_8.pdf}
        \vskip -0.5em
        \caption{Ant (404D)}
    \end{subfigure}
    \begin{subfigure}[c]{0.325\linewidth}
        \centering
        \vskip 1.2em
        \includegraphics[scale=0.9]{figs/ablation/ablation_legend.pdf}
        \vskip 1.2em
        \caption{Plot legend}
    \end{subfigure}
    \caption{Performance comparison between \MethodName and its ablation variants. ``\MethodName'' represents the original method; ``Single Net'' excludes the generative model $\bm{\gamma}'$ and related loss terms; ``Cycle GAN'' adapts the Cycle GAN model for optimization; ``\MethodName-MLP'' utilizes MLP for surrogate modeling; ``\MethodName-LCB'' utilizes the LCB optimization loss instead of the proposed composite optimization loss.}
    \label{fig:ablation-full}
\end{figure}
}





\newcommand{\figNnArchAll}{%
\begin{figure}[!thb]
    \centering
    \begin{subfigure}[c]{0.325\linewidth}
        \raggedleft
        \includegraphics[scale=0.525]{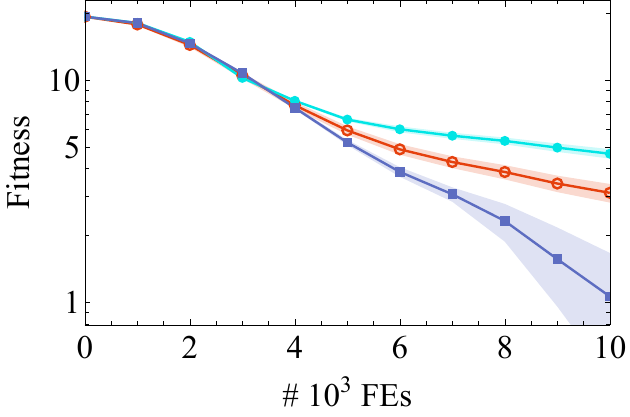}
        \vskip -0.5em
        \caption{Ackley (200D)}
    \end{subfigure}
    \begin{subfigure}[c]{0.325\textwidth}
        \raggedleft
        \includegraphics[scale=0.525]{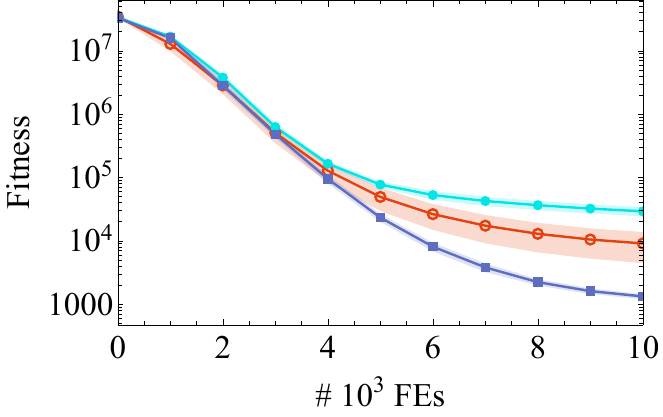}
        \vskip -0.5em
        \caption{Rosenbrock (200D)}
    \end{subfigure}
    \begin{subfigure}[c]{0.325\linewidth}
        \raggedleft
        \includegraphics[scale=0.525]{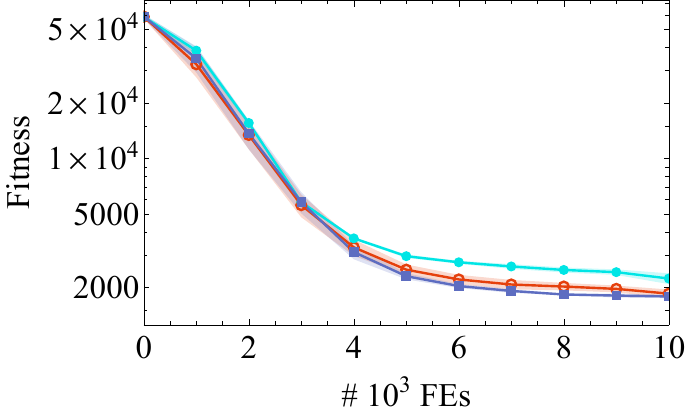}
        \vskip -0.5em
        \caption{Rastrigin (200D)}
    \end{subfigure}
    \hfill
    \vskip 0.5em
    \begin{subfigure}[c]{0.325\linewidth}
        \raggedleft
        \includegraphics[scale=0.525]{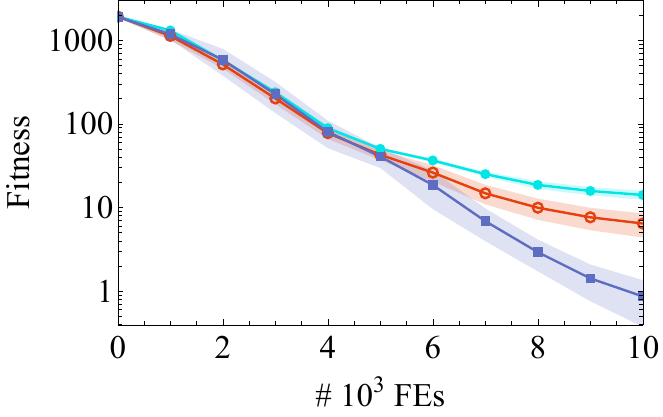}
        \vskip -0.5em
        \caption{Levy (200D)}
    \end{subfigure}
    \begin{subfigure}[c]{0.325\linewidth}
        \raggedleft
        \includegraphics[scale=0.525]{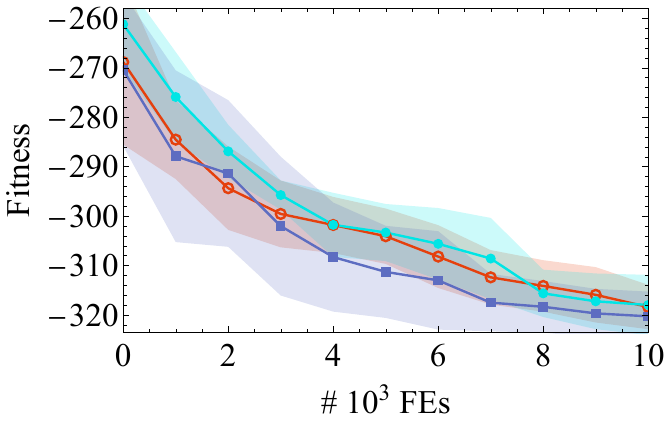}
        \vskip -0.5em
        \caption{Landing (12D)}
    \end{subfigure}
    \begin{subfigure}[c]{0.325\linewidth}
        \raggedleft
        \includegraphics[scale=0.525]{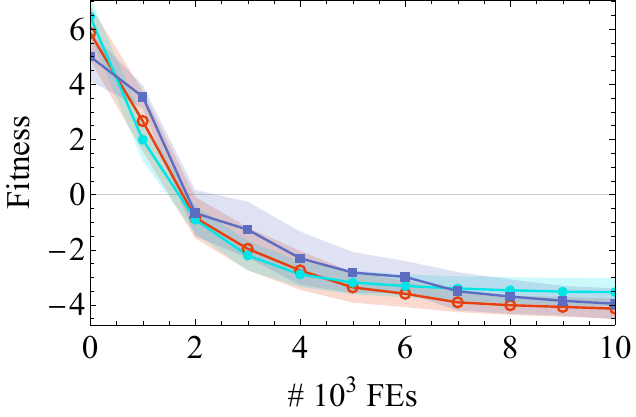}
        \vskip -0.5em
        \caption{Rover (60D)}
    \end{subfigure}
    \hfill
    \vskip 0.5em
    \begin{subfigure}[c]{0.325\textwidth}
        \raggedleft
        \includegraphics[scale=0.525]{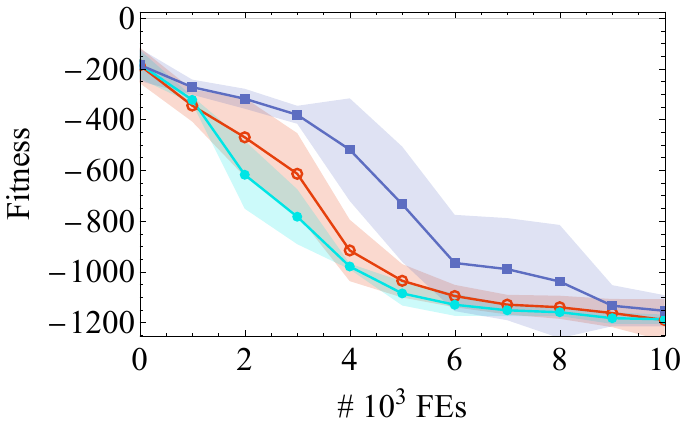}
        \vskip -0.5em
        \caption{Walker (198D)}
    \end{subfigure}
    \begin{subfigure}[c]{0.325\textwidth}
        \raggedleft
        \includegraphics[scale=0.525]{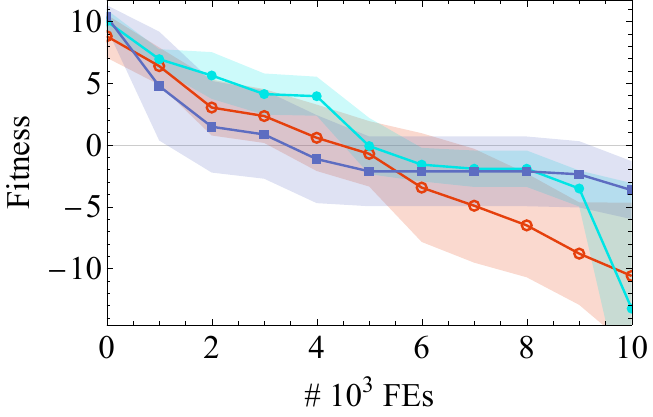}
        \vskip -0.5em
        \caption{Ant (404D)}
    \end{subfigure}
    \begin{subfigure}[c]{0.325\linewidth}
        \centering
        \vskip 1.2em
        \includegraphics[scale=0.9]{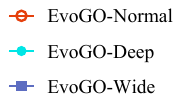}
        \vskip 1.2em
        \caption{Plot legend}
    \end{subfigure}
    \caption{Performance comparison of \MethodName and its ablations focusing on the neural network architecture of $\bm{\gamma}$ and $\bm{\gamma}'$ as described in Algorithm 2.}
    \label{fig:nn-arch-all}
\end{figure}
}

\newcommand{\figCycLossAll}{%
\begin{figure}[!thb]
    \centering
    \begin{subfigure}[c]{0.325\linewidth}
        \raggedleft
        \includegraphics[scale=0.525]{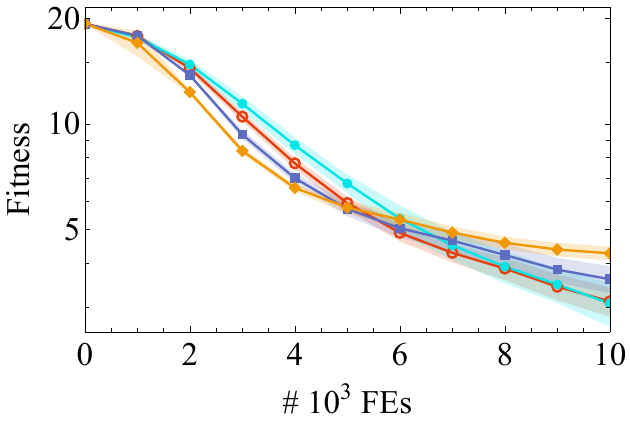}
        \vskip -0.5em
        \caption{Ackley (200D)}
    \end{subfigure}
    \begin{subfigure}[c]{0.325\textwidth}
        \raggedleft
        \includegraphics[scale=0.525]{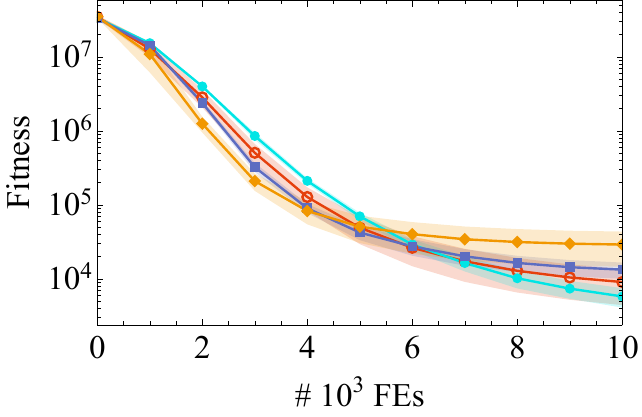}
        \vskip -0.5em
        \caption{Rosenbrock (200D)}
    \end{subfigure}
    \begin{subfigure}[c]{0.325\linewidth}
        \raggedleft
        \includegraphics[scale=0.525]{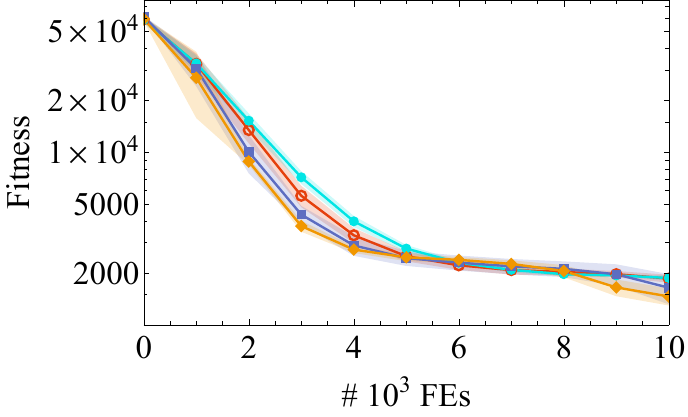}
        \vskip -0.5em
        \caption{Rastrigin (200D)}
    \end{subfigure}
    \hfill
    \vskip 0.5em
    \begin{subfigure}[c]{0.325\linewidth}
        \raggedleft
        \includegraphics[scale=0.525]{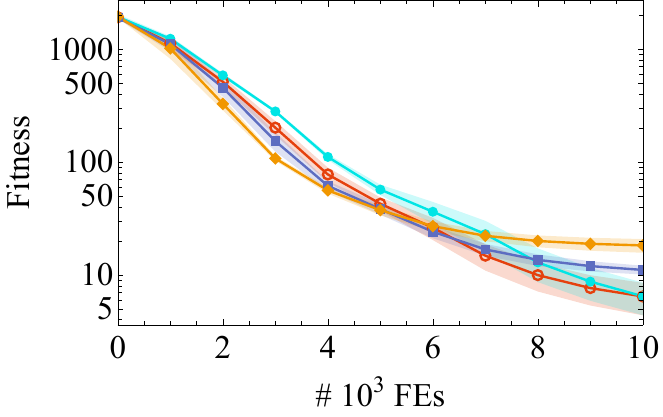}
        \vskip -0.5em
        \caption{Levy (200D)}
    \end{subfigure}
    \begin{subfigure}[c]{0.325\linewidth}
        \raggedleft
        \includegraphics[scale=0.525]{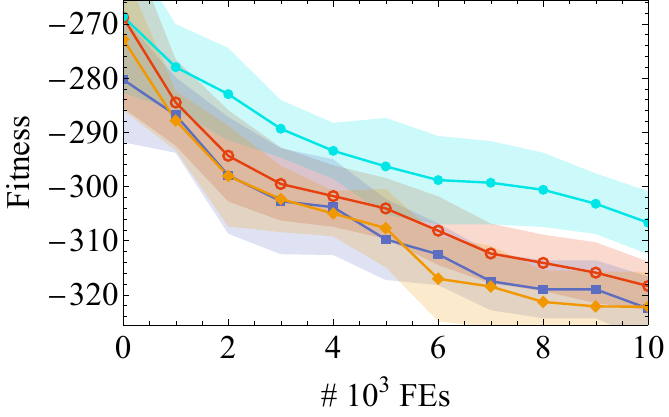}
        \vskip -0.5em
        \caption{Landing (12D)}
    \end{subfigure}
    \begin{subfigure}[c]{0.325\linewidth}
        \raggedleft
        \includegraphics[scale=0.525]{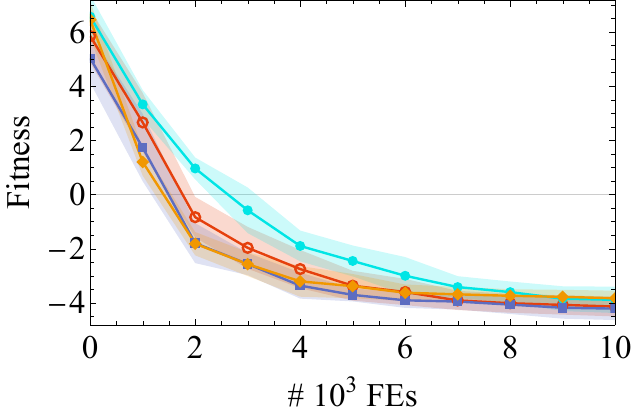}
        \vskip -0.5em
        \caption{Rover (60D)}
    \end{subfigure}
    \hfill
    \vskip 0.5em
    \begin{subfigure}[c]{0.325\textwidth}
        \raggedleft
        \includegraphics[scale=0.525]{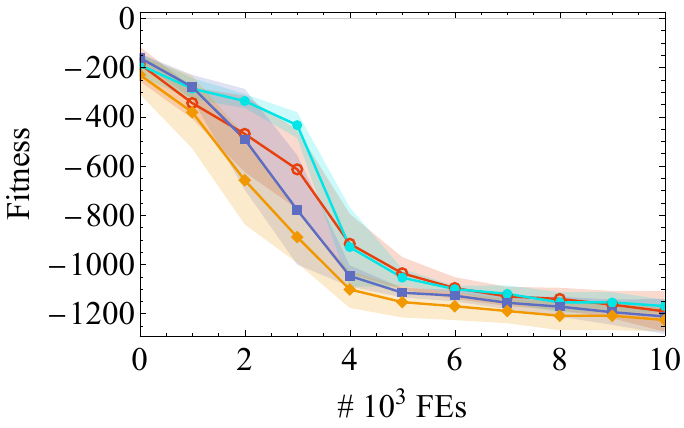}
        \vskip -0.5em
        \caption{Walker (198D)}
    \end{subfigure}
    \begin{subfigure}[c]{0.325\textwidth}
        \raggedleft
        \includegraphics[scale=0.525]{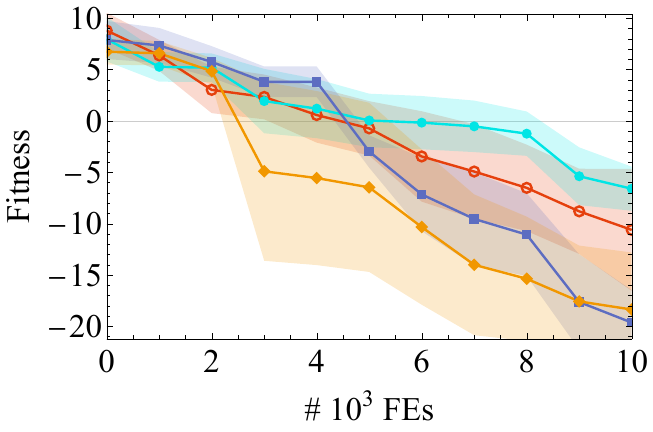}
        \vskip -0.5em
        \caption{Ant (404D)}
    \end{subfigure}
    \begin{subfigure}[c]{0.325\linewidth}
        \centering
        \vskip 1.2em
        \includegraphics[scale=0.9]{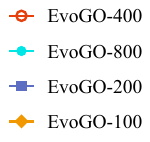}
        \vskip 1.2em
        \caption{Plot legend}
    \end{subfigure}
    \caption{Performance comparison of \MethodName and its ablations concerning the reconstruction loss scale $\lambda_1$ as outlined in the loss function as described in Eq. (2).}
    \label{fig:cyc-loss-all}
\end{figure}
}

\newcommand{\figPortionAll}{%
\begin{figure}[!thb]
    \centering
    \begin{subfigure}[c]{0.325\linewidth}
        \raggedleft
        \includegraphics[scale=0.525]{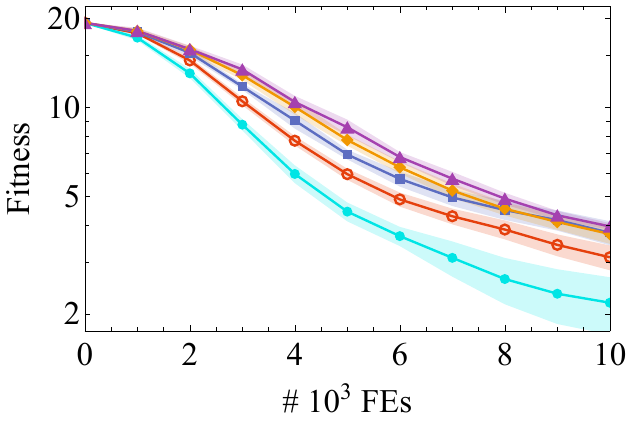}
        \vskip -0.5em
        \caption{Ackley (200D)}
    \end{subfigure}
    \begin{subfigure}[c]{0.325\textwidth}
        \raggedleft
        \includegraphics[scale=0.525]{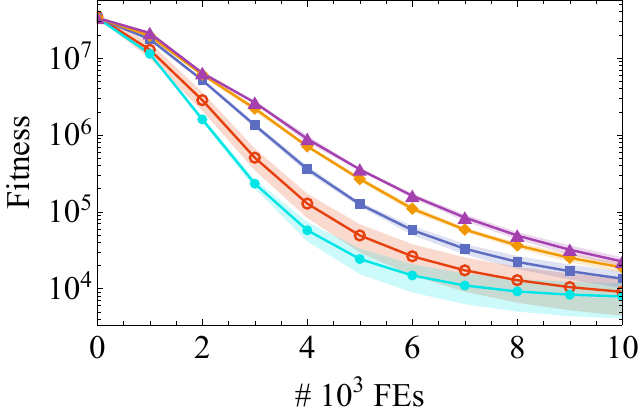}
        \vskip -0.5em
        \caption{Rosenbrock (200D)}
    \end{subfigure}
    \begin{subfigure}[c]{0.325\linewidth}
        \raggedleft
        \includegraphics[scale=0.525]{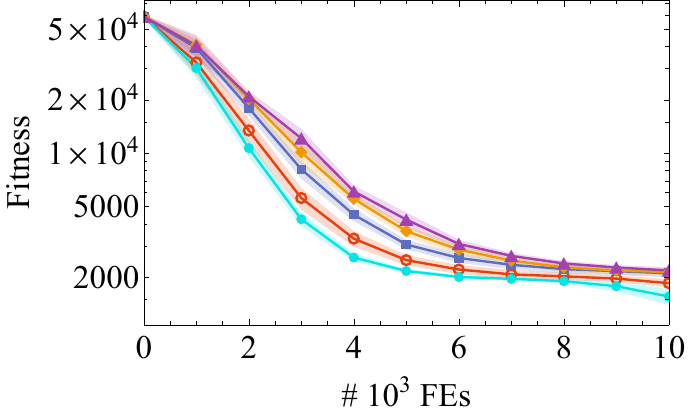}
        \vskip -0.5em
        \caption{Rastrigin (200D)}
    \end{subfigure}
    \hfill
    \vskip 0.5em
    \begin{subfigure}[c]{0.325\linewidth}
        \raggedleft
        \includegraphics[scale=0.525]{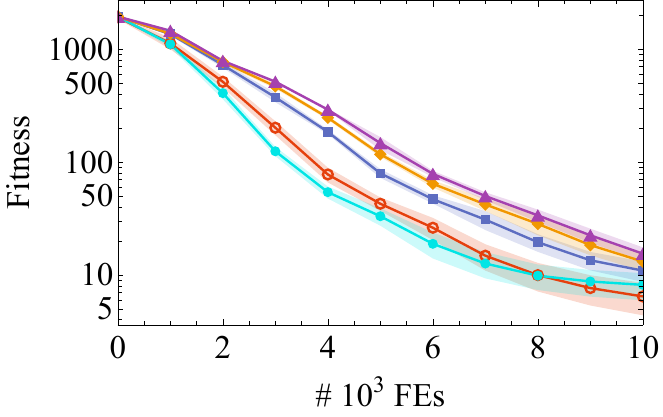}
        \vskip -0.5em
        \caption{Levy (200D)}
    \end{subfigure}
    \begin{subfigure}[c]{0.325\linewidth}
        \raggedleft
        \includegraphics[scale=0.525]{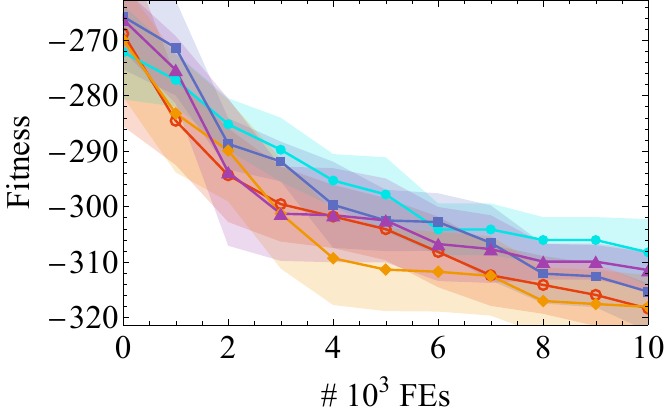}
        \vskip -0.5em
        \caption{Landing (12D)}
    \end{subfigure}
    \begin{subfigure}[c]{0.325\linewidth}
        \raggedleft
        \includegraphics[scale=0.525]{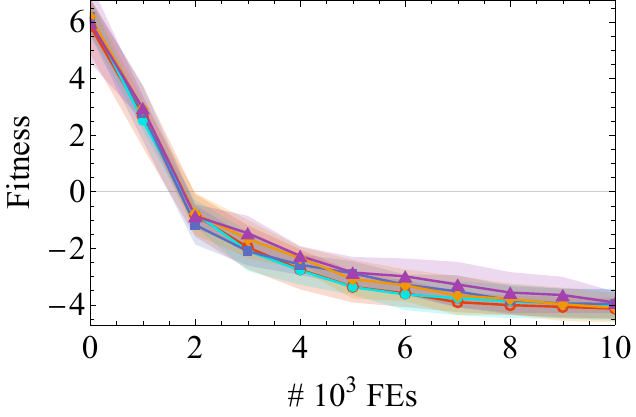}
        \vskip -0.5em
        \caption{Rover (60D)}
    \end{subfigure}
    \hfill
    \vskip 0.5em
    \begin{subfigure}[c]{0.325\textwidth}
        \raggedleft
        \includegraphics[scale=0.525]{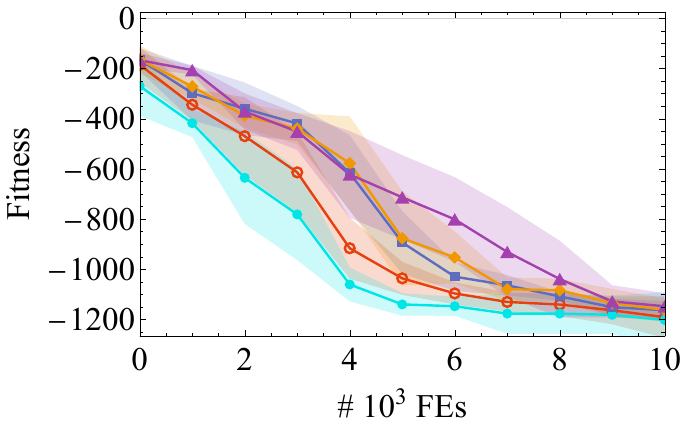}
        \vskip -0.5em
        \caption{Walker (198D)}
    \end{subfigure}
    \begin{subfigure}[c]{0.325\textwidth}
        \raggedleft
        \includegraphics[scale=0.525]{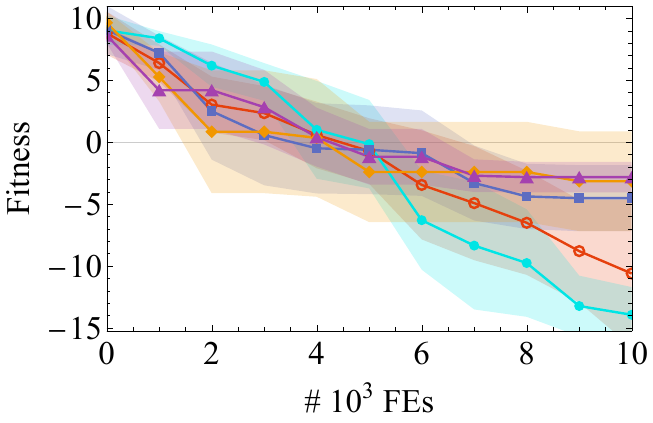}
        \vskip -0.5em
        \caption{Ant (404D)}
    \end{subfigure}
    \begin{subfigure}[c]{0.325\linewidth}
        \centering
        \vskip 1.2em
        \includegraphics[scale=0.9]{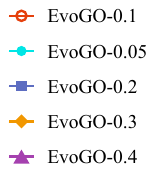}
        \vskip 1.2em
        \caption{Plot legend}
    \end{subfigure}
    \caption{Ablation analysis for the dataset split factor $\eta$, as per Algorithm 1.}
    \label{fig:portion-all}
\end{figure}
}

\newcommand{\figSlideAll}{%
\begin{figure}[!thb]
    \centering
    \begin{subfigure}[c]{0.325\linewidth}
        \raggedleft
        \includegraphics[scale=0.525]{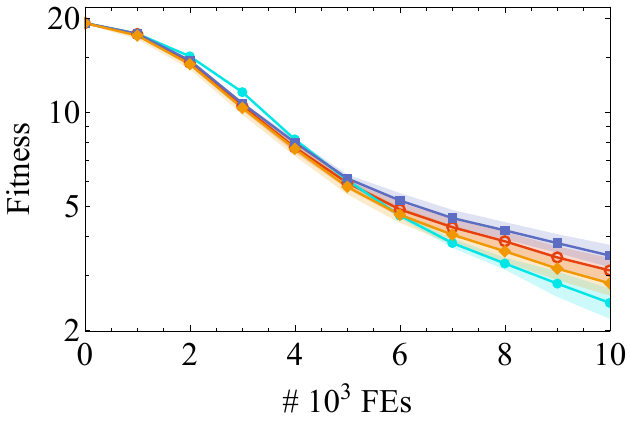}
        \vskip -0.5em
        \caption{Ackley (200D)}
    \end{subfigure}
    \begin{subfigure}[c]{0.325\textwidth}
        \raggedleft
        \includegraphics[scale=0.525]{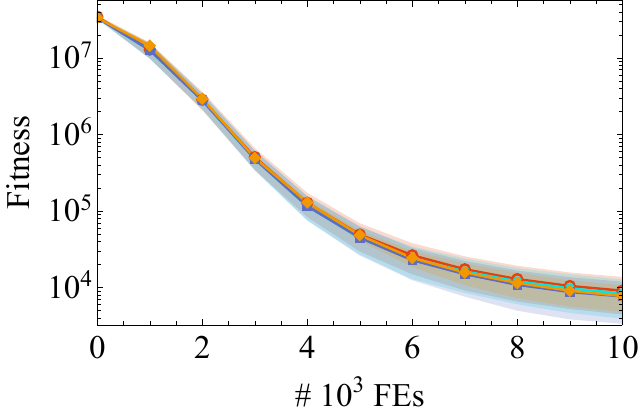}
        \vskip -0.5em
        \caption{Rosenbrock (200D)}
    \end{subfigure}
    \begin{subfigure}[c]{0.325\linewidth}
        \raggedleft
        \includegraphics[scale=0.525]{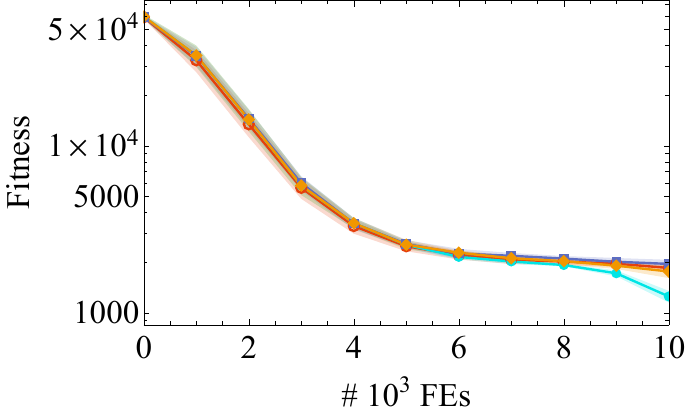}
        \vskip -0.5em
        \caption{Rastrigin (200D)}
    \end{subfigure}
    \hfill
    \vskip 0.5em
    \begin{subfigure}[c]{0.325\linewidth}
        \raggedleft
        \includegraphics[scale=0.525]{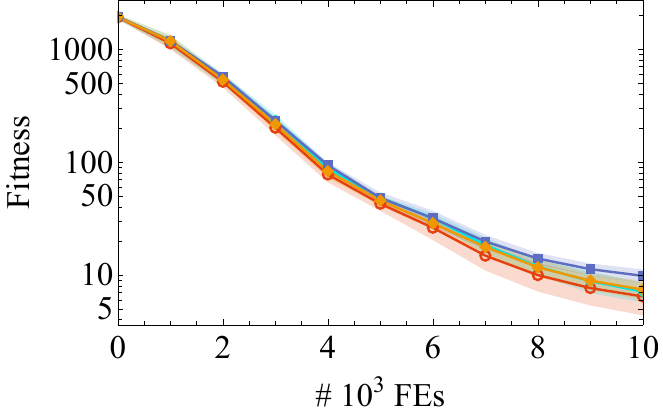}
        \vskip -0.5em
        \caption{Levy (200D)}
    \end{subfigure}
    \begin{subfigure}[c]{0.325\linewidth}
        \raggedleft
        \includegraphics[scale=0.525]{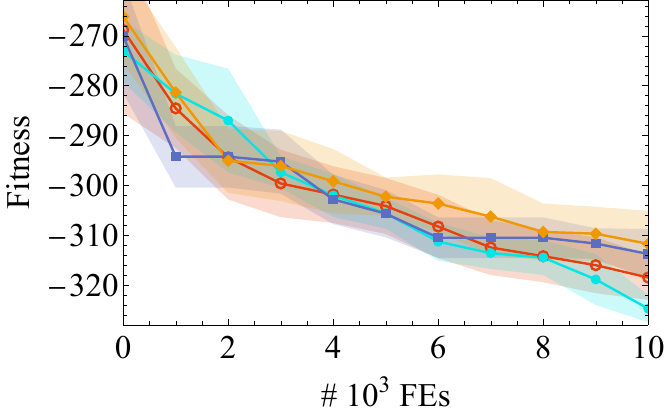}
        \vskip -0.5em
        \caption{Landing (12D)}
    \end{subfigure}
    \begin{subfigure}[c]{0.325\linewidth}
        \raggedleft
        \includegraphics[scale=0.525]{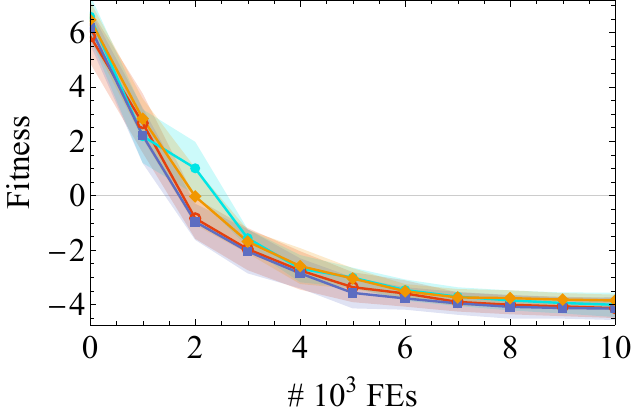}
        \vskip -0.5em
        \caption{Rover (60D)}
    \end{subfigure}
    \hfill
    \vskip 0.5em
    \begin{subfigure}[c]{0.325\textwidth}
        \raggedleft
        \includegraphics[scale=0.525]{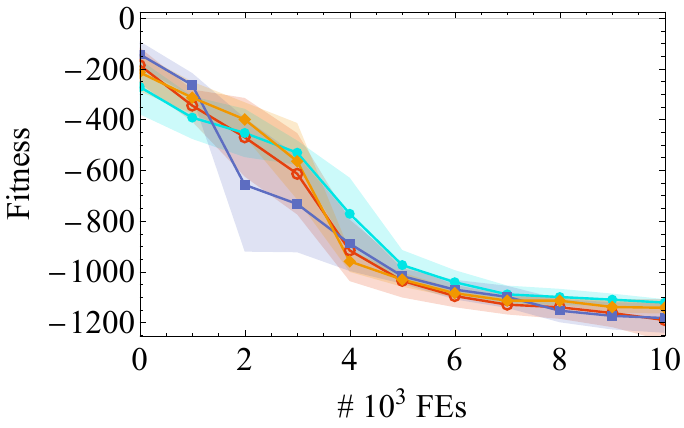}
        \vskip -0.5em
        \caption{Walker (198D)}
    \end{subfigure}
    \begin{subfigure}[c]{0.325\textwidth}
        \raggedleft
        \includegraphics[scale=0.525]{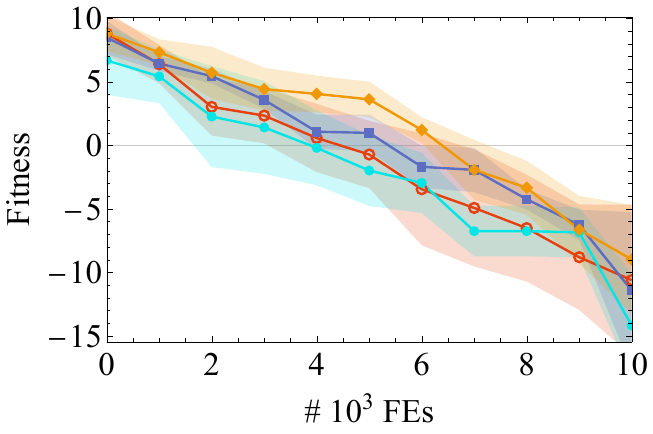}
        \vskip -0.5em
        \caption{Ant (404D)}
    \end{subfigure}
    \begin{subfigure}[c]{0.325\linewidth}
        \centering
        \vskip 1.2em
        \includegraphics[scale=0.9]{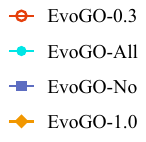}
        \vskip 1.2em
        \caption{Plot legend}
    \end{subfigure}
    \caption{Ablation analysis for the the sliding window factor $\epsilon$, as per Algorithm 1.}
    \label{fig:slide-all}
\end{figure}
}

\begin{abstract}
    Recent advances in data-driven evolutionary algorithms (EAs) have demonstrated the potential of leveraging historical data to improve optimization accuracy and adaptability. 
    \color{black}
    Despite these advancements, existing methods remain reliant on handcrafted process-level operators.  
    In contrast, \FullMethodName\ is a fully data-driven framework designed from the objective level, enabling autonomous learning of the entire search process.  
    \color{black}
    \MethodName streamlines the evolutionary optimization process into three stages: data preparation, model training, and population generation. 
    The data preparation stage constructs a pairwise dataset to enrich training diversity without incurring additional evaluation costs. 
    During model training, a tailored generative model learns to transform inferior solutions into superior ones. 
    In the population generation stage, \MethodName replaces traditional reproduction operators with a scalable and parallelizable generative mechanism.
    Extensive experiments on numerical benchmarks, classical control problems, and high-dimensional robotic tasks demonstrate that \MethodName consistently converges within merely 10 generations and substantially outperforms a wide spectrum of optimization approaches, including traditional EAs, Bayesian optimization, and reinforcement learning based methods. Code is available at: \url{https://github.com/EMI-Group/evogo}
    
    \end{abstract}
    
    \begin{IEEEkeywords}
        Generative learning, learning-based evolutionary algorithm, \color{black} black-box optimization \color{black}
    \end{IEEEkeywords}
    
    \section{Introduction}\label{sec:intro}
    
    Many real-world optimization tasks involve objectives that can be accessed only through simulations, legacy software, or physical experiments~\cite{Audet2017,Bajaj2021}. {\color{black} Such problems are typically non-analytical, noisy, and gradient-free, making them challenging for conventional optimization techniques. In these scenarios, Evolutionary Algorithms (EAs) provide a robust and widely applicable framework due to their global search capability, insensitivity to differentiability, and ease of parallelization. They have achieved success in diverse fields such as hyperparameter tuning, biomedical engineering, and aerospace design~\cite{Abdolrasol2021}. Nevertheless, while EAs offer a practical starting point for complex black-box problems, their classical forms remain limited in capturing structure or reusing information across generations~\cite{Kumagai2023,Liu2023,huang2024evox,wei2022distributed}.
    
    To enhance efficiency and stability, subsequent research reformulated EAs from an operator-based to a distribution-based paradigm. Representative examples include Evolution Strategies (ES)~\cite{Li2020} and Estimation of Distribution Algorithms (EDAs)~\cite{Larranaga2024}, which replace handcrafted variation operators with statistical modeling and parameterized sampling. This allows population updates through distributional shifts rather than discrete manipulations, while maintaining desirable invariance properties such as rotation or scale invariance~\cite{li2023multitask}. By estimating gradients in the distribution space, these methods introduce mechanisms like natural gradient ascent, covariance matrix adaptation~\cite{Hansen1996}, and factorized or low-rank modeling to better capture dependencies among decision variables\cite{Muehlenbein1997}. Such strategies effectively adjust the sampling distribution using the data collected from previous evaluations, thereby improving sample efficiency and convergence stability \cite{Baluja1994}. However, their optimization still focuses on fitting the sampling distribution itself, rather than learning a coherent decision process that couples evaluation and generation in a data-driven manner.
    
    To further enhance the consistency, adaptivity, and data efficiency of evolutionary optimization, subsequent studies have shifted attention from merely updating the sampling distribution to learning the evaluation process itself~\cite{Gutmann2001}. This line of research gave rise to data-driven evolutionary frameworks, particularly surrogate-assisted ones, where the objective landscape is approximated through learned evaluators or ranking models trained on historical samples~\cite{Zhen2023a, Wang2023a}. By replacing direct evaluations with predictive or ordinal surrogates, these methods achieve more consistent and controllable optimization dynamics while significantly reducing computational cost~\cite{Song2021, Liu2023b, Zhen2023}. Fundamentally, the evaluation step becomes a learned component that embeds accumulated experience into the search process, thereby narrowing the gap between evolutionary optimization and data-driven inference. However, their reliance on imperfect surrogates still introduces model bias, and repeated exploitation of surrogate predictions can lead to premature convergence and loss of exploration diversity.
    
    To mitigate these limitations, generative modeling-based evolutionary algorithms (GMEAs)~\cite{Pan2021,Liu2023a,Jiang2025} have emerged as a promising extension of data-driven evolutionary computation.
    Instead of learning to approximate the evaluation process, GMEAs focus on learning the generation process itself, representing the search dynamics through parameterized probabilistic models.
    By coupling the exploratory nature of evolutionary search with the expressive power of generative modeling, GMEAs establish a positive feedback loop between solution generation and model learning ~\cite{gao2024generative,wang2023learning, wang2025new,yan2024emodm}.
    GMEAs are particularly suited for complex black-box optimization problems that exhibit latent dependencies, implicit constraints, or simulation-based evaluations, where gradients or analytical forms are unavailable.
    Through data-driven modeling of the search distribution, GMEAs achieve a more balanced trade-off between exploration and exploitation, effectively alleviating the model bias and premature convergence issues often observed in surrogate-assisted approaches.
    
    However, existing GMEAs are still at an early stage and face several fundamental challenges.
    First, their integration with evolutionary workflows often relies on manually designed operators and procedural rules,
    thereby embedding task-specific heuristics into the algorithm rather than achieving automatic coordination between the generative model and the evolutionary process.
    Second, the training objectives of current generative models are typically inherited from domains such as image or text generation,
    leading to misalignment with optimization goals.
    Third, most methods depend on limited population data, which hampers training stability and generalization under tight evaluation budgets.
    These limitations highlight the need for a fully data-driven evolutionary framework that replaces handcrafted heuristics with end-to-end, learning-based mechanisms.
    To this end, we propose \FullMethodName, a unified framework that achieves evolutionary optimization entirely through generative learning.
    \color{black}}
    Our key contributions are summarized as follows:
    \begin{itemize}
        \item We design a paired generative architecture in which forward and inverse models collaborate to produce high-quality candidate solutions. This setup enables direct mapping from historical populations to new individuals, entirely replacing traditional crossover and mutation heuristics.
        \item We introduce a tailored loss function that combines reconstruction accuracy, distributional alignment, and directional guidance. This composite objective explicitly steers the generative process toward promising regions of the search space.
        \item We develop a learning-based sample synthesis strategy that generates informative training data from limited fitness evaluations. This mechanism supports stable model training even under small population sizes and tight evaluation budgets.
    \end{itemize}
    
    The remainder of the paper is organized as follows. \cref{sec:lr} reviews existing approaches for complex optimization and discusses their relationships to \MethodName.
    \cref{sec:method} introduces the proposed model architecture, loss function, and learning strategies.
    \cref{sec:experiment} presents a comprehensive empirical evaluation, including performance comparisons and ablation studies.
    Finally, \cref{sec:conclusion} concludes the paper.
    
    \section{Related Work}\label{sec:lr}
    
    \subsection{Non-Data-Driven Approaches}
    
    EAs represent the most prevalent non-data-driven paradigm for complex optimization. These approaches rely on fixed heuristics and do not leverage learned, data-driven models to guide search. EAs can be broadly categorized based on their mechanism: (i) methods that generate solutions via handcrafted search operators, and (ii) methods that probabilistically model the distribution of promising solutions to guide sampling.
    
    Methods based on handcrafted operators, such as genetic algorithms~\cite{Holland1992}, genetic programming~\cite{Koza1994}, differential evolution~\cite{Storn1997}, and particle swarm optimization~\cite{Kennedy1995}, constitute the first category. These algorithms rely on fixed, manual operators (e.g., crossover or mutation) to explore the search space. Despite their widespread adoption, their performance is heavily dependent on domain knowledge and requires substantial manual tuning, which restricts both flexibility and scalability. To address these limitations, the second category attempts to guide the evolutionary process by learning distributions over promising solutions.
    
    \color{black}
    The second category comprises Distribution-based EAs, which aim to improve search efficiency by modeling the distribution of successful solutions. This category includes Evolution Strategies (ES) and Estimation of Distribution Algorithms (EDAs). ES, originating from early stochastic optimization studies~\cite{Rechenberg1965}, achieved a major advancement with Covariance Matrix Adaptation ES (CMA-ES), which adaptively adjusts the sampling distribution based on population statistics~\cite{Hansen1996} and was later extended to a fully derandomized formulation~\cite{Hansen2001}. More recently, ES variants such as OpenAI-ES~\cite{salimans2017evolution} have demonstrated strong scalability and parallelism in large-scale machine learning tasks. In parallel, EDAs explicitly incorporate statistical modeling into evolutionary search by learning the distribution of promising solutions. Early examples include Population-Based Incremental Learning (PBIL)~\cite{Baluja1994} and the Univariate Marginal Distribution Algorithm (UMDA)~\cite{Muehlenbein1997}. Subsequent methods, such as the extended compact genetic algorithm~\cite{Harik1997} and the linkage tree genetic algorithm~\cite{Thierens2010}, capture variable dependencies through more expressive probabilistic models.
    
    \color{black}
    Despite their conceptual elegance, both ES and EDA methods remain fundamentally constrained by the non-data-driven paradigm. The underlying update and sampling procedures are still manually crafted and adhere to rigid algorithmic rules, which severely restricts their adaptability across diverse or complex optimization scenarios~\cite{Li2020,Larranaga2024}. To address this lack of flexibility, some research efforts have explored incorporating reinforcement learning (RL) into EAs to automate the selection or configuration of evolutionary operators~\cite{Eiben2007,Wang2022,Tian2023,Drugan2019,zhao2024policy}. These methods typically employ external RL agents to dynamically choose strategies based on past performance. However, while this integration offers a degree of adaptivity, the learning process is still external to the core optimization loop, often incurring substantial computational overhead and poor sample efficiency~\cite{Li2024}. Consequently, these approaches do not fundamentally depart from the heuristic nature of traditional EAs, and the search dynamics remain largely heuristic in nature.
    
    \subsection{Data-Driven Apporoaches}
    \color{black}
    Traditional EAs primarily rely on handcrafted heuristics and static update rules. However, recent research has increasingly shifted toward data-driven paradigms that exploit information accumulated during the optimization process. These methods employ machine learning techniques to enhance the efficiency, scalability, and adaptability of evolutionary optimization. Consequently, these approaches are generally categorized into three representative categories: Bayesian Optimization (BO), Surrogate-Assisted Evolutionary Algorithms (SAEAs), and Generative Modeling-based Evolutionary Algorithms (GMEAs).
    
    \color{black}
    \subsubsection{Bayesian Optimization}
    
    Bayesian Optimization (BO) traditionally employs Gaussian processes to model the objective function and constructs acquisition functions to determine promising query points~\cite{Jones1998}. Substantial efforts have been made to enhance the scalability of BO to address performance challenges in large-scale and high-dimensional settings~\cite{Srinivas2010, Snoek2012, Gardner2017, Frazier2018, Riquelme2018, Mutnỳ2018, Zhai2022, Ament2023}. Despite these advances, significant computational overhead persists, especially when BO is applied to complex tasks on modern high-throughput computing platforms.
    
    To address this persistent computational overhead, researchers have proposed multi-point acquisition strategies designed to enable efficient batch querying~\cite{Azimi2010, KonakovicLukovic2020, Daulton2021, Chen2023}. Among these approaches, Thompson Sampling (TS)~\cite{Thompson1933} has gained particular popularity for use in parallel BO schemes~\cite{Eriksson2019, Nguyen2020, Eriksson2021, DeAth2021}. For instance, Eriksson \emph{et al.} integrated TS with a trust region method to achieve scalable BO for constrained and unconstrained problems, supporting batch sizes up to 100. Nevertheless, in the majority of cases, these batch sizes remain relatively modest, thereby resulting in the underutilization of modern computational resources.
    
    In parallel with multi-point methods, Artificial Neural Networks (ANNs) have also been explored to improve the batch efficiency of BO~\cite{Swersky2020, Tiao2021, Oliveira2022}. These approaches, however, often repurpose the surrogate model as a binary classifier rather than directly generating new candidate solutions. For instance, density-ratio-based methods were proposed by Tiao \emph{et al.} and Oliveira \emph{et al.}, while Swersky \emph{et al.} adopted reinforcement learning to train a selection policy for discrete domains. Despite their contributions, these methods still fundamentally rely on direct pointwise optimization for generating new queries~\cite{Wang2023}, a mechanism that ultimately hinders the full scalability of BO.
    
    \color{black}
    \subsubsection{Surrogate-Assisted Evolutionary Algorithms}
    
    Surrogate-Assisted Evolutionary Algorithms (SAEAs) aim to enhance sample efficiency in optimization problems where fitness evaluations involve costly simulations~\cite{guo2024classifier,wei2022distributed}. Early frameworks in model-based evolutionary computation introduced approximation models to reduce evaluation cost~\cite{Bull1999,Jin2002}. Traditional SAEAs typically employ simple surrogate models, such as Radial Basis Functions (RBFs) and Support Vector Machines (SVMs)~\cite{Jin2005}; however, these models are limited in capturing the complexity of high-dimensional and nonlinear search landscapes. The subsequent introduction of Gaussian Processes (GPs) alleviated some of these constraints by incorporating uncertainty estimation into the infill process. For instance, Liu~\emph{et al.}~\cite{Liu2014} combined GPs with standard EA operators to effectively balance exploration and exploitation.
    
    However, since only a limited number of training samples are available during the optimization process, a single surrogate model often struggles to generalize across heterogeneous problem domains or adapt to different search stages. To address this issue, recent studies have proposed multi-surrogate frameworks that better capture diverse landscape features and mitigate model bias. These frameworks can be broadly classified into two categories: hierarchical-model SAEAs and ensemble-model SAEAs. The hierarchical approach combines global and local models to improve scalability~\cite{Wang2023a,Wang2019,Cai2020}, whereas ensemble paradigms integrate multiple predictors to enhance accuracy and reduce overfitting~\cite{Wu2023,Guo2019}. In parallel, adaptive-model SAEAs have been developed to automatically select appropriate surrogate models or infill criteria according to evolving problem characteristics~\cite{Liu2023b,Zhen2023,He2023}. Despite these advances, most SAEAs still rely heavily on surrogate predictions for candidate selection, which can be computationally demanding and prone to model bias~\cite{He2023}.
    
    \subsubsection{Generative Modeling-based Evolutionary Algorithms}
    
    Generative Modeling-based Evolutionary Algorithms (GMEAs) aim to learn and adapt the sampling distribution from data to guide the evolutionary search process. Current research in this area mainly follows two directions: (i) leveraging unsupervised learning for variable grouping and dimensionality reduction~\cite{Zhang2014,Tian2020,Pan2021}, and (ii) employing supervised or reinforcement learning to guide the directional improvement of candidate solutions~\cite{Bandaru2010,Gaur2017,He2021,Liu2023a,Jiang2025}.
    
    The first line of work trains generative models to discover latent structures within the search space, thereby facilitating more efficient variation. For example, Pan~\emph{et al.}~\cite{Pan2021} proposed a manifold-learning-based approach that restricts mating operations to low-dimensional subspaces, improving optimization efficiency.
    
    The second line focuses on guiding evolutionary search using data-driven feedback. He~\emph{et al.}~\cite{He2021} introduced a GAN-based multi-objective EA in which a discriminator distinguishes real from synthetic populations to train a generator that produces new candidates. Similarly, Liu~\emph{et al.}~\cite{Liu2023a} employed reinforcement learning to adaptively select variation operators based on individual context and historical performance. While GMEAs show strong potential, several open challenges remain. These include their continued reliance on manually designed heuristic components, the difficulty of seamlessly integrating generative models into the evolutionary loop, and ambiguities in defining effective training objectives.
    
    \subsection{Discussion}
    
    The evolution from non-data-driven to data-driven paradigms in evolutionary computation reflects a broader shift toward leveraging information accumulated during optimization. Despite significant progress across BO, SAEA, and GMEAs, each paradigm exhibits distinct advantages and limitations that restrict its general applicability.
    
    BO offers strong theoretical foundations and high sample efficiency but suffers from poor scalability due to the cubic complexity of Gaussian process inference and reliance on pointwise query generation. In contrast, SAEA improves efficiency by introducing learned evaluators to approximate the fitness landscape; however, its focus on local modeling often results in bias and premature convergence. Finally, GMEA emphasizes learning the distribution of promising solutions from data, enabling better structural modeling and diversity maintenance. Nevertheless, current GMEA implementations still rely on manually designed heuristic components and lack effective coupling between the generation and evaluation phases.
    
    To address the existing limitations identified in current approaches, the research imperative is to develop a unified, end-to-end learning framework that seamlessly integrates representation learning, solution generation, and fitness evaluation within a cohesive pipeline. Such a framework must satisfy key design requirements, specifically seeking to (i) eliminate reliance on ad hoc or handcrafted coordination mechanisms, (ii) ensure rigorous alignment between the learning objectives of each internal module and the global optimization goal, and (iii) effectively leverage all available information to maximize data efficiency and adaptability.
    
    \color{black}
    
    \section{Method}\label{sec:method}
    
    This section introduces the overall framework of \MethodName, followed by detailed descriptions of each key component.
    
    \figFramework
    
    \subsection{Overall Framework}
    
    \algData
    
    As illustrated in \cref{fig:framework}, \MethodName comprises three key, sequentially executed phases: data preparation, model training, and population generation. To streamline the evolutionary process and eliminate reliance on complex handcrafted rules, \MethodName adopts an end-to-end generative paradigm, replacing the heuristic components commonly used in traditional GMEAs.

    This generative architecture directly addresses two core challenges inherent in existing GMEAs: data scarcity and objective misalignment. Specifically, the data preparation phase incorporates a data augmentation strategy to mitigate the risk of overfitting caused by limited population data, thereby enhancing training diversity without introducing additional evaluation costs. Furthermore, in the model training phase, a composite model architecture and a tailored loss function are employed to align the generative objectives with the goals of evolutionary optimization, thereby resolving the objective misalignment issue.
    The time and space complexity of \MethodName are analyzed in detail in Supplementary Document~A.
    
    \subsection{Data Preparation}\label{sec:method-data}
    
    The data preparation phase is pivotal for constructing high-quality and well-structured training data that is sufficient for both the surrogate model and the generative model, particularly under limited evaluation budgets. Achieving high generalization capabilities for the generative model requires rich, diverse input data, whereas the surrogate model demands highly promising samples for accurate fitness estimation. Consequently, this phase is designed to manage the trade-off between diversity and fidelity to maximize data utility.
    
    As detailed in \cref{alg:data}, the data preparation process comprises the following three steps:
    
    \begin{itemize}
        \item \textbf{Elite Selection}:  
        A high-quality dataset $\mathcal{D}_{\text{SM}}$ is constructed by selecting individuals with the highest fitness values from all historical populations $\mathcal{D}_0 \cup \cdots \cup \mathcal{D}_t$. 
        This process ensures accurate surrogate modeling by focusing on representative and promising solutions.
    
        {\color{black}
        \item \textbf{Data Augmentation (Optional)}:  
        When the evaluation budget is limited, requiring the algorithm to operate with a small population, a learning-based data augmentation strategy is adopted. 
        Specifically, a Variational AutoEncoder (VAE) is used to generate diverse synthetic samples that resemble the current population. This process enriches the generative model dataset $\mathcal{D}_{\text{GM}}$, thereby alleviating the data scarcity issue under restricted evaluations.
    
        \item \textbf{Paired Set Construction}:  
            Instead of enlarging the dataset further, this step enhances data utilization by constructing informative sample pairs. 
            The surrogate model estimates the fitness of all candidates, after which the samples are ranked and divided into \emph{superior} ($\mathcal{D}_{\text{GM}^+}$) and \emph{inferior} ($\mathcal{D}_{\text{GM}^-}$) subsets according to a predefined ratio $\eta$. 
            The Cartesian product of the two subsets forms the paired training set $\mathcal{D}_{\text{GM}}$, enabling the generative model to learn both effective descent directions and adaptive step magnitudes through relational supervision. }
    \end{itemize}
    
    \subsection{Model Training}\label{sec:train}
    
    The model training phase focuses on establishing a composite model architecture integrated with a dedicated composite loss function. This architecture enables directional learning and optimization-aware training through an online process, where lightweight training is conducted during each evolutionary generation. Although the composite structure involves three model components, only the forward generative model $\bm{\gamma}$ is ultimately used during the population generation stage. The subsequent text details the overall training procedure, beginning with the model architecture and followed by the formal loss formulation.
    
    \subsubsection{Model Architecture}\label{sec:model}
    
    {\color{black}
    Inspired by Generative Adversarial Networks (GANs) \cite{Goodfellow2014} and CycleGANs \cite{Zhu2017}, we develop a composite-model-based GMEA framework (see \cref{fig:model}) that integrates a surrogate model $\tilde{f}$ with two complementary generative components. These components are a forward generative model $\bm{\gamma}$, which maps inferior solutions to superior ones, and a pseudo-inverse model $\bm{\gamma}'$, which performs the reverse mapping. Unlike conventional GAN-like architectures that rely on discriminators, the proposed framework employs the surrogate model to enforce optimization-aware learning, thereby aligning generative behaviors with the underlying objective landscape.
    }
    
    Given an evolving population, the composite architecture is designed to generate new solutions that possess improved function values. The dual generative models, $\bm{\gamma}$ and $\bm{\gamma}'$, implement the generative loss $\mathcal{L}_{\text{gen}}$ (Eq. \ref{eq:gen} and Eq. \ref{eq:loss-gen}), whereas the surrogate model $\tilde{f}$ addresses the optimization loss $\mathcal{L}_{\text{opt}}$. Specifically, $\bm{\gamma}$ processes \emph{inferior} solutions to generate estimated \emph{superior} ones, and $\bm{\gamma}'$ operates in reverse, both facilitating the computation of the similarity loss $\mathcal{L}_{\text{sim}}$.
    
    The integrated architecture further facilitates the transfer of optimization landscape estimation from the surrogate model to the dynamic generative components. Initially, the surrogate model $\tilde{f}$ is established and remains constant throughout the training process. Subsequently, the solutions generated by $\bm{\gamma}$ and $\bm{\gamma}'$ are reconstructed by cycling back to each other, which facilitates the computation of the reconstruction loss $\mathcal{L}_{\text{rec}}$ (Eq. \ref{eq:loss-rec}). Crucially, the generated solutions ($\bm{\gamma}(\bm{p})$ and $\bm{\gamma}'(\bm{q})$) are also evaluated against the surrogate, contributing to the optimization loss $\mathcal{L}_{\text{opt}}$. This configuration allows for the transfer of optimization landscape information via back-propagation, thereby enabling $\bm{\gamma}$ to autonomously generate optimized populations.
    
    In summary, the architecture synergistically integrates a static surrogate model $\tilde{f}$ with a dynamic generative model $\bm{\gamma}$ and its pseudo-inverse $\bm{\gamma}'$. $\bm{\gamma}$ and $\bm{\gamma}'$ are specifically trained for generating and reconstructing the \emph{superior} and \emph{inferior} solutions, respectively. Conversely, $\tilde{f}$ focuses exclusively on the estimation of the optimization landscape. This synergistic training substantially enhances the practical ability of $\bm{\gamma}$ to generate optimized solutions.
    
    \subsubsection{Loss Function} \label{sec:loss}
    
    As illustrated in \cref{fig:model}, we define a composite loss function that integrates two principal components: the \emph{generative loss} $\mathcal{L}_{\text{gen}}$ and the \emph{optimization loss} $\mathcal{L}_{\text{opt}}$:
    
    \begin{equation}
        \mathcal{L} = \mathcal{L}_{\text{gen}} + \lambda \mathcal{L}_{\text{opt}},
        \label{eq:loss}
    \end{equation}
    
    where $\lambda$ is a trade-off parameter that balances generative learning and optimization guidance.
    
    The generative loss $\mathcal{L}_{\text{gen}}$ comprises two core components. The first is the \emph{similarity loss} $\mathcal{L}_{\text{sim}}$, which encourages consistent latent representations between paired samples. The second is the \emph{reconstruction loss} $\mathcal{L}_{\text{rec}}$, which enforces structural consistency in the generated outputs:
    
    \begin{equation}
        \mathcal{L}_{\text{gen}} = \mathcal{L}_{\text{sim}} + \lambda_1 \mathcal{L}_{\text{rec}}.
        \label{eq:gen}
    \end{equation}
    
    The overall optimization loss $\mathcal{L}_{\text{opt}}$ is formulated as the combination of two components: the \emph{expectation loss} $\mathcal{L}_{\text{mean}}$, which penalizes poor average fitness among generated samples, and the \emph{variance loss} $\mathcal{L}_{\text{std}}$, which promotes diversity:
    
    \begin{equation}
        \mathcal{L}_{\text{opt}} = \mathcal{L}_{\text{mean}} + \lambda_2 \mathcal{L}_{\text{std}},
        \label{eq:loss-opt-base}
    \end{equation}
    
    where $\lambda_1$ and $\lambda_2$ are hyperparameters controlling the contributions of the reconstruction and variance terms, respectively.
    
    The similarity loss $\mathcal{L}_{\text{sim}}$ is conventionally employed in generative learning paradigms to ensure that generated outputs closely resemble the target samples~\cite{Ghifary2016,Zhu2017,Csurka2017,Kumar2020}. Specifically, for the generative model $\bm{\gamma}$, which transforms a source solution $\bm{p}$ to a target solution $\bm{q}$, we adopt a mean squared loss:
    
    \begin{equation}
        \mathcal{L}_{\text{sim}} = \|\bm{\gamma}(\bm{p}) - \bm{q}\|^2.
    \end{equation}
    
    This aligns with conventional generative paradigms, where models learn mappings from latent or random variables to target domains~\cite{Goodfellow2014,Kingma2014a,Mirza2014}. However, when applied to optimization problems, such models often struggle with the intrinsic trade-off between exploration and exploitation, especially within the EA framework~\cite{Liu2023,Liang2023a,Zhang2023}. Alternatively, domain adaptation models, which transform solutions between source and target domains~\cite{Ganin2015,Liu2016,Zhu2017,Tzeng2017}, have shown promising results in certain optimization scenarios~\cite{Zhang2023,Sun2024}. Nonetheless, without reconstruction constraints, these models are susceptible to mode collapse~\cite{Ghifary2016,Csurka2017,Kumar2020}, thereby compromising training stability. To mitigate this issue, we introduce a paired generative model $\bm{\gamma}'$ that serves as a pseudo-inverse of $\bm{\gamma}$. This yields a reconstruction loss:
    
    \begin{equation}
        \mathcal{L}_{\text{rec}} = \|\bm{\gamma}'(\bm{\gamma}(\bm{p})) - \bm{p}\|^2 + \|\bm{\gamma}(\bm{\gamma}'(\bm{q})) - \bm{q}\|^2,
        \label{eq:loss-rec}
    \end{equation}
    
    thereby leading to the similarity loss symmetrized as:
    
    \begin{equation}
        \mathcal{L}_{\text{sim}} = \|\bm{\gamma}(\bm{p}) - \bm{q}\|^2 + \|\bm{\gamma}'(\bm{q}) - \bm{p}\|^2.
        \label{eq:loss-sim}
    \end{equation}
    
    The generative loss $\mathcal{L}_{\text{gen}}$, particularly the reconstruction term, plays an essential role in stabilizing training and preserving semantic fidelity.
    This mechanism is crucial in optimization contexts, where the model must accurately learn transformations from \emph{inferior} to \emph{superior} regions of the search space without drifting off-manifold.
    To ensure this, the weight $\lambda_1$ associated with the reconstruction loss is set to dominate the loss function, emphasizing faithful reconstruction over mere proximity.
    
    By integrating the similarity and reconstruction losses for the bidirectional models $\{\bm{\gamma}, \bm{\gamma}'\}$, we obtain the full generative loss:
    \begin{align}
        \mathcal{L}_{\text{gen}} &= \left( \|\bm{\gamma}(\bm{p}) - \bm{q}\|^2 + \|\bm{\gamma}'(\bm{q}) - \bm{p}\|^2 \right) \nonumber \\
        &\quad + \lambda_1 \left( \|\bm{\gamma}'(\bm{\gamma}(\bm{p})) - \bm{p}\|^2 + \|\bm{\gamma}(\bm{\gamma}'(\bm{q})) - \bm{q}\|^2 \right).
        \label{eq:loss-gen}
    \end{align}
    
    This formulation adheres to the principles of domain-adaptive generative learning, where $\bm{p}$ and $\bm{q}$ are samples from source (\emph{inferior}) and target (\emph{superior}) domains, respectively.
    The dataset construction process is detailed in~\cref{sec:method-data}, and an alternative interpretation of~\cref{eq:loss-gen} from a probabilistic perspective is provided in Supplementary Document B.
    
    Integrating the generative model into the EA framework necessitates transforming traditional optimization objectives into suitable loss function terms for a complete data-driven process. This requirement is analogous to the infill criteria used in SAEAs, where the surrogate model selects the most promising solutions from a population of candidates \cite{He2023}. Such selection strategies have been widely studied in the literature. One of the most common approaches is to use the Upper/Lower Confidence Bound (UCB/LCB) \cite{Srinivas2010} as the infill criteria \cite{Liu2014}:
    
    \begin{equation}
        \alpha(\bm{p}) = \mathbb{E} \left[ \tilde{f}(\bm{p}) \middle| \bm{p} \right] + \lambda_2 \mathbb{V} \left[ \tilde{f}(\bm{p}) \middle| \bm{p} \right]^{1/2},
        \label{eq:lcb}
    \end{equation}
    
    Traditional Surrogate-Assisted Evolutionary Algorithms (SAEAs) utilize the infill criteria, $\alpha(\bm{p})$, to guide the selection of candidate solutions $\bm{p}$ for high-fidelity evaluation. Specifically, $\tilde{f}$ denotes the surrogate of the objective function being minimized, and $\lambda_2$ is a hyperparameter defined in \cref{eq:loss-opt-base}. All candidate solutions are evaluated using the criterion established in \cref{eq:lcb} to identify those with minimal $\alpha(\bm{p})$ values, which are subsequently passed to the true objective function for final evaluation.
    
    \figModel
    
    However, the conventional application of infill criteria, such as UCB or LCB, is not directly applicable to our proposed generative model $\bm{\gamma}$. This is because the standard infill criteria must be adapted to cohesively incorporate both the surrogate model and the generative model simultaneously. Inspired by meta-learning principles~\cite{Andrychowicz2016,Chen2017}, we propose to reinterpret the infill criteria as an optimization loss for training the generative model. Without loss of generality, we define the generative model as $\bm{\gamma}(\cdot; \bm{\phi}): \mathbb{R}^d \rightarrow \mathbb{R}^d$, parameterized by $\bm{\phi}$, which generates a new solution $\bm{x}_{t+1}$ from an existing input $\bm{x}$:
    \begin{align}
        &\bm{x}_{t+1} = \bm{\gamma}\left(\bm{x}; \argmin_{\bm{\phi}}\mathcal{L}(\bm{\phi}) \right), \nonumber \\
        &\text{where} \quad \mathcal{L}(\bm{\phi}) = \alpha(\bm{\gamma}(\bm{x};\bm{\phi})).
        \label{eq:af2model}
    \end{align}
    In this formulation, the infill criteria $\alpha(\cdot)$ explicitly defines the optimization objective used for training the parameters $\bm{\phi}$ of the generative model. Through this transformation, the task of minimizing the infill criteria is seamlessly integrated into the learning process of $\bm{\gamma}$, thereby allowing the model to internalize the required exploration-exploitation trade-offs during training.
    
    Thus, by replacing the underlying objective function $f$ with its surrogate model $\tilde{f}$ in this generalized framework, we derive the resulting optimization loss for UCB/LCB as:
    
    \begin{equation}
        \mathcal{L}_{\text{opt}} = \mathbb{E} \left[ \tilde{f}_t(\bm{\gamma}(\bm{p})) \right] - y_{\bm{p}} + \lambda_2 \mathbb{D} \left[ \bm{\gamma}(\bm{p}) \right]^{1/2},
        \label{eq:loss-opt-cb}
    \end{equation}
    
    where $\mathbb{E}$ and $\mathbb{D}$ represent the mean and variance of the surrogate model's output, respectively. This loss ties the generative model's parameters directly to the surrogate's predictions, facilitating a robust integration of the surrogate model into a generative modeling context. However, since the transformation of the infill criteria to the optimization loss is not guaranteed to be lossless, the inherent balance between exploration and exploitation may be compromised. Hence, we introduce an explicit variance optimization loss composed of UCB/LCB and Knowledge Gradient (KG) \cite{Frazier2009}:
    \begin{align}
        \mathcal{L}_{\text{opt}} &= \alpha_{\text{KG}}(\bm{p}) \nonumber \\
        &= \mathbb{E} \left[ \mu^*_{t+1} - \mu^*_t \mid \bm{p} \right] + \lambda_2 \mathbb{D} \left[ \mu^*_{t+1} - \mu^*_t \mid \bm{p} \right]^{1/2} \nonumber \\
        &= \mathbb{E} \left[ \tilde{f}_t(\bm{\gamma}(\bm{p})) \right] - y_{\bm{p}} + \lambda_2 \mathbb{D} \left[ \bm{\gamma}(\bm{p}) \right]^{1/2} + \nonumber \\
        &\lambda_2 \mathbb{C} \left[ \tilde{f}_t(\bm{\gamma}'\circ\bm{\gamma}(\bm{p})), \tilde{f}_t(\bm{q}) \right] + \lambda_2 \mathbb{C} \left[ \tilde{f}_t(\bm{\gamma}\circ\bm{\gamma}'(\bm{q})), \tilde{f}_t(\bm{p}) \right],
        \label{eq:loss-opt}
    \end{align}
    where $\mathbb{C}$ is the correlation function. Since the loss function is to be minimized, the existence of the correlation function requests that the parameter $\lambda_2$ is negative, which is not necessarily the case in \cref{eq:loss-opt-cb}. Supplementary Document B will elaborate the details of the justification of \cref{eq:loss-opt}.
    
    Unlike in conventional SAEAs, the surrogate model $\tilde{f}$ in our framework is not directly used to evaluate the quality of generated solutions. Instead, it serves only as a guidance mechanism during the training of the generative model. Furthermore, given the wide variety of infill criteria proposed in the literature, the infill criteria defined in \cref{eq:lcb} can be flexibly substituted with alternative strategies. This substitution naturally leads to different formulations of the optimization loss $\mathcal{L}_{\text{opt}}$, thereby allowing our framework to accommodate various exploration-exploitation trade-offs.
    
    \subsubsection{Training Procedure}
    
    The training procedure is summarized in \cref{alg:train}. It begins with training the surrogate model $\tilde{f}$ on the elite dataset $\mathcal{D}_{\text{SM}}$, after which the surrogate is frozen to ensure stable guidance throughout the optimization process. Subsequently, the paired generative models $\bm{\gamma}$ and $\bm{\gamma}'$ are optimized using mini-batches sampled from the paired dataset $\mathcal{D}_{\text{GM}}$, which is constructed during the data preparation phase (\cref{alg:data}).
    
    \algTrain
    
    In each training iteration (lines 4-10 of \cref{alg:train}), the model processes a paired sample $(\bm{x}_-, y_-)$ and $(\bm{x}_+, y_+)$ drawn from the \emph{inferior} and \emph{superior} subsets, respectively. The composite loss $\mathcal{L}$, as defined in \cref{eq:loss}, is evaluated and backpropagated to update the parameters of both generative models. Through this iterative process, the generative model $\bm{\gamma}$ progressively learns to map inferior solutions to superior ones, guided by the fitness landscape estimated by the surrogate model.
    
    \subsection{Population Generation}
    
    Similar to the reproduction mechanism employed in traditional EAs, the population generation phase in \MethodName{} aims to create new candidate solutions based on the current population. However, instead of relying on handcrafted heuristic operators, \MethodName{} utilizes the learned generative model $\bm{\gamma}$ for solution generation.
    
    \algGenerate
    
    As illustrated in \cref{alg:generate}, this phase employs the trained generative model $\bm{\gamma}(\cdot; \bm{\phi}^*)$ to transform \emph{inferior} solutions into \emph{superior} ones, thereby generating new candidate solutions from the current population $\mathcal{D}_t$. Each generated solution $\bm{x}^{t+1}_i$ is subsequently evaluated using the true objective function $f$ to obtain its fitness value $y^{t+1}_i$, forming the next-generation population $\mathcal{D}_{t+1}$. Due to the inherently parallelizable nature of both the generative model and the objective function $f$, this entire population generation and evaluation step can be fully parallelized, which substantially improves computational efficiency.

    \section{Experiments}\label{sec:experiment}
    
    This section presents a comprehensive evaluation of \MethodName across diverse complex optimization tasks, including low- and high-dimensional numerical benchmarks, classical control problems, and complex high-dimensional robot control tasks. The experiments are rigorously designed to assess the effectiveness, efficiency, and scalability of our proposed data-driven evolutionary framework.
    Specifically, we aim to address the following research questions: \\
    \color{black}\textbf{RQ1}: How does \MethodName perform in terms of solution quality and convergence speed compared to representative baselines?\\\color{black}
    \textbf{RQ2}: What is the individual contribution of each major component in \MethodName to the overall performance?\\
    \textbf{RQ3}: How does the optimization process of \MethodName evolve over time, and what can be observed from its search behavior through visual analysis?
    
    To systematically address these research questions, we validate the proposed framework through three cohesive groups of experiments: \color{black}(1) performance and efficiency comparisons across complex black-box optimization problems with varying dimensionalities;\color{black} (2) ablation studies to analyze the impact of different components and hyperparameters; and (3) visualizations to reveal the behavioral characteristics of the learned optimization process. These studies collectively demonstrate the scalability, adaptability, and data-efficiency of \MethodName in solving complex optimization problems without reliance on heuristic operations.
    
    \subsection{Experimental Setup}
    
    Our empirical evaluation covers three categories of optimization challenges: (i) low- and high-dimensional numerical functions, (ii) low- and high-dimensional classical control tasks, and (iii) robotic control tasks. The classical control tasks \emph{Pushing} and \emph{Rover} are taken from \cite{Wang2018}, while \emph{Landing}, \emph{Walker}, and \emph{Ant} are adapted from the Gym library \cite{Towers2023}. The maximum number of fitness evaluations (FEs) is set to 1,000 for low-dimensional tasks and 10,000 for high-dimensional tasks.
    
    \MethodName is compared against a comprehensive set of baseline methods, including:  
    (i) Bayesian optimization algorithms, covering both sequential and batch variants, such as Hyperband~\cite{Lisha2018}, TuRBO~\cite{Eriksson2019}, \textcolor{black}{SCBO~\cite{eriksson2021scalable}}, BOGP (LogEI)~\cite{Ament2023}, and the Tree structured Parzen Estimator (TPE)~\cite{Bergstra2011}, which is widely used in AutoML;  
    (ii) evolutionary strategies, represented by CMA-ES~\cite{Hansen1996};  
    (iii) Nelder Mead~\cite{Nelder1965}, a classical heuristic method;  
    \textcolor{black}{(iv) SADE-SS~\cite{zhou2025surrogate} and SA-LSEO-LE~\cite{zhao2025surrogate}, two state of the art surrogate assisted evolutionary algorithms.}

    All algorithms are evaluated over 20 independent runs with different random seeds on each benchmark.  
    \textcolor{black}{Additional details regarding the experimental environment, parameter configurations, and parallel execution settings are provided in Supplementary Document C, while ablation studies on key hyperparameters are included in Supplementary Document F.}

    \figResultAll
    
    \subsection{Convergence Performance across Benchmarks}
    
    To address our first research question (RQ1), we evaluate the performance of \MethodName against a comprehensive set of representative baselines.  
    This critical set of experiments is designed to rigorously assess the final solution quality and convergence speed across diverse tasks.  
    \textcolor{black}{Specifically, by testing on 25 benchmarks spanning numerical functions, classical control, and robot control, with dimensions ranging from 10D to 1000D,} we aim to demonstrate the broad applicability and effectiveness of the method across various dimensionalities.
    
    \tabEvoGOResults
    
    \color{black}
    The aggregated results, a representative subset of which is shown in \cref{fig:result-all} (with full details in Supplementary Document E), demonstrate that \MethodName consistently outperforms the baselines across most domains.  
    A total of 25 benchmark tasks are evaluated, including four 1000-dimensional numerical problems listed in Table~\ref{tab:evogo_1000d}. Due to the long runtime of high-dimensional cases, one representative comparison is provided for each algorithm. Overall, on low-dimensional problems, our method outperforms all baselines except the state-of-the-art SAEA algorithm SADE-SS, which can better exploit limited data in small-sample settings. However, for high-dimensional problems (e.g., 1000D), SADE-SS requires over one day of computation for a single run, making it impractical for large-scale applications. As shown in Table~\ref{tab:evogo_1000d}, our method surpasses SA-LSEO-LE (state-of-the-art high-dimensional SAEA variants), indicating its superior capability to learn underlying regularities and distributions in complex black-box optimization tasks.

    \color{black}
    
    We further evaluate \MethodName on the Hopper environment provided by Brax~\cite{Brax2021}, a high-dimensional robotic control benchmark that supports massively parallel simulation via GPU acceleration. As shown in \cref{fig:brax}, \MethodName achieves the highest final reward among all compared methods, including TPE, CMA-ES, and PPO~\cite{Schulman2017}, with respective population sizes of 1000, 1, 100, and 64. Notably, \MethodName attains comparable or even higher performance than PPO, despite the latter being an online reinforcement learning algorithm that continuously updates its policy through environment interaction. Furthermore, \MethodName exhibits substantial optimization efficiency: within a time budget of only 500 seconds, it surpasses all baselines, including those that continue training for over an hour. These results underscore the capability of \MethodName to deliver rapid and high-quality solutions in complex, GPU accelerated continuous control environments.
    
    \figBrax
    
    \figDirect
    
    \subsection{Convergence Speed across Benchmarks}
    
    Further addressing our first research question (RQ1), this section focuses specifically on the convergence speed and efficiency of \MethodName. To this end, our experiments are designed from two angles: first, we measure the number of generations required for convergence on large-scale tasks, and second, we evaluate the wall-clock efficiency within a runtime-constrained environment.
    
    First, as summarized in \cref{tab:iters}, \MethodName converges within 10 generations across all large-scale benchmark tasks. In contrast, baseline methods often require tens to thousands of generations to achieve comparable performance. This property is crucial for scenarios with tight runtime budgets but abundant parallel resources, where large populations can be evaluated concurrently.
    
    \tabIters
    
    Second, the runtime constrained setting of Brax allows us to evaluate the wall clock efficiency of each method under the same budget of 10,000 function evaluations and one hour of runtime. As shown in \cref{fig:brax}, \MethodName rapidly attains a reward of 900 within the first 100 seconds, while other methods require substantially more time to reach comparable performance, and some never achieve it. In particular, \MethodName delivers up to $134\times$ higher reward per second than CMA ES and $50\times$ higher than TPE, reflecting its superior optimization throughput. These findings highlight the robust convergence rate of \MethodName when utilizing large population sizes and parallel evaluations.
    
    \subsection{Ablation Studies}\label{sec:ablation}
    
    To answer our second research question (RQ2), we conduct a comprehensive ablation study to dissect \MethodName and quantify the contribution of its key components. These experiments are designed to isolate the impact of elements such as the surrogate model, infill criteria, and generative architecture on overall performance.
    
    \figAblation
    
    \color{black}
    
    \subsubsection{Under Perfect Surrogate Model}
    To examine the performance limit of \MethodName, we analyze its behavior when the surrogate model is assumed to be perfectly accurate.
    In this setting, the predicted objective values in the optimization loss are replaced with the ground-truth function evaluations, forming a variant referred to as \emph{Real Eval}.
    This experiment isolates the effectiveness of the overall generative architecture and optimization strategy from any surrogate-induced error, thereby assessing the theoretical upper bound of the framework.
    As shown in \cref{fig:direct}, the \emph{Real Eval} configuration achieves robust and consistent results across different numerical benchmarks, demonstrating the full potential of the proposed composite generative structure and loss formulation.
    These results indicate that \MethodName~can achieve even higher performance when the surrogate model approaches perfect accuracy.
    
    \subsubsection{Under Failed Surrogate Model}
    To simulate the scenario where the surrogate model completely fails to provide reliable guidance, we evaluated two variants of \MethodName~that remove or replace the surrogate component.
    The first variant, denoted as \emph{No Surrogate}, eliminates the surrogate model and its corresponding optimization loss, relying solely on the generative mechanism for search guidance.
    The second variant, referred to as \emph{CycleGAN}, replaces the surrogate-based optimization loss with a standard adversarial loss from the CycleGAN framework \cite{Zhu2017}.
    These two configurations effectively emulate conditions in which the surrogate model is either absent or incapable of meaningful prediction.
    As shown in \cref{fig:ablation}, both variants exhibited noticeable performance degradation and increased instability, indicating that the surrogate model plays a crucial role in maintaining effective exploration-exploitation balance.
    Nevertheless, the generative component still preserved certain diversity and structural stability, largely due to the reconstruction constraint that prevents severe mode collapse.
    These observations confirm that while \MethodName~can remain functional under surrogate failure, accurate surrogate feedback is vital for achieving its full performance potential.
    
    \subsubsection{Alternative Surrogate Model}
    To examine whether the proposed framework depends exclusively on GP surrogates, we conduct two complementary studies that evaluate both model substitution and infill criteria.
    These analyses reveal how \MethodName~behaves when the GP-based probabilistic assumptions are replaced by alternative designs.
    
    \textbf{MLP as surrogate model.}
    To evaluate the modularity of the surrogate component, we replaced the default GP with a Multi-Layer Perceptron (MLP) \cite{Haykin1994}, forming the variant \emph{\MethodName-MLP}.
    Since the MLP does not provide inherent uncertainty estimation, we removed the standard deviation and covariance terms from the optimization loss (Eq.~\ref{eq:loss-opt}).
    This configuration isolates the contribution of the probabilistic structure introduced by the GP.
    As shown in \cref{fig:ablation}, the \emph{\MethodName-MLP} variant showed slightly weaker performance on several numerical benchmarks, likely attributable to the absence of variance-guided exploration, while achieving strong results in complex scenarios such as the Ant benchmark.
    These results demonstrate that \MethodName~remains effective even when the surrogate does not include explicit uncertainty modeling, highlighting the adaptability of the proposed framework.
    
    \textbf{infill criteria as replacement.}
    We investigated the impact of the infill criteria by examining whether the proposed mechanism for balancing exploration and exploitation could be replaced by an alternative criterion.
    Specifically, we substituted the proposed optimization loss (Eq.~\ref{eq:loss-opt}) with the classical Lower/Upper Confidence Bound (LCB/UCB) criterion, thereby establishing the variant designated as \emph{\MethodName-LCB}.
    The corresponding loss function is defined as
    
    \begin{equation}
        \mathcal{L}_{\text{opt}}^{\text{LCB}} = \mathbb{E} \left[ \tilde{f}_t(\bm{\gamma}(\bm{p})) \right] - y_{\bm{p}} + \lambda_2 \mathbb{D} \left[ \bm{\gamma}(\bm{p}) \right]^{1/2},
        \label{eq:loss-opt-lcb}
    \end{equation}
    
    The derivation for the LCB infill criteria is provided in Supplementary Document~B. As illustrated in \cref{fig:ablation}, the ``\MethodName-LCB'' variant achieved competitive performance but generally performed worse than the original formulation. This observation indicates that, while the LCB/UCB-based infill criteria maintains reasonable performance, the tailored optimization loss employed in \MethodName~provides a more stable exploration mechanism across diverse problem settings.
    
    \figVisToy
    
    \subsubsection{Paired Architecture Analysis}  
    We further analyze the structural importance of the paired generative architecture, which is a fundamental component of \MethodName.  
    To this end, we design a ``Single Net'' variant that removes one generator and all consistency-enforcing mechanisms between $\bm{\gamma}$ and $\bm{\gamma}'$.  
    Specifically, this variant eliminates the reconstruction loss $\mathcal{L}_{\text{rec}}$ (Eq.~\ref{eq:loss-rec}) and one branch of the similarity loss $\mathcal{L}_{\text{sim}}$ (Eq.~\ref{eq:loss-sim}) in \cref{eq:loss-gen}, while omitting the standard deviation regularization term $\mathcal{L}_{\text{std}}$ in \cref{eq:loss-opt}.  
    This ablation assesses whether a single-generator design can sustain the stability and diversity achieved through paired modeling.  
    As presented in \cref{fig:ablation}, the ``Single Net'' configuration leads to a noticeable degradation in convergence stability and diversity maintenance, particularly under complex multimodal settings.  
    These findings demonstrate that the paired architecture, reinforced by the reconstruction constraint, is essential for ensuring balanced training dynamics, preventing mode collapse, and enhancing the robustness of the overall generative process.
    
    \color{black}
    \subsubsection{Extended Ablations}
    Additional ablation studies are detailed in Supplementary Document F. These studies examined variations in the neural network architectures and hyperparameter configurations, further elucidating the robustness and adaptability of \MethodName across various settings.
    Specifically, we explored the impact of different generative model layers and their width. We observed that the generative model's core architecture has little effect on \MethodName's performance, as shown in Fig. S4.
    Moreover, we also investigated the impact of the reconstruction loss scale $\lambda_1$, the data split factor $\eta$, and the sliding window factor $\epsilon$. These components were all found to have small or negligible impact on the performance of \MethodName, as shown in Figs. S5 to S7.
    
    \subsection{Search Dynamics}\label{sec:vis}
    To answer our third research question (RQ3) and gain deeper insights into the search dynamics of \MethodName, we conducted a visualization study on a two-dimensional Ackley function with a population size of 100, as shown in \cref{fig:vis-toy}. Specifically, we tracked the transitions from input solutions $\bm{x}_{\text{in}}$ to generated output solutions $\bm{x}_{\text{out}} = \bm{\gamma}(\bm{x}_{\text{in}})$, captured across various generations from the trained generative model $\bm{\gamma}$. Arrows in the visualization represent vectors from $\bm{x}_{\text{in}}$ to $\bm{x}_{\text{out}}$, and their colors reflect their norm-2 lengths, as described in the corresponding plot legends. This visual representation demonstrates \MethodName's capacity to precisely estimate both the direction $(\bm{x}_{\text{out}} - \bm{x}_{\text{in}}) / \|\bm{x}_{\text{out}} - \bm{x}_{\text{in}}\|$ and the magnitude $\|\bm{x}_{\text{out}} - \bm{x}_{\text{in}}\|$ of steps toward potential optima. This capability not only enhances optimization efficiency but also significantly mitigates mode collapse. Furthermore, the learning progress of $\bm{\gamma}$ is illustrated through the grayscale contours of the optimization landscapes captured by the solutions at different generations. Throughout the search process, $\bm{\gamma}$ adaptively hones in on specific regions, thereby demonstrating a refined focus that enhances its search efficacy. This adaptiveness suggests \MethodName's potential for achieving superlinear convergence given ample data, which is accentuated by the dashed rectangles that delineate the boundaries of solutions generated at different generations.
    
    \section{Conclusion}\label{sec:conclusion}
    
    This paper presented \FullMethodName, a data-driven evolutionary optimization framework. This framework was specifically designed to address three critical limitations inherent in existing generative model-based evolutionary algorithms: reliance on handcrafted heuristics, objective misalignment between generative models and evolutionary goals, and underutilization of available data. By eliminating traditional search operators and adopting globally consistent training objectives, \MethodName enables robust end-to-end learning for optimization, offering enhanced generality, objective alignment, and data efficiency.
    
    The proposed framework is underpinned by a paired generative architecture that integrates forward and inverse models to produce high-fidelity candidate solutions, thereby effectively replacing conventional crossover and mutation operators. To enhance training diversity under limited evaluation budgets, we employed a pairwise data construction strategy. Furthermore, a composite loss function integrates reconstruction, distributional alignment, and directional guidance to ensure efficient model convergence toward promising objective regions.
    
    Extensive experiments conducted across numerical, control, and robotic tasks demonstrated the superior performance of \MethodName when compared against state-of-the-art baselines. Specifically, \MethodName consistently achieved convergence within 10 generations. This performance, combined with GPU-parallelized inference, delivered up to a $134\times$ speedup on computationally complex optimization tasks.
    
    These results underscore the substantial potential of fully data-driven evolutionary frameworks to generalize effectively across diverse domains without requiring manual heuristic design. Future work will initially focus on enhancing robustness under extremely low data regimes and providing theoretical insights into the framework's convergence behavior and optimization dynamics. \textcolor{black}{In addition, future extensions will explore offline pre-training and cross-task transfer to further improve data efficiency in settings where online evaluation is limited.}

\bibliographystyle{IEEEtran}
\bibliography{refs.bib}

@book{Rasmussen2005,
  author    = {Rasmussen, Carl Edward and Williams, Christopher K. I.},
  publisher = {The MIT Press},
  title     = {Gaussian processes for machine learning},
  year      = {2005},
  month     = {11}
}

@article{Frazier2009,
  author  = {Frazier, Peter and Powell, Warren and Dayanik, Savas},
  journal = {INFORMS Journal on Computing},
  title   = {The knowledge-gradient policy for correlated normal beliefs},
  year    = {2009},
  number  = {4},
  pages   = {599-613},
  volume  = {21}
}

@article{Jones1998,
  author    = {Jones, Donald R. and Schonlau, Matthias and Welch, William J.},
  journal   = {Journal of Global Optimization},
  title     = {Efficient global optimization of expensive black-box functions},
  year      = {1998},
  month     = dec,
  number    = {4},
  pages     = {455–492},
  volume    = {13},
  address   = {USA},
  numpages  = {38},
  publisher = {Kluwer Academic Publishers}
}

@misc{Frazier2018,
  author        = {Peter I. Frazier},
  title         = {A tutorial on {B}ayesian optimization},
  year          = {2018},
  archiveprefix = {arXiv},
  eprint        = {1807.02811},
  primaryclass  = {stat.ML}
}

@article{Sherman1950,
  author    = {Jack Sherman and Winifred J. Morrison},
  journal   = {The Annals of Mathematical Statistics},
  title     = {Adjustment of an inverse matrix corresponding to a change in one element of a given matrix},
  year      = {1950},
  number    = {1},
  pages     = {124 -- 127},
  volume    = {21},
  publisher = {Institute of Mathematical Statistics}
}

@inproceedings{Chen2023,
  author    = {Chen, Lesi and Xu, Jing and Luo, Luo},
  booktitle = {International Conference on Machine Learning},
  title     = {Faster gradient-free algorithms for nonsmooth nonconvex stochastic optimization},
  year      = {2023},
  pages     = {5219--5233},
  publisher = {PMLR},
  volume    = {202}
}

@inproceedings{Zhai2022,
  author    = {Zhai, Yaoguang and Gao, Sicun},
  booktitle = {Advances in Neural Information Processing Systems},
  title     = {Monte {C}arlo tree descent for black-box optimization},
  year      = {2022},
  pages     = {12581--12593},
  publisher = {Curran Associates, Inc.},
  volume    = {35}
}

@inproceedings{Ament2023,
  author    = {Sebastian Ament and Sam Daulton and David Eriksson and Maximilian Balandat and Eytan Bakshy},
  booktitle = {Advances in Neural Information Processing Systems},
  title     = {Unexpected improvements to expected improvement for {B}ayesian optimization},
  year      = {2023}
}

@inproceedings{Eriksson2019,
  author    = {Eriksson, David and Pearce, Michael and Gardner, Jacob and Turner, Ryan D. and Poloczek, Matthias},
  booktitle = {Advances in Neural Information Processing Systems},
  title     = {Scalable global optimization via local {B}ayesian optimization},
  year      = {2019},
  volume    = {32}
}

@inproceedings{Zhu2017,
  author    = {Zhu, Jun-Yan and Park, Taesung and Isola, Phillip and Efros, Alexei A.},
  booktitle = {Proceedings of the IEEE International Conference on Computer Vision},
  title     = {Unpaired image-to-image translation using cycle-consistent adversarial networks},
  year      = {2017},
  pages     = {2223--2232}
}

@inproceedings{Wang2018,
  author    = {Wang, Zi and Gehring, Clement and Kohli, Pushmeet and Jegelka, Stefanie},
  booktitle = {Proceedings of the International Conference on Artificial Intelligence and Statistics},
  title     = {Batched large-scale {B}ayesian optimization in high-dimensional spaces},
  year      = {2018},
  pages     = {745--754},
  publisher = {PMLR},
  volume    = {84}
}

@misc{Towers2023,
  author    = {Towers, Mark and Terry, Jordan K. and Kwiatkowski, Ariel and Balis, John U. and Cola, Gianluca de and Deleu, Tristan and Goulão, Manuel and Kallinteris, Andreas and Arjun, K. G. and Krimmel, Markus and Perez-Vicente, Rodrigo and Pierré, Andrea and Schulhoff, Sander and Tai, Jun Jet and Shen, Andrew Tan Jin and Younis, Omar G.},
  month     = mar,
  title     = {Gymnasium},
  year      = {2023},
  publisher = {Zenodo}
}

@inproceedings{Kennedy1995,
  author    = {Kennedy, J. and Eberhart, R.},
  booktitle = {Proceedings of the International Conference on Neural Networks},
  title     = {Particle swarm optimization},
  year      = {1995},
  pages     = {1942-1948},
  volume    = {4}
}

@article{Nelder1965,
  author    = {Nelder, John A. and Mead, Roger},
  journal   = {The Computer Journal},
  title     = {A simplex method for function minimization},
  year      = {1965},
  number    = {4},
  pages     = {308--313},
  volume    = {7},
  publisher = {The British Computer Society}
}

@inproceedings{Kingma2015,
  author    = {Diederik P. Kingma and Jimmy Ba},
  booktitle = {International Conference on Learning Representations},
  title     = {Adam: A method for stochastic optimization},
  year      = {2015}
}

@article{Goodfellow2014,
  author   = {Goodfellow, Ian J. and Pouget-Abadie, Jean and Mirza, Mehdi and Xu, Bing and Warde-Farley, David and Ozair, Sherjil and Courville, Aaron and Bengio, Yoshua},
  journal  = {Advances in Neural Information Processing Systems},
  title    = {Generative adversarial nets},
  year     = {2014},
  pages    = {2672--2680},
  language = {english}
}

@book{Audet2017,
  author    = {Audet, Charles and Hare, Warren},
  publisher = {Springer},
  title     = {Derivative-free and blackbox optimization},
  year      = {2017}
}

@incollection{Bajaj2021,
  author    = {Bajaj, Ishan and Arora, Akhil and Hasan, M. M. Faruque},
  booktitle = {Black Box Optimization, Machine Learning, and No-Free Lunch Theorems},
  publisher = {Springer},
  title     = {Black-box optimization: Methods and applications},
  year      = {2021},
  pages     = {35--65}
}

@article{Abdolrasol2021,
  author         = {Abdolrasol, Maher G. M. and Hussain, S. M. Suhail and Ustun, Taha Selim and Sarker, Mahidur R. and Hannan, Mahammad A. and Mohamed, Ramizi and Ali, Jamal Abd and Mekhilef, Saad and Milad, Abdalrhman},
  journal        = {Electronics},
  title          = {Artificial neural networks based optimization techniques: A review},
  year           = {2021},
  number         = {21},
  volume         = {10},
  article-number = {2689}
}

@inproceedings{Kumar2020,
  author    = {Kumar, Aviral and Levine, Sergey},
  booktitle = {Advances in Neural Information Processing Systems},
  title     = {Model inversion networks for model-based optimization},
  year      = {2020},
  pages     = {5126--5137},
  volume    = {33}
}

@inproceedings{Andrychowicz2016,
  author    = {Andrychowicz, Marcin and Denil, Misha and Colmenarejo, Sergio G{\'o}mez and Hoffman, Matthew W. and Pfau, David and Schaul, Tom and Shillingford, Brendan and de Freitas, Nando},
  booktitle = {Advances in Neural Information Processing Systems},
  title     = {Learning to learn by gradient descent by gradient descent},
  year      = {2016},
  pages     = {3988--3996}
}

@inproceedings{Chen2017,
  author       = {Chen, Yutian and Hoffman, Matthew W. and Colmenarejo, Sergio G{\'o}mez and Denil, Misha and Lillicrap, Timothy P. and Botvinick, Matt and Freitas, Nando},
  booktitle    = {International Conference on Machine Learning},
  title        = {Learning to learn without gradient descent by gradient descent},
  year         = {2017},
  organization = {PMLR},
  pages        = {748--756}
}

@inproceedings{Srinivas2010,
  author    = {Srinivas, Niranjan and Krause, Andreas and Kakade, Sham and Seeger, Matthias},
  booktitle = {International Conference on Machine Learning},
  title     = {Gaussian process optimization in the bandit setting: No regret and experimental design},
  year      = {2010},
  pages     = {1015--1022}
}

@inproceedings{Snoek2012,
  author    = {Snoek, Jasper and Larochelle, Hugo and Adams, Ryan P.},
  booktitle = {Advances in Neural Information Processing Systems},
  title     = {Practical {B}ayesian optimization of machine learning algorithms},
  year      = {2012},
  pages     = {2951--2959},
  volumn    = {2}
}

@inproceedings{Riquelme2018,
  author    = {Riquelme, Carlos and Tucker, George and Snoek, Jasper},
  booktitle = {International Conference on Learning Representations},
  title     = {Deep {B}ayesian bandits showdown: An empirical comparison of {B}ayesian deep networks for {T}hompson sampling},
  year      = {2018}
}

@inproceedings{Gardner2017,
  author       = {Gardner, Jacob and Guo, Chuan and Weinberger, Kilian and Garnett, Roman and Grosse, Roger},
  booktitle    = {Artificial Intelligence and Statistics},
  title        = {Discovering and exploiting additive structure for {B}ayesian optimization},
  year         = {2017},
  organization = {PMLR},
  pages        = {1311--1319}
}

@inproceedings{Mutnỳ2018,
  author    = {Mutn{\`y}, Mojm{\'\i}r and Krause, Andreas},
  booktitle = {Advances in Neural Information Processing Systems},
  title     = {Efficient high dimensional {B}ayesian optimization with additivity and quadrature fourier features},
  year      = {2018},
  pages     = {9019--9030}
}

@book{Holland1992,
  author    = {Holland, John H.},
  publisher = {{MIT} Press},
  title     = {Adaptation in natural and artificial systems: An introductory analysis with applications to biology, control, and artificial intelligence},
  year      = {1992}
}

@book{Koza1994,
  author    = {Koza, John R.},
  publisher = {MIT Press},
  title     = {Genetic programming {II}},
  year      = {1994}
}

@article{Storn1997,
  author    = {Storn, Rainer and Price, Kenneth},
  journal   = {Journal of Global Optimization},
  title     = {Differential evolution--a simple and efficient heuristic for global optimization over continuous spaces},
  year      = {1997},
  pages     = {341--359},
  volume    = {11},
  publisher = {Springer}
}

@software{Brax2021,
  author  = {C. Daniel Freeman and Erik Frey and Anton Raichuk and Sertan Girgin and Igor Mordatch and Olivier Bachem},
  title   = {Brax - A Differentiable Physics Engine for Large Scale Rigid Body Simulation},
  url     = {http://github.com/google/brax},
  version = {0.9.8},
  year    = {2021}
}

@inproceedings{Kingma2014a,
  author    = {Kingma, Diederik P. and Welling, Max},
  booktitle = {Proceedings of the International Conference on Learning Representations},
  title     = {Auto-encoding variational {B}ayes},
  year      = {2014},
  pages     = {1},
  volume    = {1050}
}

@article{Mirza2014,
  author  = {Mirza, Mehdi and Osindero, Simon},
  journal = {arXiv preprint arXiv:1411.1784},
  title   = {Conditional generative adversarial nets},
  year    = {2014}
}

@article{Sun2024,
  author  = {Sun, Kebin and Wang, Weituo and Cheng, Ran and Liang, Yu and Xie, Hairun and Wang, Jing and Zhang, Miao},
  journal = {Complex \& Intelligent Systems},
  title   = {Evolutionary generative design of supercritical airfoils: {A}n automated approach driven by small data},
  year    = {2024},
  month   = feb,
  pages   = {1167-1183},
  volume  = {10}
}

@inproceedings{Frazier2007,
  author    = {Frazier, Peter and Powell, Warren},
  booktitle = {IEEE International Symposium on Approximate Dynamic Programming and Reinforcement Learning},
  title     = {The knowledge gradient policy for offline learning with independent normal rewards},
  year      = {2007},
  pages     = {143--150}
}

@article{Liu2023,
  author  = {Liu, Songbai and Lin, Qiuzhen and Li, Jianqiang and Tan, Kay Chen},
  journal = {IEEE Transactions on Evolutionary Computation},
  title   = {A survey on learnable evolutionary algorithms for scalable multiobjective optimization},
  year    = {2023},
  number  = {6},
  pages   = {1941-1961},
  volume  = {27}
}

@Article{Liang2023a,
  author  = {Liang, Zhengping and Zhu, Yingmiao and Wang, Xiyu and Li, Zhi and Zhu, Zexuan},
  journal = {IEEE Transactions on Evolutionary Computation},
  title   = {Evolutionary multitasking for optimization based on generative strategies},
  year    = {2023},
  number  = {4},
  pages   = {1042-1056},
  volume  = {27},
}

@article{Zhang2023,
  author  = {Zhang, Yabin and Lian, Hairong and Yang, Guang and Zhao, Suyun and Ni, Peng and Chen, Hong and Li, Cuiping},
  journal = {IEEE Transactions on Cybernetics},
  title   = {Inaccurate-supervised learning with generative adversarial nets},
  year    = {2023},
  number  = {3},
  pages   = {1522-1536},
  volume  = {53}
}

@inproceedings{Ganin2015,
  author    = {Ganin, Yaroslav and Lempitsky, Victor},
  booktitle = {International Conference on Machine Learning},
  title     = {Unsupervised domain adaptation by backpropagation},
  year      = {2015},
  month     = jul,
  pages     = {1180--1189},
  publisher = {PMLR},
  volume    = {37}
}

@inproceedings{Liu2016,
  author    = {Liu, Ming-Yu and Tuzel, Oncel},
  booktitle = {Advances in Neural Information Processing Systems},
  title     = {Coupled generative adversarial networks},
  year      = {2016},
  publisher = {Curran Associates, Inc.},
  volume    = {29}
}

@inproceedings{Tzeng2017,
  author    = {Tzeng, Eric and Hoffman, Judy and Saenko, Kate and Darrell, Trevor},
  booktitle = {IEEE Conference on Computer Vision and Pattern Recognition},
  title     = {Adversarial discriminative domain adaptation},
  year      = {2017},
  pages     = {2962-2971}
}

@inproceedings{Ghifary2016,
  author    = {Muhammad Ghifary and W. Bastiaan Kleijn and Mengjie Zhang and David Balduzzi and Wen Li},
  booktitle = {The European Conference on Computer Vision},
  title     = {Deep reconstruction-classification networks for unsupervised domain adaptation},
  year      = {2016},
  pages     = {597--613},
  publisher = {Springer},
  volume    = {9908}
}

@inbook{Csurka2017,
  author    = {Csurka, Gabriela},
  pages     = {1--35},
  publisher = {Springer International Publishing},
  title     = {A Comprehensive Survey on Domain Adaptation for Visual Applications},
  year      = {2017},
  booktitle = {Domain Adaptation in Computer Vision Applications}
}

@article{Thompson1933,
  author  = {William R. Thompson},
  journal = {Biometrika},
  title   = {On the likelihood that one unknown probability exceeds another in view of the evidence of two samples},
  year    = {1933},
  pages   = {285-294},
  volume  = {25}
}

@inproceedings{Azimi2010,
  author    = {Azimi, Javad and Fern, Alan and Fern, Xiaoli},
  booktitle = {Advances in Neural Information Processing Systems},
  title     = {Batch {B}ayesian optimization via simulation matching},
  year      = {2010},
  publisher = {Curran Associates, Inc.},
  volume    = {23}
}

@inproceedings{DeAth2021,
  author       = {De Ath, George and Everson, Richard M. and Fieldsend, Jonathan E.},
  booktitle    = {Uncertainty in Artificial Intelligence},
  title        = {Asynchronous $\epsilon$-greedy {B}ayesian optimisation},
  year         = {2021},
  organization = {PMLR},
  pages        = {578--588}
}

@inproceedings{Nguyen2020,
  author    = {Nguyen, Dang and Gupta, Sunil and Rana, Santu and Shilton, Alistair and Venkatesh, Svetha},
  booktitle = {Proceedings of the AAAI Conference on Artificial Intelligence},
  title     = {{B}ayesian optimization for categorical and category-specific continuous inputs},
  year      = {2020},
  pages     = {5256--5263},
  volume    = {34}
}

@article{KonakovicLukovic2020,
  author  = {Konakovic Lukovic, Mina and Tian, Yunsheng and Matusik, Wojciech},
  journal = {Advances in Neural Information Processing Systems},
  title   = {Diversity-guided multi-objective {B}ayesian optimization with batch evaluations},
  year    = {2020},
  pages   = {17708--17720},
  volume  = {33}
}

@article{Wang2023,
  author    = {Wang, Xilu and Jin, Yaochu and Schmitt, Sebastian and Olhofer, Markus},
  journal   = {ACM Computing Surveys},
  title     = {Recent advances in {B}ayesian optimization},
  year      = {2023},
  number    = {13s},
  pages     = {1--36},
  volume    = {55},
  publisher = {ACM New York, NY}
}

@inproceedings{Eriksson2021,
  author       = {Eriksson, David and Poloczek, Matthias},
  booktitle    = {International Conference on Artificial Intelligence and Statistics},
  title        = {Scalable constrained {B}ayesian optimization},
  year         = {2021},
  organization = {PMLR},
  pages        = {730--738}
}

@article{Daulton2021,
  author  = {Daulton, Samuel and Balandat, Maximilian and Bakshy, Eytan},
  journal = {Advances in Neural Information Processing Systems},
  title   = {Parallel {B}ayesian optimization of multiple noisy objectives with expected hypervolume improvement},
  year    = {2021},
  pages   = {2187--2200},
  volume  = {34}
}

@inproceedings{Tiao2021,
  author       = {Tiao, Louis C. and Klein, Aaron and Seeger, Matthias W. and Bonilla, Edwin V. and Archambeau, Cedric and Ramos, Fabio},
  booktitle    = {International Conference on Machine Learning},
  title        = {{BORE}: {B}ayesian optimization by density-ratio estimation},
  year         = {2021},
  organization = {PMLR},
  pages        = {10289--10300}
}

@inproceedings{Swersky2020,
  author       = {Swersky, Kevin and Rubanova, Yulia and Dohan, David and Murphy, Kevin},
  booktitle    = {Conference on Uncertainty in Artificial Intelligence},
  title        = {Amortized bayesian optimization over discrete spaces},
  year         = {2020},
  organization = {PMLR},
  pages        = {769--778}
}

@inproceedings{Oliveira2022,
  author    = {Oliveira, Rafael and Tiao, Louis C. and Ramos, Fabio},
  booktitle = {Proceedings of the 36th International Conference on Neural Information Processing Systems},
  title     = {Batch {B}ayesian optimisation via density-ratio estimation with guarantees},
  year      = {2022},
  pages     = {29816--29829}
}

@inproceedings{Bergstra2011,
  author    = {Bergstra, James and Bardenet, R{\'e}mi and Bengio, Yoshua and K{\'e}gl, Bal{\'a}zs},
  booktitle = {Proceedings of the 24th International Conference on Neural Information Processing Systems},
  title     = {Algorithms for hyper-parameter optimization},
  year      = {2011},
  pages     = {2546--2554}
}

@article{Kumagai2023,
  author    = {Kumagai, Wataru and Yasuda, Keiichiro},
  journal   = {Innovative Systems Approach for Facilitating Smarter World},
  title     = {Black-box optimization and its applications},
  year      = {2023},
  pages     = {81--100},
  publisher = {Springer}
}

@article{Lisha2018,
  author  = {Lisha Li and Kevin Jamieson and Giulia DeSalvo and Afshin Rostamizadeh and Ameet Talwalkar},
  title   = {Hyperband: A Novel Bandit-Based Approach to Hyperparameter Optimization},
  journal = {Journal of Machine Learning Research},
  year    = {2018},
  volume  = {18},
  number  = {185},
  pages   = {1--52}
}

@book{Haykin1994,
  title     = {Neural networks: A comprehensive foundation},
  author    = {Haykin, Simon},
  year      = {1994},
  publisher = {Prentice Hall PTR}
}

@article{Schulman2017,
  title={Proximal policy optimization algorithms},
  author={Schulman, John and Wolski, Filip and Dhariwal, Prafulla and Radford, Alec and Klimov, Oleg},
  journal={arXiv preprint arXiv:1707.06347},
  year={2017}
}

@Article{Li2020,
  author  = {Zhenhua Li and Xi Lin and Qingfu Zhang and Hailin Liu},
  journal = {Swarm and Evolutionary Computation},
  title   = {Evolution strategies for continuous optimization: A survey of the state-of-the-art},
  year    = {2020},
  issn    = {2210-6502},
  pages   = {100694},
  volume  = {56},
}

@Article{Larranaga2024,
  author  = {Larrañaga, Pedro and Bielza, Concha},
  journal = {IEEE Transactions on Evolutionary Computation},
  title   = {Estimation of distribution algorithms in machine learning: A survey},
  year    = {2024},
  number  = {5},
  pages   = {1301-1321},
  volume  = {28},
}

@Article{He2021,
  author  = {He, Cheng and Huang, Shihua and Cheng, Ran and Tan, Kay Chen and Jin, Yaochu},
  journal = {IEEE Transactions on Cybernetics},
  title   = {Evolutionary multiobjective optimization driven by generative adversarial networks ({GANs})},
  year    = {2021},
  number  = {6},
  pages   = {3129-3142},
  volume  = {51},
}

@Article{Jiang2025,
  author  = {Jiang, Yi and Zhan, Zhi-Hui and Chen Tan, Kay and Zhang, Jun},
  journal = {IEEE Transactions on Evolutionary Computation},
  title   = {Knowledge learning for evolutionary computation},
  year    = {2025},
  number  = {1},
  pages   = {16-30},
  volume  = {29},
}

@Article{Rechenberg1965,
  author   = {Rechenberg, Ingo},
  journal  = {Royal Aircraft Establishment, Library Translation},
  title    = {Cybernetic solution path of an experimental problem},
  year     = {1965},
  volume   = {1122},
}

@InProceedings{Hansen1996,
  author       = {Hansen, Nikolaus and Ostermeier, Andreas},
  booktitle    = {IEEE International Conference on Evolutionary Computation},
  title        = {Adapting arbitrary normal mutation distributions in evolution strategies: The covariance matrix adaptation},
  year         = {1996},
  organization = {IEEE},
  pages        = {312--317},
}

@Article{Hansen2001,
  author  = {Hansen, Nikolaus and Ostermeier, Andreas},
  journal = {Evolutionary Computation},
  title   = {Completely derandomized self-adaptation in evolution strategies},
  year    = {2001},
  month   = {06},
  number  = {2},
  pages   = {159-195},
  volume  = {9},
}

@article{salimans2017evolution,
  title={Evolution strategies as a scalable alternative to reinforcement learning},
  author={Salimans, Tim and Ho, Jonathan and Chen, Xi and Sidor, Szymon and Sutskever, Ilya},
  journal={arXiv preprint arXiv:1703.03864},
  year={2017}
}

@TechReport{Baluja1994,
  author    = {Baluja, Shummet},
  title     = {Population-based incremental learning: A method for integrating genetic search based function optimization and competitive learning},
  year      = {1994},
  publisher = {Carnegie Mellon University},
}

@Article{Muehlenbein1997,
  author   = {M\"{u}hlenbein, Heinz},
  journal  = {Evol. Comput.},
  title    = {The equation for response to selection and its use for prediction},
  year     = {1997},
  month    = sep,
  number   = {3},
  pages    = {303–346},
  volume   = {5},
  numpages = {44},
}

@InProceedings{Thierens2010,
  author    = {Thierens, Dirk},
  booktitle = {International Conference on Parallel Problem Solving from Nature},
  title     = {The linkage tree genetic algorithm},
  year      = {2010},
  pages     = {264–273},
  numpages  = {10},
}

@PhdThesis{Harik1997,
  author = {Harik, Georges Raif},
  school = {University of Michigan},
  title  = {Learning gene linkage to efficiently solve problems of bounded difficulty using genetic algorithms},
  year   = {1997},
}

@Article{Bull1999,
  author  = {Bull, Larry},
  journal = {Soft Computing},
  title   = {On model-based evolutionary computation},
  year    = {1999},
  pages   = {76--82},
  volume  = {3},
}

@Article{Jin2002,
  author  = {Jin, Yaochu and Olhofer, Markus and Sendhoff, Bernhard},
  journal = {IEEE Transactions on Evolutionary Computation},
  title   = {A framework for evolutionary optimization with approximate fitness functions},
  year    = {2002},
  number  = {5},
  pages   = {481--494},
  volume  = {6},
}

@Article{Jin2005,
  author  = {Jin, Yaochu},
  journal = {Soft Computing},
  title   = {A comprehensive survey of fitness approximation in evolutionary computation},
  year    = {2005},
  number  = {1},
  pages   = {3--12},
  volume  = {9},
}

@Article{He2023,
  author  = {Chunlin He and Yong Zhang and Dunwei Gong and Xinfang Ji},
  journal = {Expert Systems with Applications},
  title   = {A review of surrogate-assisted evolutionary algorithms for expensive optimization problems},
  year    = {2023},
  pages   = {119495},
  volume  = {217},
}

@Article{Liu2014,
  author  = {Liu, Bo and Zhang, Qingfu and Gielen, Georges G. E.},
  journal = {IEEE Transactions on Evolutionary Computation},
  title   = {A {G}aussian process surrogate model assisted evolutionary algorithm for medium scale expensive optimization problems},
  year    = {2014},
  number  = {2},
  pages   = {180-192},
  volume  = {18},
}

@InProceedings{Bandaru2010,
  author    = {Bandaru, Sunith and Deb, Kalyanmoy},
  booktitle = {IEEE Congress on Evolutionary Computation},
  title     = {Automated discovery of vital knowledge from pareto-optimal solutions: First results from engineering design},
  year      = {2010},
  pages     = {1-8},
}

@InProceedings{Gaur2017,
  author    = {Gaur, Abhinav and Deb, Kalyanmoy},
  booktitle = {2017 IEEE Congress on Evolutionary Computation (CEC)},
  title     = {Effect of size and order of variables in rules for multi-objective repair-based innovization procedure},
  year      = {2017},
  pages     = {2177-2184},
}

@Article{Liu2023a,
  author  = {Liu, Songbai and Li, Jun and Lin, Qiuzhen and Tian, Ye and Tan, Kay Chen},
  journal = {IEEE Transactions on Evolutionary Computation},
  title   = {Learning to accelerate evolutionary search for large-scale multiobjective optimization},
  year    = {2023},
  number  = {1},
  pages   = {67-81},
  volume  = {27},
}

@InProceedings{Zhang2014,
  author    = {Zhang, Hu and Song, Shenmin and Zhou, Aimin and Gao, Xiao-Zhi},
  booktitle = {2014 IEEE Congress on Evolutionary Computation (CEC)},
  title     = {A clustering based multiobjective evolutionary algorithm},
  year      = {2014},
  pages     = {723-730},
}

@Article{Pan2021,
  author  = {Pan, Linqiang and Li, Lianghao and Cheng, Ran and He, Cheng and Tan, Kay Chen},
  journal = {IEEE Transactions on Cybernetics},
  title   = {Manifold learning-inspired mating restriction for evolutionary multiobjective optimization with complicated pareto sets},
  year    = {2021},
  number  = {6},
  pages   = {3325-3337},
  volume  = {51},
}

@Article{Tian2020,
  author  = {Tian, Ye and Zhang, Xingyi and Wang, Chao and Jin, Yaochu},
  journal = {IEEE Transactions on Evolutionary Computation},
  title   = {An evolutionary algorithm for large-scale sparse multiobjective optimization problems},
  year    = {2020},
  number  = {2},
  pages   = {380-393},
  volume  = {24},
}

@Article{Tian2023,
  author  = {Tian, Ye and Li, Xiaopeng and Ma, Haiping and Zhang, Xingyi and Tan, Kay Chen and Jin, Yaochu},
  journal = {IEEE Transactions on Emerging Topics in Computational Intelligence},
  title   = {Deep reinforcement learning based adaptive operator selection for evolutionary multi-objective optimization},
  year    = {2023},
  number  = {4},
  pages   = {1051-1064},
  volume  = {7},
}

@InProceedings{Eiben2007,
  author    = {Eiben, A. E. and Horvath, Mark and Kowalczyk, Wojtek and Schut, Martijn C.},
  booktitle = {Engineering Self-Organising Systems},
  title     = {Reinforcement learning for online control of evolutionary algorithms},
  year      = {2007},
  pages     = {151--160},
}

@Article{Wang2022,
  author  = {Wang, Jing-Jing and Wang, Ling},
  journal = {IEEE Transactions on Evolutionary Computation},
  title   = {A cooperative memetic algorithm with learning-based agent for energy-aware distributed hybrid flow-shop scheduling},
  year    = {2022},
  number  = {3},
  pages   = {461-475},
  volume  = {26},
}

@Article{Li2024,
  author  = {Li, Pengyi and Hao, Jianye and Tang, Hongyao and Fu, Xian and Zhen, Yan and Tang, Ke},
  journal = {IEEE Transactions on Evolutionary Computation},
  title   = {Bridging evolutionary algorithms and reinforcement learning: A comprehensive survey on hybrid algorithms},
  year    = {2024},
}

@Article{Drugan2019,
  author  = {Madalina M. Drugan},
  journal = {Swarm and Evolutionary Computation},
  title   = {Reinforcement learning versus evolutionary computation: A survey on hybrid algorithms},
  year    = {2019},
  issn    = {2210-6502},
  pages   = {228-246},
  volume  = {44},
}

@article{gao2024generative,
  title={Generative learning for forecasting the dynamics of high-dimensional complex systems},
  author={Gao, Han and Kaltenbach, Sebastian and Koumoutsakos, Petros},
  journal={Nature Communications},
  volume={15},
  number={1},
  pages={8904},
  year={2024},
  publisher={Nature Publishing Group UK London}
}

@article{wang2023learning,
  title={Learning regularity for evolutionary multiobjective search: A generative model-based approach},
  author={Wang, Shuai and Zhou, Aimin and Zhang, Guixu and Fang, Faming},
  journal={IEEE Computational Intelligence Magazine},
  volume={18},
  number={4},
  pages={29--42},
  year={2023},
  publisher={IEEE}
}

@article{zhao2024policy,
  title={A policy-based meta-heuristic algorithm for energy-aware distributed no-wait flow-shop scheduling in heterogeneous factory systems},
  author={Zhao, Fuqing and Song, Lisi and Jiang, Tao and Wang, Ling and Dong, Chenxin},
  journal={IEEE Transactions on Systems, Man, and Cybernetics: Systems},
  year={2024},
  publisher={IEEE}
}

@article{yan2024emodm,
  title={Emodm: A diffusion model for evolutionary multi-objective optimization},
  author={Yan, Xueming and Jin, Yaochu},
  journal={arXiv preprint arXiv:2401.15931},
  year={2024}
}

@article{wang2025new,
  title={A new prediction strategy for dynamic multi-objective optimization using diffusion model},
  author={Wang, Feng and Xie, Jinsong and Zhou, Aimin and Tang, Ke},
  journal={IEEE Transactions on Evolutionary Computation},
  year={2025},
  publisher={IEEE}
}

@Article{Song2021,
  author   = {Song, Zhenshou and Wang, Handing and He, Cheng and Jin, Yaochu},
  journal  = {IEEE Transactions on Evolutionary Computation},
  title    = {A Kriging-Assisted Two-Archive Evolutionary Algorithm for Expensive Many-Objective Optimization},
  year     = {2021},
  number   = {6},
  pages    = {1013-1027},
  volume   = {25},
  comment  = {adaptive},
  doi      = {10.1109/TEVC.2021.3073648},
}

@Article{Wang2019,
  author   = {Wang, Xinjing and Wang, G. Gary and Song, Baowei and Wang, Peng and Wang, Yang},
  journal  = {IEEE Transactions on Evolutionary Computation},
  title    = {A Novel Evolutionary Sampling Assisted Optimization Method for High-Dimensional Expensive Problems},
  year     = {2019},
  number   = {5},
  pages    = {815-827},
  volume   = {23},
  comment  = {sample},
  doi      = {10.1109/TEVC.2019.2890818},
}

@Article{Gutmann2001,
  author     = {Gutmann, H.-M.},
  journal    = {J. of Global Optimization},
  title      = {A Radial Basis Function Method for Global Optimization},
  year       = {2001},
  issn       = {0925-5001},
  month      = mar,
  number     = {3},
  pages      = {201–227},
  volume     = {19},
  address    = {USA},
  comment    = {RBF},
  doi        = {10.1023/A:1011255519438},
  issue_date = {March 2001},
  numpages   = {27},
  publisher  = {Kluwer Academic Publishers},
  url        = {https://doi.org/10.1023/A:1011255519438},
}

@Article{Wang2023a,
  author   = {Wang, Weizhong and Liu, Hai-Lin and Tan, Kay Chen},
  journal  = {IEEE Transactions on Cybernetics},
  title    = {A Surrogate-Assisted Differential Evolution Algorithm for High-Dimensional Expensive Optimization Problems},
  year     = {2023},
  number   = {4},
  pages    = {2685-2697},
  volume   = {53},
  comment  = {multi models},
  doi      = {10.1109/TCYB.2022.3175533},
}

@Article{Zhen2023,
  author   = {Zhen, Huixiang and Gong, Wenyin and Wang, Ling},
  journal  = {IEEE Transactions on Evolutionary Computation},
  title    = {Evolutionary Sampling Agent for Expensive Problems},
  year     = {2023},
  number   = {3},
  pages    = {716-727},
  volume   = {27},
  comment  = {adaptive},
  doi      = {10.1109/TEVC.2022.3177605},
}

@article{huang2024evox,
  title={EvoX: A distributed GPU-accelerated framework for scalable evolutionary computation},
  author={Huang, Beichen and Cheng, Ran and Li, Zhuozhao and Jin, Yaochu and Tan, Kay Chen},
  journal={IEEE Transactions on Evolutionary Computation},
  year={2024},
  publisher={IEEE}
}

@Article{Liu2023b,
  author   = {Liu, Zhening and Wang, Handing and Jin, Yaochu},
  journal  = {IEEE Transactions on Cybernetics},
  title    = {Performance Indicator-Based Adaptive Model Selection for Offline Data-Driven Multiobjective Evolutionary Optimization},
  year     = {2023},
  number   = {10},
  pages    = {6263-6276},
  volume   = {53},
  comment  = {adaptive model selection},
  doi      = {10.1109/TCYB.2022.3170344},
}

@Article{Zhen2023a,
  author   = {Zhen, Huixiang and Gong, Wenyin and Wang, Ling and Ming, Fei and Liao, Zuowen},
  journal  = {IEEE Transactions on Cybernetics},
  title    = {Two-Stage Data-Driven Evolutionary Optimization for High-Dimensional Expensive Problems},
  year     = {2023},
  number   = {4},
  pages    = {2368-2379},
  volume   = {53},
  comment  = {multi model},
  doi      = {10.1109/TCYB.2021.3118783},
}

@Article{Wu2023,
  author   = {Wu, Xunfeng and Lin, Qiuzhen and Li, Jianqiang and Tan, Kay Chen and Leung, Victor C. M.},
  journal  = {IEEE Transactions on Cybernetics},
  title    = {An Ensemble Surrogate-Based Coevolutionary Algorithm for Solving Large-Scale Expensive Optimization Problems},
  year     = {2023},
  number   = {9},
  pages    = {5854-5866},
  volume   = {53},
  doi      = {10.1109/TCYB.2022.3200517},
}

@Article{Cai2020,
  author   = {Cai, Xiwen and Gao, Liang and Li, Xinyu},
  journal  = {IEEE Transactions on Evolutionary Computation},
  title    = {Efficient Generalized Surrogate-Assisted Evolutionary Algorithm for High-Dimensional Expensive Problems},
  year     = {2020},
  number   = {2},
  pages    = {365-379},
  volume   = {24},
  doi      = {10.1109/TEVC.2019.2919762},
}

@Article{Guo2019,
  author   = {Guo, Dan and Jin, Yaochu and Ding, Jinliang and Chai, Tianyou},
  journal  = {IEEE Transactions on Cybernetics},
  title    = {Heterogeneous Ensemble-Based Infill Criterion for Evolutionary Multiobjective Optimization of Expensive Problems},
  year     = {2019},
  number   = {3},
  pages    = {1012-1025},
  volume   = {49},
  doi      = {10.1109/TCYB.2018.2794503},
}

@inproceedings{eriksson2021scalable,
  title={Scalable constrained Bayesian optimization},
  author={Eriksson, David and Poloczek, Matthias},
  booktitle={International conference on artificial intelligence and statistics},
  pages={730--738},
  year={2021},
  organization={PMLR}
}

@article{zhou2025surrogate,
  title={Surrogate-Assisted Differential Evolution With Search Space Tightening for High-Dimensional Expensive Optimization Problems},
  author={Zhou, Rongfeng and Ren, Chongle and Meng, Zhenyu and Zhu, Haibin},
  journal={IEEE Transactions on Systems, Man, and Cybernetics: Systems},
  year={2025},
  publisher={IEEE}
}

@article{zhao2025surrogate,
  title={Surrogate-assisted large-scale expensive optimization enhanced by local exploitation},
  author={Zhao, Kaili and Wang, Xilu and Sun, Chaoli and Khan, Wazir Zada},
  journal={Complex \& Intelligent Systems},
  volume={11},
  number={12},
  pages={1--14},
  year={2025},
  publisher={Springer}
}

@article{li2023multitask,
  title={Multitask evolution strategy with knowledge-guided external sampling},
  author={Li, Yanchi and Gong, Wenyin and Li, Shuijia},
  journal={IEEE Transactions on Evolutionary Computation},
  year={2023},
  publisher={IEEE}
}

@article{guo2024classifier,
  title={A classifier-ensemble-based surrogate-assisted evolutionary algorithm for distributed data-driven optimization},
  author={Guo, Xiao-Qi and Wei, Feng-Feng and Zhang, Jun and Chen, Wei-Neng},
  journal={IEEE Transactions on Evolutionary Computation},
  year={2024},
  publisher={IEEE}
}

@article{wei2022distributed,
  title={Distributed and expensive evolutionary constrained optimization with on-demand evaluation},
  author={Wei, Feng-Feng and Chen, Wei-Neng and Li, Qing and Jeon, Sang-Woon and Zhang, Jun},
  journal={IEEE Transactions on Evolutionary Computation},
  volume={27},
  number={3},
  pages={671--685},
  year={2022},
  publisher={IEEE}
}

\clearpage

\onecolumn
\begin{@twocolumnfalse}
\begin{center}
    \fontsize{24}{29}\selectfont  Evolutionary Generative Optimization: Towards Fully Data-Driven Evolutionary Optimization via Generative Learning\\
    (Supplementary Document)
    \vspace{0.5em} 

\end{center}
\end{@twocolumnfalse}

\begin{center}
      Tao Jiang, 
        Kebin Sun,
        Zhenyu Liang,
        Ran Cheng,~
        Yaochu Jin,
        and Kay Chen Tan
\end{center}


\setcounter{algorithm}{0}
\setcounter{table}{0}
\setcounter{figure}{0}
\setcounter{section}{0}
\setcounter{equation}{0}
\renewcommand\thealgorithm{S\arabic{algorithm}}
\renewcommand\thetable{S.\Roman{table}}
\renewcommand{\figurename}{\normalsize Fig.}
\renewcommand\thefigure{S\arabic{figure}}
\renewcommand\thesection{S.\Roman{section}}
\renewcommand\theequation{S.\arabic{equation}}

\makeatother
\vspace{1em} 

\setcounter{page}{1}
\setcounter{section}{0}

\subsection{Complexity Analysis}\label{app:complexity}
    
    The overall computational cost of \MethodName mainly consists of data preparation, model training, and population generation. Let the population size be $P$, problem dimension $D$, number of training epochs $E$, and the depth and width of the generative model be $L_g$ and $h_g$, respectively. Since the surrogate model is a GP, its training on $P$ samples requires $O(P^3 + P^2 D)$ time and $O(P^2)$ memory due to kernel matrix construction and inversion, while a mean and variance prediction for each sample costs $O(P^2 + P D)$ owing to kernel evaluation and Cholesky-based inference. During the model training phase, the paired dataset construction yields approximately $P^2$ samples, and the bidirectional generative models are trained for $E$ epochs, resulting in a total cost of $T_{\text{train}} = O(P^3 + P^2 D) + O(E\,P^2\,L_g h_g^2)$. In the population generation phase, each of the $P$ individuals is transformed once by the trained generative model and evaluated by the true objective function, giving $T_{\text{gen}} = O(P\,(L_g h_g^2 + D))$. Summing up and keeping dominant terms, the overall per-generation complexity can be approximated as $T_{\text{total}} = O(P^3 + P^2 D + E\,P^2\,L_g h_g^2 + P D)$. The corresponding space complexity is mainly determined by the GP kernel matrix and the paired dataset, yielding $S_{\text{total}} = O(P^2 + P^2 D + L_g h_g^2) \approx O(P^2 D + L_g h_g^2)$. Although the algorithm exhibits relatively high theoretical time complexity compared with conventional evolutionary methods, it is highly parallelizable, and its practical runtime remains efficient due to GPU-based batch operations and concurrent model evaluations.
    
\subsection{Supplementary Justifications}\label{app:justify}

\subsubsection{Notations of Gaussian Process}
The Gaussian Process (GP) regression model, as described in \cite{Rasmussen2005}, defines the posterior probability distribution used in this section:

\begin{equation}
    \left(\tilde{\bm{f}} \, \middle| \, \bm{Q} = \left[\bm{q}_1, \ldots, \bm{q}_p\right], \bm{X}_t, \bm{y}_t\right) \sim \mathcal{N}\left(\bm{\mu}_t(\bm{Q}), \bm{\Sigma}_t(\bm{Q})\right),
\end{equation}

where $t$ is the step indicator, the mean vector is given by

\begin{equation}
    \bm{\mu}_t(\bm{Q}) = \kappa\left(\bm{Q}, \bm{X}_t\right)^\top \left[\kappa\left(\bm{X}_t, \bm{X}_t\right)\right]^{-1} \bm{y}_t,
\end{equation}

and the covariance matrix is defined as

\begin{equation}
    \bm{\Sigma}_t(\bm{Q}) = \kappa\left(\bm{Q}, \bm{Q}\right) - \kappa\left(\bm{Q}, \bm{X}_t\right)^\top \left[\kappa\left(\bm{X}_t, \bm{X}_t\right)\right]^{-1} \kappa\left(\bm{Q}, \bm{X}_t\right).
\end{equation}

We denote $\kappa(\cdot,\cdot)$ as the kernel function. The resulting covariance matrix $\kappa_t(\bm{Q})$ is required to be symmetric, positive semi-definite, and permutation-invariant, satisfying the following formal constraints:
\begin{align}
    &\kappa_t(\bm{Q}_\pi) = \pi\left(\kappa_t(\bm{Q})\right), \\
    &\kappa_t(\bm{Q}) = \kappa_t(\bm{Q})^\top, \\
    &\bm{v}^\top \kappa_t(\bm{Q}) \bm{v} \geq 0 \quad \forall \bm{v} \neq \bm{0},
\end{align}
where $\kappa_t(\cdot)$ represents the composed covariance matrix at step $t$, and $\bm{Q}_\pi$ is a permutation of $\bm{Q}$ adhering to the standard permutation notation $\pi$.

Given the inherent separability of common Radial Basis Function (RBF) and Mat\'ern kernels, we subsequently express $\kappa_t(\bm{Q})$ as

\begin{equation}
    \kappa_t(\bm{Q}) = \begin{bmatrix}
        \kappa(\bm{q}_1, \bm{q}_1) & \cdots & \kappa(\bm{q}_1, \bm{q}_p) \\
        \vdots & \ddots & \vdots \\
        \kappa(\bm{q}_p, \bm{q}_1) & \cdots & \kappa(\bm{q}_p, \bm{q}_p)
    \end{bmatrix},
\end{equation}

where 
\begin{align}
    \kappa(\bm{a}, \bm{b}) &= \alpha \exp{\left(-\|\bm{a} - \bm{b}\|^2\right)}, \\
    \kappa(\bm{a}, \bm{b}) &= \alpha \frac{2^{1-\nu}}{\Gamma(\nu)} \left(\sqrt{2\nu}\|\bm{a} - \bm{b}\|\right)^\nu K_\nu\left(\sqrt{2\nu}\|\bm{a} - \bm{b}\|\right),
\end{align}
for RBF and Mat\'ern kernels, respectively. 
The functions $\Gamma(\cdot)$ and $K_\nu(\cdot)$ denote the Gamma function and the modified Bessel function of the second kind, respectively. For clarity in the subsequent derivations, we will assume that the covariance matrix is separable throughout the remainder of this section.

\color{black}
\subsubsection{Justification of Optimization Loss $\mathcal{L}_{\text{opt}}$}
We initiate the formal derivation by defining the core components necessary for the optimization loss function. Recall that $\mu_t^* \equiv \max_{\bm{x}'} \mu_t(\bm{x}')$ represents the maximum of the posterior mean at step $t$. Furthermore, $\mu_{t+1}^* \equiv \max_{\bm{x}'} \mu_{t+1}(\bm{x}' \, | \, \bm{x}, f(\bm{x}))$ denotes the maximum of the posterior mean at step $t+1$, contingent upon observing the new solution and its fitness $(\bm{x}, f(\bm{x}))$. The optimization loss, $\mathcal{L}_{\text{opt}}$, is formally defined as:
\begin{align}
    \mathcal{L}_{\text{opt}} &= \mathcal{L}_{\text{mean}} + \mathcal{L}_{\text{std}} \equiv \mathbb{E} \left[ \mu_{t+1}^*-\mu_t^* \middle| \bm{x} \right] + \lambda_2 \mathbb{D} \left[ \mu_{t+1}^*-\mu_t^* \middle| \bm{x} \right]^{1/2} \nonumber \\
    &= \left( \mu_t(\bm{\gamma}(\bm{p})) - f(\bm{p}) \right) + \lambda_2 \left( \sigma_t(\bm{\gamma}(\bm{p})) + \rho_t(\bm{\gamma}'(\bm{\gamma}(\bm{p})), \bm{q}) + \rho_t(\bm{\gamma}(\bm{\gamma}'(\bm{q})), \bm{p}) \right),
\end{align}
For simplification, we denote

\begin{equation}
    \ell \equiv \mu_{t+1}^*-\mu_t^*.
\end{equation}

Let $\bm{x}^*_t$ denote the maximizer of the posterior mean at step $t+1$. We have

\begin{equation}
    \mu_{t+1}^* = \max_{\bm{x}'} \mu_{t+1}\left(\bm{x}'\middle|\bm{x},f(\bm{x})\right) = \max_{\bm{x}'} \left[{\kappa\left(\bm{x}', \left[\bm{X}_t, \bm{x}\right]\right)}^\top \left[\kappa\left(\left[\bm{X}_t, \bm{x}\right], \left[\bm{X}_t, \bm{x}\right]\right)\right]^{-1} \left[\bm{y}_t; f(\bm{x})\right]\right].
\end{equation}

Since $f(\bm{x})$ is unknown, we substitute it with the GP posterior at iteration $t$ \cite{Frazier2007,Frazier2009}:

\begin{equation}
    \mu_{t+1}\left(\bm{x}' \, \middle| \, \bm{x}, f(\bm{x})\right) = \mu_{t+1}\left(\bm{x}' \, \middle| \, \bm{x}, \mu_t(\bm{x}) + \Sigma_t(\bm{x}) Z\right),
\end{equation}

where $Z \sim \mathcal{N}(0,1)$ is a random variable of standard normal distribution.
With a separable covariance matrix, we obtain:
\begin{align}
    \mu_{t+1}\left(\bm{x}_t^* \, \middle| \, \bm{x}_t, \mu_t(\bm{x}_t) + \Sigma_t(\bm{x}_t) Z\right) = &\left[{\bm{k}^*}^\top, k_t^*\right] \cdot \left[\begin{matrix} \bm{K} & \bm{k}_t \\ \bm{k}_t^\top & \alpha \end{matrix}\right]^{-1}  \nonumber \\
    &\cdot \left[\bm{y}_t; \bm{k}_t^\top \bm{K}^{-1} \bm{y}_t + \left(\alpha - \bm{k}_t^\top \bm{K}^{-1} \bm{k}_t\right)^{1/2} Z\right],
    \label{eq:app-mu1}
\end{align}
where $\bm{x}_t^*$ is the maximizer of $\mu_{t+1}$, $\bm{k}_t \equiv \kappa(\bm{x}_t, \bm{X}_t)$, $\bm{k}^* \equiv \kappa(\bm{x}_t^*, \bm{X}_t)$, $k_t^* \equiv \kappa(\bm{x}_t^*, \bm{x}_t)$, $\bm{K} \equiv \kappa(\bm{X}_t, \bm{X}_t)$, and $\alpha \equiv \kappa(\bm{x}_t, \bm{x}_t)$.
Furthermore, the variable $\bm{x}_t$ in \cref{eq:app-mu1} is defined such that $\bm{x}_t = \bm{\gamma}(\bm{x})$. This mapping requires that $\bm{\gamma}$ acts as a bijection for the assumption to hold. We enforce this necessary invertibility of $\bm{\gamma}$ via the reconstruction loss term $\mathcal{L}_{\text{rec}}$ (Eq. 5) within $\mathcal{L}_{\text{gen}}$, which promotes the use of its pseudo-inverse model $\bm{\gamma}'$.

Applying the inverse of a block matrix and the Sherman-Morrison formula \cite{Sherman1950}, we derive:

\begin{equation}
    \left[{\bm{k}^*}^\top, k_t^*\right] \left[\begin{matrix} \bm{K} & \bm{k}_t \\ \bm{k}_t^\top & \alpha \end{matrix}\right]^{-1} = \left[{\bm{k}^*}^\top \bm{K}^{-1} + \frac{{\bm{k}^*}^\top \bm{K}^{-1} \bm{k}_t \bm{k}_t^\top \bm{K}^{-1} - k_t^* \bm{k}_t^\top \bm{K}^{-1}}{\alpha - \bm{k}_t^\top \bm{K}^{-1} \bm{k}_t}, \, \frac{k_t^* - {\bm{k}^*}^\top \bm{K}^{-1} \bm{k}_t}{\alpha - \bm{k}_t^\top \bm{K}^{-1} \bm{k}_t}\right].
\end{equation}

Consequently,
\begin{dmath}
    \mu_{t+1}\left(\bm{x}_t^* \, \middle| \, \bm{x}_t, \mu_t(\bm{x}_t) + \Sigma_t(\bm{x}_t) Z\right) \sim \mathcal{N}\left(\kappa(\bm{x}_t^*, \bm{X}_t)^\top {\kappa(\bm{X}_t, \bm{X}_t)}^{-1} \bm{y}_t, \, \frac{\left[\kappa(\bm{x}_t^*, \bm{x}_t) - {\kappa(\bm{x}_t^*, \bm{X}_t)}^\top {\kappa(\bm{X}_t, \bm{X}_t)}^{-1} \kappa(\bm{x}_t, \bm{X}_t)\right]^2}{\alpha - {\kappa(\bm{x}_t, \bm{X}_t)}^\top {\kappa(\bm{X}_t, \bm{X}_t)}^{-1} \kappa(\bm{x}_t, \bm{X}_t)}\right).
\end{dmath}
Using the covariance matrix of a bivariate normal distribution,

\begin{equation}
    \mathbf{\Sigma} = \left[\begin{matrix} \sigma_1^2 & \rho \sigma_1 \sigma_2 \\ \rho \sigma_1 \sigma_2 & \sigma_2^2 \end{matrix}\right],
\end{equation}

and by applying the correlation function definition, we formally derive:

\begin{equation}
    \mathbb{E}[\ell] = \mu_t(\bm{x}_t^*) - \mu_t^*, \quad \text{and} \quad \mathbb{D}[\ell] = \rho_t^2\left(\bm{x}_t^*, \bm{x}_t\right) \Sigma_t(\bm{x}_t^*),
    \label{eq:ev-loss-opt}
\end{equation}

where

\begin{equation}
    \rho_t\left(\bm{x}_1, \bm{x}_2\right) = \frac{{\Sigma_t\left(\left[\bm{x}_1, \bm{x}_2\right]\right)}_{1,2}}{\sqrt{\Sigma_t(\bm{x}_1) \Sigma_t(\bm{x}_2)}}
\end{equation}

is the GP correlation coefficient function at the current step.

If the solution at step $t$, $\bm{x}_{t}$, is also the maximizer of $\mu_{t}$, we can write $\bm{x}_t^* = \bm{x}_{t+1} = \bm{\gamma}(\bm{x}_t; \bm{\phi})$, i.e., the output of the generative model at step $t$ is the maximizer of $\mu_{t+1}$.
This assumption holds provided that the generated solutions are optimal, a necessary condition explicitly ensured by the inclusion of the mean loss term $\mathcal{L}_{\text{mean}}$ in the infill criteria, which incorporates a direct optimization component.
Then, \cref{eq:ev-loss-opt} leads to

\begin{equation}
    \mathbb{E}[\ell] + \lambda_2 \mathbb{D}[\ell]^{1/2} = \mu_t(\bm{\gamma}(\bm{x}_t)) - f(\bm{x}_t) + \lambda_2 \left[\Sigma_t(\bm{x}_t) + \Sigma_{t,12}^2(\bm{\gamma}(\bm{x}_t), \bm{x}_t)\right]^{1/2},
    \label{eq:app-ell2-1}
\end{equation}

The derivation proceeds by assuming that the derivative of $f(\bm{x}_t)$ and $\mu_t^*$ with respect to the generative model $\bm{\gamma}$ is zero, alongside the constraint that $\Sigma_{t,12}^2(\bm{\gamma}(\bm{x}_t), \bm{x}_t) + \Sigma_t(\bm{x}_t) \geq -2 \Sigma_{t,12}(\bm{\gamma}(\bm{x}_t), \bm{x}_t) \sqrt{\Sigma_t(\bm{x}_t)}$. Furthermore, with the assumption of paired generative modeling, i.e., $\bm{\gamma} \circ \bm{\gamma}' = \bm{\gamma}' \circ \bm{\gamma} = \text{identity}$ (with $\circ$ being the function composition), which holds because the reconstruction loss is minimized, we can simplify the above result to

\begin{equation}
    \mu_t(\bm{\gamma}(\bm{x}_t)) - f(\bm{x}_t) + \lambda_2 \left[\Sigma_t(\bm{x}_t) + \Sigma_{t,12}^2((\bm{\gamma}' \circ \bm{\gamma})(\bm{x}_t), \bm{x}_{t+1}) + \Sigma_{t,12}^2((\bm{\gamma} \circ \bm{\gamma}')(\bm{x}_{t+1}), \bm{x}_t)\right]^{1/2},
\end{equation}

noting the symmetry of $\Sigma_{t,12}$.

Finally, extending these results, we consider $\bm{p} \equiv \bm{x}_t \sim \bm{\mathcal{X}}_t$ and $\bm{q} \equiv \bm{x}_{t+1} \sim \bm{\mathcal{X}}_{t+1}$ as samples from distributions in the search space having \emph{inferior} and \emph{superior} function values, respectively. Based on this formulation, we derive the optimization loss term $\mathcal{L}_{\text{opt}}$ for training the generative model pair $\bm{\gamma}(\bm{p}; \bm{\phi})$ and $\bm{\gamma}'(\bm{q}; \bm{\phi}')$, simplifying notation as follows:
\begin{align}
    \mathcal{L}_{\text{opt}} &= \mathcal{L}_{\text{mean}} + \mathcal{L}_{\text{std}} \equiv \mathbb{E} \left[ \mu_{t+1}^*-\mu_t^* \middle| \bm{x} \right] + \lambda_2 \mathbb{D} \left[ \mu_{t+1}^*-\mu_t^* \middle| \bm{x} \right]^{1/2} \nonumber \\
    &= \left( \mu_t(\bm{\gamma}(\bm{p})) - f(\bm{p}) \right) + \lambda_2 \left( \sigma_t(\bm{\gamma}(\bm{p})) + \rho_t(\bm{\gamma}'(\bm{\gamma}(\bm{p})), \bm{q}) + \rho_t(\bm{\gamma}(\bm{\gamma}'(\bm{q})), \bm{p}) \right),
\end{align}
where $\mu_t$, $\sigma_t$, and $\rho_t$ are the mean, standard deviation, and correlation function of the surrogate model at step $t$, respectively.

If we adopt other commonly used infill criteria, the resulting optimization losses are expected to follow a similar structure. For instance, using the steps detailed above, we can derive the optimization loss corresponding to the Upper/Lower Confidence Bound (UCB/LCB):

\begin{equation}
    \mathcal{L}_{\text{opt}}^{\text{UCB/LCB}} = \mu_t(\bm{\gamma}(\bm{p})) - f(\bm{p}) + \lambda_2 \sigma_t(\bm{\gamma}(\bm{p})),
\end{equation}

A positive $\lambda_2$ corresponds to the Lower Confidence Bound (LCB) in our minimization context, whereas a negative $\lambda_2$ corresponds to the Upper Confidence Bound (UCB). Consequently, the boundary term $\mathcal{L}_{\text{opt}}^{\text{UCB/LCB}}$ is incorporated into the overall optimization loss defined in (Eq. 11).

\subsubsection{Justification of Generative Loss $\mathcal{L}_{\text{gen}}$ \texorpdfstring{\protect\footnotemark}{} }
\footnotetext{This analysis provides a foundational justification for the generative learning loss term $\mathcal{L}_{\text{gen}}$ from the perspective of Gaussian Processes (GP). It is presented as a supporting argument rather than a rigorous formal proof, and thus, some inherent assumptions may only strictly hold under specific or extreme boundary conditions.}

By limiting the number of input solutions to one, we can express the GP regression model as

\begin{equation}
    \left(\tilde{f} \, \middle| \, \bm{q}, \bm{X}_t, \bm{y}_t\right) \sim \mathcal{N}\left(\mu_t(\bm{q}), \Sigma_t(\bm{q})\right).
\end{equation}

Due to the linearity property of differentiation, the derivative of a GP model remains another GP model:

\begin{equation}
    \left(\nabla_{\bm{q}} \tilde{f} \, \middle| \, \bm{q}, \bm{X}_t, \bm{y}_t\right) \sim \mathcal{N}\left(\nabla_{\bm{q}} \mu_t(\bm{q}), \nabla_{\bm{q}}^2 \Sigma_t(\bm{q})\right),
\end{equation}

where

\begin{equation}
    \nabla_{\bm{q}}^2 \Sigma_t(\bm{q}) \equiv \left.\frac{\partial^2}{\partial \bm{u} \partial \bm{v}} \cov{\left[\tilde{f}(\bm{u}), \tilde{f}(\bm{v})\right]} \right|_{\bm{u}=\bm{q}, \bm{v}=\bm{q}}.
\end{equation}

Following \cite{Rasmussen2005}, the derivative of a GP model yields

\begin{equation}
    \left(\nabla_{\bm{q}} \tilde{f} \, \middle| \, \bm{q}, \bm{X}_t, \bm{y}_t\right) \sim \mathcal{N}\left(\mu_t'(\bm{q}), \bm{\Sigma}_t'(\bm{q})\right).
\end{equation}

Here, the mean and covariance matrix are given by:

\begin{equation}
    \mu_t'(\bm{q}) = \left[\nabla_{\bm{q}} \kappa\left(\bm{q}, \bm{X}_t\right)\right]^\top \left[\kappa\left(\bm{X}_t, \bm{X}_t\right)\right]^{-1} \bm{y}_t,
\end{equation}

Despite significant advancements in parallel computing hardware, the efficient implementation of high-throughput evolutionary search remains constrained by inherent structural and data management challenges. Consequently, achieving substantial speedups for population-based methods, especially those involving tree-like structures, requires specialized architectural solutions. To address this critical limitation, we introduce EvoGP, a novel framework designed to enable large-scale, population-level parallelism for Genetic Programming on modern GPU architectures. EvoGP facilitates this seamless integration through three primary, cohesive contributions. First, we develop a tensorized representation that standardizes variable-length tree structures for uniform execution. Second, we implement an adaptive population management strategy that mitigates thread divergence and optimizes resource utilization. Third, EvoGP embeds custom, specialized CUDA kernels to accelerate fundamental operations, thereby maximizing evaluation throughput.

\begin{equation}
    \bm{\Sigma}_t'(\bm{q}) = -\left[\nabla_{\bm{q}} \kappa\left(\bm{q}, \bm{X}_t\right)\right]^\top \left[\kappa\left(\bm{X}_t, \bm{X}_t\right)\right]^{-1} \nabla_{\bm{q}} \kappa\left(\bm{q}, \bm{X}_t\right),
\end{equation}

In these equations, the term $\nabla_{\bm{q}} \kappa\left(\bm{q}, \bm{X}_t\right) \in \mathbb{R}^{t \times d}$ denotes a matrix. Based on the preceding context, we further assume that $\kappa\left(\bm{q}, \bm{q}\right) = \alpha$ is a constant.

We proceed to perform gradient search by utilizing the gradient of the GP posterior as the surrogate gradient. In any gradient search procedure, the new query at step $t+1$ is derived directly from the gradient of the function evaluated at step $t$:

\begin{equation}
    \bm{x}_{t+1} = \bm{x}_t + \bm{H}_t\left( \nabla_{\bm{x}_t} f(\bm{x}_t) \right),
\end{equation}

where $\nabla_{\bm{x}_t}$ denotes the gradient operator, and $\bm{H}_t$ represents the vector modification operator, which is typically the negative learning rate or the negative Hessian. For simplicity, we assume that the gradient can be approximated by the gradient of GP and that $\bm{H}_t$ can be approximated by a matrix. Since the exact value of $f(\bm{x}_{t+1})$ is unknown, we aim to minimize the squared norm of its covariance matrix:

\begin{equation}
    \min \left\| \cov{\left[ \bm{x}_t + \bm{H}_t\nabla_{\bm{x}_t} \tilde{f}(\bm{x}_t) \right]} \right\|^2.
\end{equation}

The justification for this minimization corresponds directly to minimizing the loss term $\mathcal{L}_{\text{gen}}$. Subsequently, by incorporating the gradient of the GP model, we can obtain the derived expression:

\begin{equation}
    \bm{x}_t + \bm{H}_t\nabla_{\bm{x}_t} \tilde{f}(\bm{x}_t) = \bm{x}_t + \bm{H}_t \bm{k}_t^\top \bm{K}^{-1} \bm{y}_t + \bm{H}_t \left(-\bm{k}_t^\top \bm{K}^{-1} \bm{k}_t\right)^{1/2} Z,
\end{equation}

where $\bm{K}^{-1} \equiv \kappa\left(\bm{X}_t, \bm{X}_t\right)$, $\bm{k}_t \equiv \nabla_{\bm{x}_t} \kappa\left(\bm{x}_t, \bm{X}_t\right)$, and $Z \sim \mathcal{N}(0,1)$ denotes a standard normal distribution. 
Consequently, the distribution of $\bm{x}_{t+1}$ conditioned on $\bm{X}_t$ and $\bm{y}_t$ is

\begin{equation}
    \left(\bm{x}_{t+1} \, \middle| \, \bm{X}_t, \bm{y}_t\right) \sim \mathcal{N}\left(\bm{x}_t + \bm{H}_t \bm{k}_t^\top \bm{K}^{-1} \bm{y}_t, -\bm{H}_t \bm{k}_t^\top \bm{K}^{-1} \bm{k}_t \bm{H}_t^\top\right).
\end{equation}

By applying expectations on both sides, we obtain

\begin{equation}
    \mathbb{E}\left[\bm{x}_{t+1}\right] = \bm{x}_t + \bm{H}_t \bm{k}_t^\top \bm{K}^{-1} \bm{y}_t \equiv \bm{x}_t + \bm{H}_t \bm{k}_t^\top \bm{L}^{-\top} \bm{L}^{-1} \bm{y}_t \equiv \bm{x}_t + \mathbf{P}(\bm{x}_t) \bm{z}_t,
\end{equation}

where $\bm{L} \bm{L}^\top \equiv \bm{K} = \kappa\left(\bm{X}_t, \bm{X}_t\right)$ is the Cholesky decomposition of $\bm{K}$, $\bm{z}_t \equiv \bm{L}^{-1} \bm{y}_t$, and $\mathbf{P}(\bm{x}_t) \equiv \bm{H}_t \bm{k}_t^\top \bm{L}^{-\top} = \bm{H}_t \left[\nabla_{\bm{x}_t} \kappa\left(\bm{x}_t, \bm{X}_t\right)\right]^\top \bm{L}^{-\top} \in \mathbb{R}^{d \times n}$ is a function of $\bm{x}_t$. The specific conditions are omitted for simplicity.

We assume that the difference $\bm{v}$ between the output of the generative model $\bm{\gamma}$ at step $t$, $\bm{\gamma}(\bm{x}_t;\bm{\phi})$, and the solution at step $t$, $\bm{x}_t$, is relatively small:

\begin{equation}
    \left\|\bm{v}\right\| \ll \left\|\bm{\gamma}(\bm{x}_t;\bm{\phi})\right\|, \text{ and } \left\|\bm{v}\right\| \ll \left\|\bm{x}_t\right\|, \text{ where } \bm{v}\equiv\bm{\gamma}(\bm{x}_t;\bm{\phi}) - \bm{x}_t.
\end{equation}

If $\bm{\gamma}$ only marginally modifies the inputs, the generative model can be approximated by a linear map $\bm{G}$. Furthermore, assuming $\bm{G}$ maps $\bm{x}_t$ to $\bm{x}_{t+1}$ and that both distributions are congruent, we can presume $\bm{G}$ to be an isometry. Since the simplest isometry form is realized by a unitary matrix, which implies $\bm{G}^\top = \bm{G}^{-1}$, we utilize this property. Consequently, $\bm{G}$ is replaced by the parametric generative model $\bm{\gamma}(\bm{x}_t; \bm{\phi})$, and $\bm{G}^\top$ is approximated by the parametric generative model $\bm{\gamma}'(\bm{x}_t; \bm{\phi}') \approx \bm{\gamma}^{-1}(\bm{x}_t; \bm{\phi})$. The selection of $\bm{\gamma}'$ over the direct inverse $\bm{\gamma}^{-1}$ is motivated by two key issues: first, the potential non-invertibility of an arbitrary parametric model; and second, the unsuitability of a direct inverse for limited-precision computation due to possible ill-conditioning. Based on these arguments, we formalize the subsequent assumption: when $\bm{q}$ is near the solutions of step $t$, the generative model $\bm{\gamma}$ and its pseudo-inverse $\bm{\gamma}'$ at step $t$ can be approximated as an orthogonal linear operator $\bm{G}_t$ and its inverse $\bm{G}^{-1}_t$, respectively.

\begin{equation}
    \bm{\gamma}(\bm{q};\bm{\phi}) \simeq \bm{G}_t \bm{q} \text{ and } \bm{\gamma}'(\bm{q};\bm{\phi}') \simeq \bm{G}^{-1}_t \bm{q} \text{ given } \bm{q}\in\mathcal{B}_r(\bm{x}_t) \text{ and } \bm{G}^\top_t = \bm{G}^{-1}_t,
\end{equation}

where $\mathcal{B}_r(\bm{x}_t)$ is the open ball neighborhood of $\bm{x}_t$ with radius $r$.

Based on these foundational assumptions, we bilinearize $\mathbf{P}$ into a tensor $\bm{P}^{ijk}$, yielding

\begin{equation}
    \mathbb{E}\left[\bm{x}_{t+1}\right] = \bm{x}_t + \bm{P}^{ijk} \bm{x}_t^i \bm{z}_t^j \equiv \bm{P}' \bm{x}_t,
\end{equation}

The subsequent derivation employs the Einstein summation convention for repeated indices.
Aiming to minimize covariance, we have:

\begin{equation}
    \cov\left[\bm{x}_{t+1}\right] = \bm{H}_t \bm{k}_t^\top \bm{K}^{-1} \bm{k}_t \bm{H}_t^\top.
\end{equation}

Under the same assumptions, we can also write

\begin{equation}
    \bm{\gamma}(\bm{x}_t) \equiv \bm{G} \bm{x}_t = \bm{P}' \bm{x}_t,
\end{equation}

Consequently,

\begin{equation}
    \cov\left[\bm{x}_{t+1}\right] = -\frac{\bm{z}_t \bm{z}_t^\top}{{\|\bm{z}_t\|}^2} \bm{x}_t^\top \left(\bm{G} \bm{G}^\top + \bm{I} - \bm{G} - \bm{G}^\top\right) \bm{x}_t,
\end{equation}

based on the linearization approach described earlier. 
The covariance comprises two components. Given that the former component ($\bm{z}_t \bm{z}_t^\top / {\|\bm{z}_t\|}^2$) is a projection matrix, we focus exclusively on the scalar part of the expression:
\begin{align}
    \bm{x}_t^\top \left(\bm{G} \bm{G}^\top + \bm{I} - \bm{G} - \bm{G}^\top\right) \bm{x}_t = &\left[\left(\bm{x}_t, (\bm{\gamma}' \circ \bm{\gamma})(\bm{x}_t)\right) - \left(\bm{x}_t, \bm{x}_t\right)\right] \nonumber \\
    &\phantom{a} - \lambda \left[\left(\bm{x}_{t+1}, \bm{\gamma}(\bm{x}_t)\right) - \left(\bm{x}_{t+1}, \bm{x}_{t+1}\right)\right],
    \label{eq:app-loss-gen}
\end{align}
where $\lambda$ is a scaling factor used to balance the variances of the distributions for $\bm{x}_{t+1}$ and $\bm{x}_{t}$, $\circ$ denotes function composition, and $(\bm{x},\bm{y})$ represents the inner product of two vectors. 
Given that

\begin{equation}
    \|\bm{x} + \Delta \bm{v} - \bm{x}\|^2 = 2 \left(\|\bm{x}\|^2 - \left(\bm{x}, \bm{x} + \Delta \bm{v}\right)\right)
\end{equation}

Due to the symmetry of \cref{eq:app-loss-gen} with respect to $\bm{G}$ and $\bm{G}^\top$, the covariance minimization reformulates to

\begin{equation}
    \left\| (\bm{\gamma}' \circ \bm{\gamma})(\bm{x}_t) - \bm{x}_t \right\|^2 + \left\| (\bm{\gamma} \circ \bm{\gamma}')(\bm{x}_{t+1}) - \bm{x}_{t+1} \right\|^2 + \lambda \left\| \bm{\gamma}(\bm{x}_t) - \bm{x}_{t+1} \right\|^2 + \lambda \left\| \bm{\gamma}'(\bm{x}_{t+1}) - \bm{x}_t \right\|^2,
\end{equation}

where \emph{$\circ$} represents the function composition.
In this context, $\bm{\gamma} : \bm{\mathcal{X}}_t \rightarrow \bm{\mathcal{X}}_{t+1}$ and $\bm{\gamma}' \approx \bm{\gamma}^{-1}$ represent two unknown parametric mappings, which are modeled using neural network generative models.

Finally, the generative loss, $\mathcal{L}_{\text{gen}}$, can be justified from the perspective of GP:
\begin{align}
    \mathcal{L}_{\text{gen}} = \left( \|\bm{\gamma}(\bm{p}) - \bm{q}\|^2 + \|\bm{\gamma}'(\bm{q}) - \bm{p}\|^2 \right) + \lambda_1 \left( \|\bm{\gamma}'(\bm{\gamma}(\bm{p})) - \bm{p}\|^2 + \|\bm{\gamma}(\bm{\gamma}'(\bm{q})) - \bm{q}\|^2 \right).
\end{align}

\subsection{Experimental Setup and Implementation Details}\label{app:exp}

\subsubsection{Implementation Details}\label{app:impl}

The primary implementation of \MethodName\ is based on JAX (v0.4.16) with XLA enabled, utilizing \texttt{jit} and \texttt{vmap} for efficient vectorized computation on a single GPU without cross-device communication. For framework validation, we also implemented a PyTorch version (v2.5.1) and observed comparable runtime efficiency and optimization performance. Profiling with \texttt{jax.profiler} indicates that the \emph{parallelization overhead}, which arises primarily from occasional XLA compilation and kernel launch scheduling for batched execution, accounts for less than 5\% of the total wall-clock time. Crucially, all compared methods were executed under identical hardware and software configurations, ensuring that this minimal overhead does not impact the fairness of the comparisons.

All of the experiments were conducted using Ubuntu 22.04, Python 3.10, and NVIDIA CUDA 12.4, on hardware equipped with NVIDIA A100 GPUs and Intel Xeon Gold 6226R CPUs.

\subsubsection{Setup Details}\label{app:setup}
In \MethodName, the dataset is configured according to Algorithm 1, using a split factor $\eta=0.1$ and a sliding window factor $\epsilon=0.3$. The initial dataset, generated via Latin Hypercube Sampling within normalized predefined regions ($[0,1]$ for all dimensions), is sized at 1000. We analyze the impact of these hyperparameters in \cref{app:ablation}. To manage dataset evolution, each iteration of Algorithm 1 incorporates a sorting step: $\mathcal{D}_{t+1} \leftarrow \left\{ (\bm{x}, f) \in \mathcal{D}_{t} \cup \mathcal{D}_{t+1} \middle| f < \mathrm{median}(\mathcal{F}_{t} \cup \mathcal{F}_{t+1}) \right\}$, where $\mathcal{F}_t$ denotes the function values in $\mathcal{D}_t$. This procedure allows dataset selection based solely on $\mathcal{D}_{t-1}$ and subsequently facilitates the splitting process on $\mathcal{D}_t$ detailed in Algorithm 2.

The GP model within \MethodName\ employs the Mat\'ern covariance function with $\nu = 5/2$. The optimization of the GP model's learnable parameters utilizes the Marginal Log Likelihood (MLL) loss and the Adam optimizer \cite{Kingma2015} ($\beta_1 = 0.9$, $\beta_2 = 0.999$, $\varepsilon = 1 \times 10^{-8}$). Training is limited to 1000 epochs across all benchmarks. The learning rate is determined by $|\tilde{\mathcal{D}}| / (6 \times 10^4)$ to accommodate the inherently small gradient of the MLL loss function.

The generative model in \MethodName\ is structured as a multi-layer perceptron (MLP) with 5 hidden layers. Given input/output dimension $d$, the layer sizes are sequentially defined as $\max{\{2d,128\}}$, $\max{\{4d,256\}}$, $\max{\{4d,256\}}$, $\max{\{4d,256\}}$, and $\max{\{2d,128\}}$. Further analysis of these architectural choices is provided in \cref{app:ablation}.

Generative model training parameters include a maximum of 200 epochs and a batch size equal to $|\mathcal{D}_t|$. We use the same Adam optimizer settings as the GP model, setting the learning rate to $0.015 / |\tilde{\mathcal{D}}|$. The loss function hyperparameters are specifically configured: $\lambda_1 = 400$ for reconstruction loss, $\lambda = 0.1$ for optimization loss, and $\lambda_2 = 10$ for standard deviation loss.

Furthermore, we utilize a noisy Mat\'ern covariance function kernel in the GP model of \MethodName.

\begin{equation}
	\kappa_\nu(d_{ij}) = \sigma^2 \frac{2^{1-\nu}}{\Gamma(\nu)} \left(\sqrt{2\nu} d_{ij}\right)^\nu K_\nu\left(\sqrt{2\nu} d_{ij}\right) + \varepsilon \delta_{ij},
\end{equation}

where $\sigma$ is the scale factor, $d_{ij}$ is the squared distance between normalized sample solutions, $\Gamma$ denotes the gamma function, $K_\nu$ represents the modified Bessel function of the second kind, $\nu$ is the parameter influencing the covariance, $\varepsilon$ is the noise factor, and \change{$\delta_{ij}$} is the delta function for $i=j$.

\subsection{Benchmark Problems}

\SmallTitle{Numerical Functions}
The formulations for four numerical functions are detailed in \cref{tab:numerical}.

\tabNumerical

\SmallTitle{Pushing}
Introduced by \cite{Wang2018}, this problem involves a scenario where a two-handed robot is tasked with relocating two objects to designated locations. The task necessitates the optimization of $d = 14$ parameters, which define the robot's hand positions, orientations, push velocities, directions, and durations, to achieve the desired object placements.

\SmallTitle{Landing}
Derived from the OpenAI Gym \cite{Towers2023}, this problem focuses on controlling a lunar lander to optimize fuel consumption and landing precision while avoiding crash landings. The control strategy is defined by $d = 12$ parameters (provided by OpenAI Gym) and is assessed based on the final reward obtained within a maximum of 1000 steps.

\SmallTitle{Rover}
First described by \cite{Wang2018}, this problem entails optimizing the trajectory of a rover navigating through challenging terrain. The trajectory is defined by a B-spline curve, which uses 30 control points on a 2D plane. Optimizing this path requires adjusting a total of $d = 60$ parameters.

\SmallTitle{Walker}
Derived from the OpenAI Gym \cite{Towers2023}, this problem focuses on controlling a robot's dual set of legs to maximize forward walking distance. The control strategy is implemented as a simple MLP with one hidden layer of 8 neurons. This network accepts 17 observation values as input and outputs 6 action values. Consequently, the model encompasses 198 trainable parameters with a range of $[-5, 5]$, defining the search space for this problem.

\SmallTitle{Ant}
Derived from the OpenAI Gym \cite{Towers2023}, this problem requires optimizing a four-legged robotic agent tasked with navigating complex terrain. The control mechanism is a neural network model, specifically an MLP with one hidden layer consisting of 11 neurons. This network processes 27 observation inputs and outputs 8 distinct action commands. This setup encapsulates 404 trainable parameters, ranging from $[-5, 5]$, thereby defining the search space for optimal locomotion strategies.

\subsection{Supplementary Experiments}\label{app:more-exp}
We report the complete results for the same batch size experiment (Fig. 5) in \cref{fig:same-batch-full}. Furthermore, the complete results for the main ablation experiment (Fig. 8) are provided in \cref{fig:ablation-full}.

\figResultAllFullA

\figResultAllFullB

\figSameBatchFull

\figAblationFull

\renewcommand{\thetable}{S.\Roman{table}}

\begin{table}[t]
\centering
\caption{Statistical Results (Mean and Standard Deviation) of the Tested Algorithms on 5-Dimensional Problems under 100 Function Evaluations. Each Algorithm Was Executed Ten Times, and the Best Results Are Highlighted in Bold.}
\label{tab:evogo_5d_100fe}
\resizebox{\textwidth}{!}{
\begin{tabular}{lcccccc} 
\toprule
Problem & BOGP & TPE & SADE-SS & EvoGO (No Aug) & EvoGO (VAE-100) & EvoGO (VAE-1000)\\
\midrule
Ackley     & 1.9952e$+$01 (5.1820e$+$00) & 1.6320e$+$01 (3.0432e$+$00) & 1.6496e$+$01 (3.8948e$+$00) & 1.4231e$+$01 (1.8130e$+$00) & \textbf{6.8832e$+$00 (5.0488e$+$00)} & 1.2975e$+$01 (3.4978e$+$00) \\
Rosenbrock & 7.4111e$+$05 (1.4876e$+$06) & 8.3383e$+$05 (1.9668e$+$06) & 7.0584e$+$04 (1.0762e$+$05) & 3.5799e$+$04 (5.2380e$+$04) & \textbf{1.7382e$+$04 (5.1928e$+$04)} & 5.5068e$+$04 (6.5246e$+$04) \\
Rastrigin  & 3.3367e$+$03 (6.8820e$+$03) & 1.9304e$+$03 (2.0557e$+$03) & 5.2282e$+$02 (5.0540e$+$02) & 9.9146e$+$02 (8.3324e$+$02) & 4.7234e$+$02 (1.0223e$+$03) & \textbf{6.6046e$+$02 (1.0606e$+$03)} \\
Levy       & 2.9658e$+$03 (3.5070e$+$03) & 6.3936e$+$02 (1.0340e$+$03) & 2.8033e$+$02 (2.0761e$+$02) & 5.0654e$+$02 (7.7947e$+$01) & 4.4347e$+$02 (2.3513e$+$02) & \textbf{1.3794e$+$02 (1.9793e$+$02)} \\
\bottomrule
\end{tabular}
}
\end{table}

\subsection{Supplementary Ablation Studies}\label{app:ablation}
We report additional ablation studies concerning the influence of the neural network architectures in \cref{fig:nn-arch-all}, and the settings of various hyperparameters in \crefrange{fig:cyc-loss-all}{fig:slide-all}.

\SmallTitle{Ablation Study on VAE-Based Data Augmentation}
To evaluate the effectiveness of VAE-based data augmentation within EvoGO, an ablation experiment was conducted on four standard 5-dimensional benchmark problems: Ackley, Rosenbrock, Rastrigin, and Levy. This experiment was performed under a strict evaluation budget of 100 function evaluations to assess algorithmic efficiency in data-scarce conditions. For population-based frameworks, the population size was uniformly set to 100. Each algorithm was independently executed ten times to account for stochastic variations, and the mean and standard deviation of the final objective values were recorded. Three key variants of EvoGO were included for comparison. The first variant, EvoGO (No Aug), performs optimization without any data augmentation. The second variant, EvoGO (VAE-100), employs 100 solutions generated by the VAE model to enrich the search space. The third variant, EvoGO (VAE-1000), uses 1000 generated samples for augmentation to expand the candidate pool further. These variants were also compared against representative baselines, including BOGP, TPE, and SADE-SS.

As reported in Table~\ref{tab:evogo_5d_100fe}, EvoGO demonstrates consistent superiority over the baseline algorithms under this limited-budget setting. Both augmented variants achieve significant improvements over the non-augmented version (EvoGO (No Aug)), confirming the effectiveness of incorporating learned samples from the VAE model. Among the augmented variants, EvoGO (VAE-100) achieves the most stable performance, suggesting that a moderate level of augmentation provides a favorable balance between exploration and model reliability. In contrast, the use of excessive augmentation, as seen in EvoGO (VAE-1000), can lead to marginal performance degradation due to the accumulation of less reliable synthetic samples. Consequently, these results validate that VAE-based augmentation effectively enhances the search capability and robustness of EvoGO under constrained evaluation conditions.
\color{black}

\SmallTitle{Neural Network Architectures}
Figure \cref{fig:nn-arch-all} evaluates the influence of different neural network configurations within \MethodName, including standard (\MethodName), deeper (\MethodName-Deep), and wider (\MethodName-Wide) networks. The hidden layers of the wider networks are sized at $\max{\{4d,256\}}$, $\max{\{6d,384\}}$, and $\max{\{4d,256\}}$, respectively. Conversely, the hidden layers of the deeper networks are sized at $\max{\{2d,128\}}$, $\max{\{3d,192\}}$, $\max{\{3d,192\}}$, $\max{\{3d,192\}}$, $\max{\{3d,192\}}$, $\max{\{3d,192\}}$, and $\max{\{2d,128\}}$, respectively. The resulting number of parameters for these architectures is shown in \cref{tab:params}. Overall, the findings suggest that \MethodName\ is robust to architectural changes, with the standard configuration generally providing the most effective results.

\tabParams

\SmallTitle{Reconstruction Loss Scale ($\lambda_1$)}  
The influence of the reconstruction loss scale $\lambda_1$ is examined in \cref{fig:cyc-loss-all}. Across tested values of 800, 400, 200, and 100, \MethodName\ exhibits only minor variations in performance. This indicates robust stability and low sensitivity to the setting of this hyperparameter.

\SmallTitle{Dataset Split Factor ($\eta$)}
Figure \cref{fig:portion-all} investigates the effect of varying the dataset split factor $\eta$. Although optimal performance is observed at $\eta = 0.1$, alternative values also yield competitive outcomes, highlighting the method's inherent adaptability.

\SmallTitle{Sliding Window Factor ($\epsilon$)} 
Figure \cref{fig:slide-all} assesses the impact of different $\epsilon$ settings on performance. The method achieves optimal results at $\epsilon = 0.3$, suggesting this value provides an effective balance between computational efficiency and search effectiveness.

\figNnArchAll

\figCycLossAll

\figPortionAll

\figSlideAll

\color{black}
\subsection{Other Limitations}\label{app:limit}
While \MethodName\ demonstrates considerable strengths in online optimization settings, it is important to recognize several limitations that must be addressed in future work to ensure broader applicability and performance robustness.

Firstly, the algorithm heavily relies on the availability of sufficient online data streams. This dependence can be a significant constraint in data-scarce environments, where the absence of sufficient online data leads to suboptimal performance because the model lacks the necessary information for accurate predictions and dynamic adjustments.

Moreover, the current implementation of \MethodName\ does not fully exploit the potential benefits of incorporating offline data. Excluding historical datasets means the algorithm overlooks valuable information that could substantially enhance its training process and improve predictive accuracy. This limitation becomes particularly evident when analyzing complex, long-term patterns that require extensive historical data capture.

Additionally, \MethodName\ may underperform when applied to highly irregular optimization landscapes. In environments characterized by unpredictable or erratic patterns, the algorithm's reliance on data preparation via splitting may be insufficient to navigate such complex terrains effectively. Nevertheless, empirical evidence shows \MethodName\ only slightly underperforms in classic control benchmarks, suggesting that typical real-world optimization landscapes possess sufficient regularity for effective application of the method.

\end{document}